%% file: main.tex
\documentclass[journal]{IEEEtran}

\pdfoutput=1
\usepackage[utf8]{inputenc}
\usepackage[T1]{fontenc}
\usepackage{amsmath}
\usepackage{amsfonts}
\usepackage{amssymb}
\usepackage{graphicx}
\usepackage{color}
\usepackage{array}
\usepackage{booktabs}
\usepackage{amsthm}
\usepackage[pdftex]{hyperref}

\theoremstyle{definition}
\newtheorem{definition}{Definition}[section]

\title{Time-Scale Coupling Between States and Parameters in Recurrent Neural Networks}

\author{Lorenzo Livi$^{*}$
\thanks{*lorenz.livi@gmail.com}
\thanks{Lorenzo Livi is with the Open Institute of Technology, The Core, Triq Il-Wied Ta' L-Imsida, Msida, MSD 9021, Malta}}

\begin{document}

\maketitle

\begin{abstract}
We show that gating mechanisms in recurrent neural networks (RNNs) induce lag-dependent and direction-dependent effective learning rates, even when training uses a fixed, global step size. This behavior arises from a coupling between state-space time-scales (parametrized by the gates) and parameter-space dynamics during gradient descent. By deriving exact Jacobians for leaky-integrator and gated RNNs and applying a first-order expansion, we make explicit how constant, scalar, and multi-dimensional gates reshape gradient propagation, modulate effective step sizes, and introduce anisotropy in parameter updates. These findings reveal that gates act not only as filters of information flow, but also as data-driven preconditioners of optimization, with formal connections to learning-rate schedules, momentum, and adaptive methods such as Adam. Empirical simulations corroborate these predictions: across several sequence tasks, gates produce lag-dependent effective learning rates and concentrate gradient flow into low-dimensional subspaces, matching or exceeding the anisotropic structure induced by Adam.
Notably, gating and optimizer-driven adaptivity shape complementary aspects of credit assignment: gates align state-space transport with loss-relevant directions, while optimizers rescale parameter-space updates. Overall, this work provides a unified dynamical systems perspective on how gating couples state evolution with parameter updates, clarifying why gated architectures achieve robust trainability in practice.
\end{abstract}
\begin{IEEEkeywords}
Recurrent neural networks; Dynamical systems; Gates; Time-scale coupling; Gradient descent.
\end{IEEEkeywords}

\section{Introduction}

Training recurrent neural networks (RNNs) is typically analyzed from two largely separate perspectives. On the one hand, state-space dynamics emphasize how gating mechanisms stabilize hidden trajectories, regulate memory retention, and mitigate vanishing or exploding gradients. On the other hand, parameter-space dynamics emphasize the role of optimization algorithms--such as momentum or Adam--in adapting learning rates and reshaping update directions to ease training. Yet these two domains interact in ways that have not been made explicit.

Why do gated RNNs often train stably even with plain gradient descent? This robustness suggests that gates must influence not only the flow of information in state space, but also the dynamics of parameter updates. However, the precise mechanisms by which state-space time-scales couple into optimization have remained implicit.

In this work, we show that gating mechanisms directly shape the dynamics of parameter updates.
State-space time scales, parametrized by gates, determine the structure of Jacobian products in backpropagation through time.
These products, in turn, induce lag-dependent \emph{effective learning rates} and directional anisotropy in parameter updates, even when the optimizer itself is non-adaptive.
From this perspective, gates function as implicit, data-driven preconditioners of the optimization process, with formal connections to learning-rate schedules, momentum, and Adam--but here these behaviors emerge endogenously from gating rather than being externally imposed.
Our contributions are:
\begin{itemize}
    \item We show analytically that gates act as parametrized time scales that modulate lag-dependent effective learning rates. Our analysis introduces a perturbative expansion of Jacobian products, combining Fr\'{e}chet derivative techniques with the structure of gated recurrent dynamics, making explicit how constant, scalar, and multi-gate RNNs modulate effective learning rates.
    \item We extend the theoretical analysis to directional effects, introducing the anisotropy index to quantify how gates shape the dominant subspaces of gradient propagation.
    \item We validate these predictions through targeted simulations on canonical tasks. Our results highlight that both gating and optimizer dynamics shape not only the magnitude but also the directional structure of temporal credit assignment, with task-dependent trade-offs.
\end{itemize}

The remainder of the paper is organized as follows.
Section~\ref{sec:related_works} reviews related work.
Section~\ref{sec:time_scales} introduces the interpretation of gating
mechanisms as parametrized time scales and formalizes the RNN models
considered in our analysis.
Section~\ref{sec:rnn-models-affect-learning-dynamics} develops the
theoretical results linking state-space time scales with parameter
dynamics, while Section~\ref{sec:gradient_descent_algorithms}
discusses connections with classical optimization methods.
Section~\ref{sec:simulations} presents empirical validations on
canonical sequence tasks, and
Section~\ref{sec:conclusions} concludes the paper.
Supplementary material provides the derivation of the matrix-product
expansion used in the analysis together with additional experimental
figures.

\section{Related work}
\label{sec:related_works}

The study of RNNs has long been shaped by the well-known vanishing and exploding gradient problem~\cite{pascanu2013difficulty}, which affects the stability and learnability of long-term dependencies. While this issue traditionally focuses on the magnitude of gradients, recent work by~Zucchet et al.~\cite{zucchet2024recurrent} highlights a complementary phenomenon: as the memorization capacity of an RNN increases during training, the network output can become highly sensitive to small parameter variations. This heightened sensitivity arises even in the absence of vanishing or exploding gradients, suggesting that RNNs may undergo abrupt changes in behavior due to intrinsic instabilities in the learned parameter–state mapping. In a related vein, Ceni \cite{10905028} proposed Random Orthogonal Additive Filters, a class of architectures designed to stabilize and enrich recurrent dynamics via orthogonal transformations, mitigating instability without sacrificing representational power.

Several lines of research have explored simplified linear models as a means of studying and designing recurrent dynamics with greater interpretability. Continuous-time state-space models, such as those in~\cite{gu2021efficiently,gu2021combining}, have enabled efficient training of sequence models with long-range dependencies by leveraging structured state-space kernels. More recently, Muca et al.~\cite{muca2024theoretical} provided a theoretical analysis of linear state-space models, clarifying their expressive power and the role of parameterization in controlling stability and memory retention. These models offer a valuable bridge between classical control theory and modern deep learning approaches to sequence modeling.

From a training dynamics perspective, Lee et al.~\cite{lee2019wide} analyzed wide neural networks through the lens of the neural tangent kernel, providing insights into the coupling between parameters and outputs during optimization. Although this framework is often applied to feedforward networks, it has implications for understanding the parameter–state interaction in recurrent settings, particularly in regimes where network width and gating jointly influence effective learning rates. Complementary work by~Saxe et al. \cite{saxe2013exact} investigated exact learning dynamics in deep linear networks, showing how singular value spectra dictate time scales of parameter evolution. This resonates with our perspective that recurrent gates modulate effective time scales in both state and parameter spaces.

Another rich research area involves constraining recurrent weight matrices to improve stability, expressivity, or trainability. Orthogonal and unitary RNNs~\cite{pmlr-v80-helfrich18a,10.5555/3305890.3305929,arjovsky2016unitary,NIPS2016_6327,vorontsov2017orthogonality} maintain constant gradient norms over time, thus alleviating the vanishing/exploding gradient problem. Lipschitz RNNs~\cite{lipschitz_rnn} extend this idea by explicitly controlling the Lipschitz constant to ensure robustness. Non-normal RNNs~\cite{kerg2019non} and their variants incorporating Schur decompositions~\cite{kag2019rnns} have been proposed to exploit transient amplification phenomena for richer dynamics. Other designs include antisymmetric RNNs~\cite{chang2018antisymmetricrnn}, which draw inspiration from Hamiltonian systems to ensure stability, coupled RNNs~\cite{rusch2021coupled}, which model interacting subsystems, and multiscale RNNs~\cite{rusch2021long}, which embed explicit time-scale separation into the architecture.

Gating mechanisms have received particular attention for their role in modulating information flow and improving trainability. From a theoretical standpoint, \cite{pmlr-v80-chen18i} and \cite{gilboa2019dynamical} analyzed the effects of gating on dynamical isometry and mean-field properties in RNNs, showing that gates can help preserve gradient flow and condition the optimization landscape. Related analyses in feedforward networks~\cite{pennington2017resurrecting} underline the generality of dynamical isometry principles across architectures. Architecturally, stacked gated RNNs~\cite{9373965} and hybrid designs such as the ``Just Another Network''~\cite{van2018unreasonable} demonstrate the versatility of gating in achieving both memory retention and efficient optimization. More recent theoretical work further strengthens this connection: Krishnamurthy et al.~\cite{krishnamurthy2022theory} and Can et al.~\cite{can2020gates} showed that gates can create slow modes in recurrent dynamics, directly modulating Jacobian spectra and shaping effective memory time scales. Empirical studies such as~\cite{quax2020adaptive} confirmed that RNNs tend to develop adaptive time scales when trained on multi-scale sequential data, further illustrating the deep connection between gating and temporal structure.

A distinct line of research studies RNNs directly on a mathematical time scale in the sense of Hilger's time-scale calculus, where the time domain itself unifies continuous and discrete dynamics (see, e.g., Martsenyuk et al.~\cite{martsenyuk2025rnn} for exponential stability results in this framework). Our usage of time scale is different and strictly dynamical: it refers to the characteristic temporal scales of the state and parameter dynamics induced by gating, and to the effective learning rates that couple them.

\section{RNNs and time-scales}
\label{sec:time_scales}

A summary of the main symbols used throughout the paper is provided in the Supplementary Material (Table~\ref{tab:notation}), which readers may consult as a quick-access reference while following the derivations.

We begin with a simple continuous-time RNN model~\cite{tallec2018can}:
\begin{equation}
\label{eq:rnn_ct}
\frac{dx(t)}{dt} = \phi\!\left(W^{r}x(t) + W^i u(t)\right) - x(t), 
\quad x(0) = x_0,
\end{equation}
where the state vector $x(t) \in \mathbb{R}^{N_r}$ evolves under the recurrent weights $W^r \in \mathbb{R}^{N_r \times N_r}$, input weights $W^i \in \mathbb{R}^{N_r \times N_i}$, and elementwise nonlinearity $\phi(\cdot)$.  

The output is generated via a readout mapping
\begin{equation}
\label{eq:output_ct}
z(t) = \psi\!\left(x(t)\right),
\end{equation}
which, in the common linear case, reduces to
\begin{equation}
\label{eq:output_linear_ct}
z(t) = W^o x(t), \quad W^o \in \mathbb{R}^{N_o \times N_r}.
\end{equation}

\subsection{From continuous to discrete time}

Applying a first-order Taylor expansion around $t$ (i.e., Euler discretization) to~\eqref{eq:rnn_ct} gives
\begin{equation}
\label{eq:taylor}
x(t+\delta t) \approx x(t) + \delta t \,\frac{dx(t)}{dt}.
\end{equation}
With unit step $\delta t=1$, this yields the standard discrete-time RNN update
\begin{equation}
\label{eq:rnn_dt}
x_{t+1} = \phi\!\left(W^{r}x_t + W^i u_t\right).
\end{equation}

\subsection{Global time rescaling}

Suppose we want to model a time-rescaled input $u(\alpha t)$ with~\eqref{eq:rnn_ct}.  
Let the state trajectory be reparameterized as $x(\alpha t)$ via a linear time-warping function
\begin{equation}
c(t) = \alpha t, \quad \alpha > 0.
\end{equation}
Rewriting \eqref{eq:rnn_ct} with the new time variable, $\tau=c(t)=\alpha t$, we obtain:
\begin{equation}
\label{eq:rnn_ct_leaky}
\frac{dx(c(t))}{dt} = \frac{dx(\tau)}{d\tau}\frac{d \tau}{dt} = \alpha\phi(W^{r}x(t) + W^iu(t)) - \alpha x(t),
\end{equation}
where we relabel the time variable back to $t$ on the right-hand side to simplify the notation.  
This expression shows that a time-warping $c(t)=\alpha t$ scales both the recurrent and decay terms by the same factor $\alpha$.  
Therefore, the original dynamics in \eqref{eq:rnn_ct} are \emph{not} invariant under time-rescaling: the evolution is accelerated for $\alpha > 1$ and slowed down for $\alpha < 1$.  
In this sense, $\alpha$ acts as a global update rate, with $\alpha \to 0$ yielding nearly frozen dynamics and $\alpha \to \infty$ corresponding to very fast updates.

Discretizing~\eqref{eq:rnn_ct_leaky} with~\eqref{eq:taylor} and $\delta t = 1$ yields
\begin{equation}
\label{eq:rnn_dt_leaky}
x_{t+1} = \alpha \,\phi\!\left(W^{r}x_t + W^i u_t\right) + (1-\alpha) x_t,
\end{equation}
which corresponds to \emph{leaky integrator} neurons~\cite{jaeger2007optimization}.  
Here $\alpha$ is the \emph{global state-update rate}; its reciprocal $1/\alpha \in [1, \infty)$ sets the associated global discrete-time scale, so that rate and time scale act as dual variables.  
When $\alpha \to 0$, state updates occur very slowly, approaching a static memory.

\subsection{General time warping and gating}

To go beyond global rescaling and achieve invariance to more general time transformations, we allow an arbitrary monotonic, differentiable warping $c : \mathbb{R} \to \mathbb{R}$.
\begin{equation}
\begin{aligned}
\frac{dx(c(t))}{dt} 
&= \frac{dc(t)}{dt} \,\phi\!\left(W^{r}x(t) + W^i u(t)\right) - \frac{dc(t)}{dt} \,x(t), \\
\label{eq:rnn_ct_warping}
&= g(t) \,\phi\!\left(W^{r}x(t) + W^i u(t)\right) - g(t) \,x(t),
\end{aligned}
\end{equation}
where $g(t) := \frac{dc(t)}{dt}$ is the instantaneous state-update rate at time $t$.
When $g(t) \equiv \alpha$, we recover the leaky case~\eqref{eq:rnn_ct_leaky}.

We parametrize $g(t)$ as a \emph{gate}:
\begin{equation}
\label{eq:ggate}
g(t) = \sigma\!\left(W^{r, g} x(t) + W^{i, g}u(t)\right),
\end{equation}
where $\sigma(\cdot) \in (0, 1)$ is the logistic sigmoid, $W^{r, g} \in \mathbb{R}^{1 \times N_r}$, and $W^{i, g} \in \mathbb{R}^{1 \times N_i}$ are gate parameters.  
This parametrization ensures that the update rate $g(t)$ remains bounded in $(0,1)$.  
Because $g(t)$ depends both on the current state $x(t)$ and the input $u(t)$, the effective time scale of the network becomes data- and state-dependent.

Discretizing~\eqref{eq:rnn_ct_warping} with $\delta t = 1$ gives
\begin{equation}
\label{eq:rnn_dt_singlegate}
x_{t+1} = g_t \,\phi\!\left(W^{r}x_t + W^i u_t\right) + (1-g_t)\,x_t,
\end{equation}
which makes the dynamics invariant to a global rescaling of time via $g_t$.

For neuron-specific time scales, we assign an individual gate $g_t^{(j)}$ to each neuron $j$:
\begin{equation}
\label{eq:rnn_dt_multigate}
x_{t+1} = g_t \odot \phi\left(W^{r}x_t + W^{i}u_t\right) + (1-g_t) \odot x_t,
\end{equation}
where $\odot$ denotes elementwise multiplication and $g_t \in (0,1)^{N_r}$.
Here $\sigma(\cdot)$ in~\eqref{eq:ggate} is applied componentwise, with $W^{r,g} \in \mathbb{R}^{N_r \times N_r}$ and $W^{i,g} \in \mathbb{R}^{N_r \times N_i}$.  
Each neuron thus possesses its own update rate (or time scale), enabling fine-grained adaptation of the network’s temporal dynamics.

\section{BPTT-based training of RNNs}
\label{sec:bptt_ve_problem}

\subsection{Stochastic gradient descent in neural networks}
\label{sec:bptt}

The most basic form of the \emph{stochastic gradient descent} (SGD) update rule for a parameter vector~$\theta$ is
\begin{equation}
\label{eq:gd}
\theta_{l+1} \;=\; \theta_{l} \;-\; \mu \,\nabla_{\!\theta}\,\mathcal{E}(\theta_l),
\end{equation}
where $l$ indexes the training iteration, $\mu>0$ is the learning rate (or step size), and $\mathcal{E}$ is the loss function.
For a single sequence of length~$T$, the gradient can be expanded using the chain rule.
For recurrent models, the dependence of the current state $x_t$ on all previous states $\{x_k\}_{k < t}$ must be taken into account:
\begin{equation}
\label{eq:gradient_t_full}
\frac{\partial \mathcal{E}}{\partial \theta} 
\;=\; 
\sum_{t=1}^{T} 
\frac{\partial \mathcal{E}_t}{\partial x_t} 
\sum_{k=1}^{t} 
\frac{\partial x_t}{\partial x_k} 
\frac{\partial x_k}{\partial \theta}.
\end{equation}
Here, $\mathcal{E} = \sum_{t=1}^{T} \mathcal{E}_t$ decomposes over time steps, and the term $\frac{\partial x_t}{\partial x_k}$ is the product of Jacobian factors linking the state at time $t$ to the state at time $k$.

Because the gradients must be propagated backward through the sequence of states, this procedure is known as \emph{backpropagation}. When applied to recurrent neural networks, where parameters are shared across time steps, the method is referred to as \emph{backpropagation through time} (BPTT)~\cite{werbos1990bptt}.

In practice, gradients are averaged over a mini-batch of independent sequences, leading to noisy gradient estimates; since this averaging does not affect the per-sequence dynamics, we present the derivation for a single sequence without loss of generality.

\subsection{Vanishing and exploding gradients}
\label{eq:vanishing-exploding_grad}
A central difficulty in computing~\eqref{eq:gradient_t_full} arises from the long Jacobian products
\begin{align}
\label{eq:product_jacobian}
\frac{\partial x_t}{\partial x_k} = \prod_{j=k+1}^{t} J_j, \qquad J_j = \frac{\partial x_j}{\partial x_{j-1}},
\end{align}
whose norm governs the magnitude of the gradient contribution from time~$k$:
\begin{equation}
\label{eq:error_gradient_norm}
\left\lVert \frac{\partial \mathcal{E}_t}{\partial \theta_l} \right\rVert
\;\leq\;
\left\lVert \frac{\partial \mathcal{E}_t}{\partial x_t} \right\rVert
\prod_{j=k+1}^{t} \left\lVert J_j \right\rVert.
\end{equation}
For the standard RNN~\eqref{eq:rnn_dt}, $J_j = D_{j-1} W^r$ with $D_{j} = \mathrm{diag}(\phi^{\prime}(a_{j}))$, so the product contracts or expands exponentially with $(t-k)$ depending on whether $\lVert D_{j-1} \rVert \lVert W^r \rVert$ is persistently below or above~$1$; the classical vanishing/exploding gradient problem~\cite{pascanu2013difficulty}.

\section{How time-scales in RNN states affect parameter dynamics}
\label{sec:rnn-models-affect-learning-dynamics}

In Section~\ref{sec:jacobian_matrices}, we derive the exact expressions for the Jacobian matrices associated with the RNN variants introduced in Section~\ref{sec:time_scales}.  
These Jacobians govern how perturbations to the hidden state propagate through time, and thus play a central role in both forward signal evolution and backward gradient flow.  
Building on these results, Section~\ref{sec:time_scale_interaction} analyzes how the time-scales embedded in the state-space dynamics interact with the optimization process.  
In particular, we show how the presence of constant, scalar, or multiple gating mechanisms influences the effective learning rate and the direction of parameter updates, thereby shaping the behaviour of the gradient descent algorithm.

\subsection{Jacobian matrices}
\label{sec:jacobian_matrices}

\subsubsection{Leaky-integrator neurons}

For the leaky-integrator model \eqref{eq:rnn_dt_leaky}, the Jacobian at time $j$ is
\begin{equation}
\label{eq:jacobian_rnn_dt_leaky}
J_j = \alpha D_{j-1}W^r + (1-\alpha)I,
\end{equation}
where $I$ is the identity matrix and $D_{j-1} = \mathrm{diag}(\phi'(a_{j-1}))$.  
Relative to the standard RNN Jacobian $D_{j-1}W^r$, the constant gate $\alpha$ scales the recurrent contribution while adding a skip connection proportional to $(1-\alpha)I$, thereby tempering gradient decay or explosion.

\subsubsection{Single scalar gate}

Consider now the scalar-gated model \eqref{eq:rnn_dt_singlegate}.  
Defining the gate pre-activation $a^g_t = W^{r,g} x_t + W^{i,g}u_t,\ a^g_t \in \mathbb{R}$, the Jacobian becomes
\begin{equation}
\begin{aligned}
J_j &= \Big[ \phi(a_{j-1})\frac{\partial g_{j-1}}{\partial x_{j-1}} + g_{j-1} D_{j-1}W^r \Big] \\
&+ \Big[ x_{j-1}\frac{\partial(1-g_{j-1})}{\partial x_{j-1}} + (1-g_{j-1})I \Big] \\
&= \big( \phi(a_{j-1}) - x_{j-1} \big) J^{g}_{j-1} + g_{j-1}D_{j-1}W^r + (1-g_{j-1})I \\
\label{eq:jacobian_rnn_dt_warping}
&= G_{j-1} + g_{j-1}D_{j-1}W^r + (1-g_{j-1})I,
\end{aligned}
\end{equation}
where
\begin{equation}
\begin{aligned}
J^{g}_{j} = \frac{\partial g_{j}}{\partial x_{j}} = \sigma'(a^g_j) W^{r,g}, 
\qquad G_{j-1} = d_{j-1} J^{g}_{j-1}, \\
\quad d_{j-1} = \phi(a_{j-1}) - x_{j-1}.
\end{aligned}
\end{equation}
Here $J^g_j \in \mathbb{R}^{1\times N_r}$ and $G_{j-1}$ is rank-1, being the outer product of $d_{j-1}$ and $J^{g}_{j-1}$.  
Thus, the scalar gate introduces a low-rank correction $G_{j-1}$ in addition to rescaling the recurrent and skip terms.

\subsubsection{Multiple gates}

For the multi-gated model \eqref{eq:rnn_dt_multigate}, the gate pre-activation is a vector $a^g_j \in \mathbb{R}^{N_r}$, yielding the gate Jacobian
\begin{equation}
\label{eq:jacobian_gate_multiple}
J^{g}_{j} = \frac{\partial g_j}{\partial x_j} 
          = \mathrm{diag}(\sigma'(a^g_j)) W^{r,g}
          = D^g_j W^{r,g},
\end{equation}
with $D^g_j = \mathrm{diag}(\sigma'(a^g_j))$.  
In this case, $g_j \in \mathbb{R}^{N_r}$ is a vector of neuron-wise update rates.

Introducing the shorthand $\hat{d} = \mathrm{diag}(d)$ for a diagonal matrix with vector $d$ on the diagonal, and $\Gamma = \mathrm{diag}(g)$, the Jacobian of the multi-gate RNN model reads
\begin{equation}
\begin{aligned}
J_j &= \hat{\phi}(a_{j-1}) J^{g}_{j-1} + \Gamma_{j-1} D_{j-1}W^r - \hat{x}_{j-1} J^{g}_{j-1} + (I - \Gamma_{j-1}) \\
&= \hat{d}_{j-1} J^{g}_{j-1} + \Gamma_{j-1} D_{j-1}W^r + (I - \Gamma_{j-1}) \\
\label{eq:jacobian_rnn_dt_multiple_warping}
&= G_{j-1} + \Gamma_{j-1}D_{j-1}W^r + (I - \Gamma_{j-1}).
\end{aligned}
\end{equation}
Unlike the scalar case, here $G_{j-1}$ is not rank-deficient, since it results from multiplying a diagonal matrix by a full matrix.

\subsection{Time-scale coupling}
\label{sec:time_scale_interaction}

In this section, we show that the interaction between state-space time scales and parameter updates gives rise to an effective learning rate $\mu^{*}$, which generally differs from the nominal learning rate $\mu$ in~\eqref{eq:gd}. The precise form of $\mu^{*}$ depends on the chosen state-space model.
To this end, without loss of generality we consider only one time-step $t$ and expand the gradient formula in \eqref{eq:gradient_t_full} as follows:
\begin{align}
\frac{\partial\mathcal{E}_t}{\partial \theta_l} = \frac{\partial \mathcal{E}_t}{\partial x_t} \sum_{k=1}^{t} \left[ \prod_{j = k+1}^{t} J_{j} \right] \frac{\partial x_k}{\partial \theta_l}.
\end{align}

\subsubsection{Constant gate case}
\label{sec:effective_lr_constant_gate}

In the constant gate case $g_t \equiv \alpha \in (0,1]$, the BPTT gradient can be written as
\begin{equation}
\frac{\partial E_t}{\partial \theta_l}
=
\frac{\partial E_t}{\partial x_t}
\sum_{k=1}^{t}
\left[
\prod_{j=k+1}^t \big(I + \alpha\,A_{j-1}\big)
\right]
\frac{\partial x_k}{\partial \theta_l}.
\end{equation}
where the Jacobian \eqref{eq:jacobian_rnn_dt_leaky} is $J_j \;=\; I + \alpha\,A_{j-1}, \ A_{j-1} := D_{j-1} W_r - I$.
Factoring $\alpha$ out of each Jacobian in the product gives
\begin{align}
&\prod_{j=k+1}^{t} \big(I + \alpha\,A_{j-1}\big) \\
&\;=\; \left(\prod_{j=k+1}^{t}\alpha\right)
\; \prod_{j=k+1}^{t} \big(\alpha^{-1} I + A_{j-1}\big)
\;=\; \alpha^{\,t-k} \;\mathcal{P}_{t,k},
\end{align}
where we defined the normalized product $\mathcal{P}_{t,k} \;:=\; \prod_{j=k+1}^{t} \big(\alpha^{-1} I + A_{j-1}\big)$.

Substituting back, the gradient becomes
\begin{equation}
\frac{\partial E_t}{\partial \theta_l}
=
\frac{\partial E_t}{\partial x_t}
\sum_{k=1}^{t}
\alpha^{\,t-k} \;
\mathcal{P}_{t,k} \;
\frac{\partial x_k}{\partial \theta_l}.
\end{equation}
The factor $\alpha^{t-k}$ is the exact multiplicative decay applied to gradient components that travel through $(t - k)$ recurrent steps.
This means that gating induces a lag-dependent reweighting of gradient contributions, equivalent to a lag-dependent effective learning rate:
\begin{equation}
\label{eq:effective_lr_alpha}
\mu^{*}_{t,k} \;=\; \mu\,\alpha^{\,t-k} \;+\; \text{(perturbative correction terms)}.
\end{equation}
Here the perturbative correction terms stem from the residual $(1-\alpha)I$ contribution and from activation derivatives, which slightly distort the pure exponential decay. This makes explicit that the learning rate decays exponentially with temporal distance, at a rate mainly determined by $\alpha$. For $\alpha < 1$, long-range dependencies are progressively downweighted, leading to a vanishing gradient phenomenon for long temporal distances.

\subsubsection{RNNs with scalar gate}

When considering the Jacobian in \eqref{eq:jacobian_rnn_dt_warping}, the product of Jacobians inside the gradient update formula can be rewritten as:
\begin{equation}
\begin{aligned}
\label{eq:jacobian_product_scalar_gate}
&\prod_{j = k+1}^{t} G_{j-1} + g_{j-1}D_{j-1}W^r + (1 - g_{j-1})I \\
&= \prod_{j = k+1}^{t} G_{j-1} + I + g_{j-1}(D_{j-1}W^r - I).
\end{aligned}
\end{equation}
The gradient equation becomes:
\begin{equation}
\label{eq:gradient_t_rnn_dt_warping}
\begin{aligned}
&\frac{\partial\mathcal{E}_t}{\partial \theta_l} \\
&= \frac{\partial \mathcal{E}_t}{\partial x_t} \sum_{k=1}^{t} \left[ \prod_{j = k+1}^{t} G_{j-1} + I + g_{j-1}(D_{j-1}W^r - I) \right] \frac{\partial x_k}{\partial \theta_l}.
\end{aligned}
\end{equation}

In what follows, and in Section~\ref{sec:effective_lr_multiple_gates}, we use the general first-order expansion from Appendix~\ref{sec:matrix_product_expansion}, applied to products of Jacobian matrices with $\epsilon=1$. 
Note that in the present setting the products correspond to BPTT Jacobian chains 
$\prod_{j=k+1}^{t} J_j$, which propagate sensitivities backward through time. 
Consequently, perturbation insertions appear between factors indexed by earlier 
and later time steps, but the ordering remains fully consistent with the general 
matrix product expansion derived in Appendix~\ref{sec:matrix_product_expansion}.

Unlike the constant--$\alpha$ case, here the gate contributions cannot be factored out in closed form; the product involves time-varying, input-driven gates that interact with state dynamics.
We rewrite each factor inside the product as
\begin{equation}
M_{j-1} = I + g_{j-1} A_{j-1} + G_{j-1},
\end{equation}
where $A_{j-1} = D_{j-1} W^r - I$ and $G_{j-1}$ is a rank-1 matrix.

Expanding the matrix product \eqref{eq:jacobian_product_scalar_gate} to the first order in the rank-1 corrections $G_{j-1}$, we obtain
\begin{equation}
\begin{aligned}
\label{eq:product-expansion_single_gate}
\prod_{j=k+1}^{t} M_{j-1} &\approx 
\prod_{j=k+1}^{t} \big( I + g_{j-1} A_{j-1} \big) \\
+ \sum_{m=k+1}^{t} &\Bigg( \prod_{j=m+1}^{t} (I + g_{j-1} A_{j-1}) \Bigg) \\
&G_{m-1} \Bigg( \prod_{j=k+1}^{m-1} (I + g_{j-1} A_{j-1}) \Bigg).
\end{aligned}
\end{equation}

The first term describes the main dynamics without the rank-1 updates, while the second term is a sum of rank-1 perturbative corrections, each inserted at a different time step $m$ and propagated by the main dynamics.
These corrections remain small in practice because they arise from gate activation derivatives, whose range is inherently bounded by the saturating nature of the gating nonlinearity.
We now factor the gate scalars $g_{j-1}$ explicitly from the first term in \eqref{eq:product-expansion_single_gate} like we did in Section~\ref{sec:effective_lr_constant_gate}.
\begin{equation}
\begin{aligned}
&\prod_{j=k+1}^{t} (I + g_{j-1} A_{j-1}) = \\
&\Bigg( \prod_{j=k+1}^{t} g_{j-1} \Bigg)
\Bigg( \prod_{j=k+1}^{t} \big( g_{j-1}^{-1} I + A_{j-1} \big) \Bigg).
\end{aligned}
\end{equation}

Thus, the gate values contribute as a multiplicative attenuation factor given by the product $\prod_{j=k+1}^{t} g_{j-1}$, while the remaining matrices describe the dynamics normalized by $g_{j-1}$.
Similarly, the rank-1 corrections remain rank-1 but acquire products of gates from the intervals before and after the insertion of $G_{m-1}$.

Collecting all terms, we can write
\begin{equation}
\begin{aligned}
\label{eq:single_gate-product-decomposition}
&\prod_{j=k+1}^{t} M_{j-1} \approx
\Bigg( \prod_{j=k+1}^{t} g_{j-1} \Bigg) \mathcal{P}_{t,k} \\
&+ \sum_{m=k+1}^{t} \Bigg( \prod_{j=m+1}^{t} g_{j-1} \Bigg) R_{t,m} G_{m-1} \, \Bigg( \prod_{j=k+1}^{m-1} g_{j-1} \Bigg) S_{k,m},
\end{aligned}
\end{equation}
where $\mathcal{P}_{t,k} = \prod_{j=k+1}^{t} \big( g_{j-1}^{-1} I + A_{j-1} \big)$ and the normalized matrices are defined as $R_{t,m} = \prod_{j=m+1}^{t} \big( g_{j-1}^{-1} I + A_{j-1} \big)$ and $S_{k,m} = \prod_{j=k+1}^{m-1} \big( g_{j-1}^{-1} I + A_{j-1} \big)$.

Plugging \eqref{eq:single_gate-product-decomposition} into the gradient expression, we obtain
\begin{equation}
\begin{aligned}
\label{eq:gradient-gated-decomposition}
\frac{\partial E_t}{\partial \theta_l} &= \frac{\partial E_t}{\partial x_t} \sum_{k=1}^{t} \left( \prod_{j=k+1}^{t} g_{j-1} \right) \mathcal{P}_{t,k} \\
&+ \sum_{m=k+1}^{t} \left( \prod_{j=m+1}^{t} g_{j-1} \right) R_{t,m} \\
& G_{m-1} \, \left( \prod_{j=k+1}^{m-1} g_{j-1} \right) S_{k,m} \frac{\partial x_k}{\partial \theta_l}.
\end{aligned}
\end{equation}

Unlike the constant gate case, here the gates $g_{j-1}$ are time-varying and driven by the input and state. Therefore, the attenuation factor is not a simple power (as with $\alpha$), but a product of gate values along the time interval $[k+1, t]$. This product leads to the following effective learning rate:
\begin{equation}
\label{eq:effective_lr_g}
\mu^{*}_{t,k} \;=\; \mu \prod_{j=k+1}^{t} g_{j-1} \;+\; \text{(perturbative correction terms)}.
\end{equation}
Here the perturbative correction terms arise from the residual $(1-g_t)x_t$ pathway and from gate-gradient contributions. These introduce rank-1 modifications to the Jacobian products, modulating gradient propagation in a way that cannot be collapsed into a single scalar factor.

\subsubsection{RNNs with multiple gates}
\label{sec:effective_lr_multiple_gates}

When considering the Jacobian in \eqref{eq:jacobian_rnn_dt_multiple_warping} the gradient equation becomes:
\begin{equation}
\begin{aligned}
\label{eq:gradient_t_rnn_dt_warping_multiple_gates}
&\frac{\partial\mathcal{E}_t}{\partial \theta_l} 
= \frac{\partial \mathcal{E}_t}{\partial x_t} \\
&\sum_{k=1}^{t}\left[ \prod_{j = k+1}^{t}G_{j-1} + \left( \Gamma_{j-1}D_{j-1}W^r \right) + \left( I - \Gamma_{j-1} \right) \right] \frac{\partial x_k}{\partial \theta_l}.
\end{aligned}
\end{equation}
Each factor in the product appearing in the gradient expression reads
\begin{equation}
M_{j-1} = (I - \Gamma_{j-1}) + \Gamma_{j-1} A_{j-1} + G_{j-1},
\end{equation}
where $A_{j-1} = D_{j-1} W^r$, $\Gamma_{j-1}$ is the diagonal matrix of neuron-specific gates and $G_{j-1} = \hat{d}_{j-1} J^g_{j-1}$ is a full-rank matrix in this case.

Unlike the scalar gate case, here $\Gamma_{j-1}$ does not commute with $A_{j-1}$, and $G_{j-1}$ is no longer rank-1. Therefore, the factorization is more involved. We can rewrite in terms of perturbation as follows:
\begin{equation}
M_{j-1} = \Gamma_{j-1} \big( \tilde{A}_{j-1} \big) + G_{j-1},
\end{equation}
where $\tilde{A}_{j-1} = A_{j-1} + \Gamma_{j-1}^{-1} (I - \Gamma_{j-1})$,
with $\Gamma_{j-1}^{-1}(I - \Gamma_{j-1})$ defined element-wise as $\mathrm{diag}((1 - g^{(i)}_{j-1}) / g^{(i)}_{j-1})$ for $g^{(i)}_{j-1} > 0$.

Similarly to the scalar gate case, we expand the product to first order in $G_{j-1}$:
\begin{equation}
\begin{aligned}
\label{eq:product-expansion-multigate}
&\prod_{j=k+1}^{t} M_{j-1} \approx 
\prod_{j=k+1}^{t} \big( \Gamma_{j-1} \tilde{A}_{j-1} \big) \\
&+ \sum_{m=k+1}^{t} \Bigg( \prod_{j=m+1}^{t} \Gamma_{j-1} \tilde{A}_{j-1} \Bigg)
G_{m-1}
\Bigg( \prod_{j=k+1}^{m-1} \Gamma_{j-1} \tilde{A}_{j-1} \Bigg).
\end{aligned}
\end{equation}

The first term describes the main dynamics through diagonal gates and normalized matrices $\tilde{A}_{j-1}$. The second term accounts for corrections due to $G_{j-1}$.

We now separate the diagonal gate products explicitly:
\begin{equation}
\prod_{j=k+1}^{t} \big( \Gamma_{j-1} \tilde{A}_{j-1} \big) =
\Bigg( \prod_{j=k+1}^{t} \Gamma_{j-1} \Bigg)
\mathcal{P}_{t,k},
\end{equation}
where $\mathcal{P}_{t,k} = \prod_{j=k+1}^{t} \tilde{A}_{j-1}$,
and the product $\prod_{j=k+1}^{t} \Gamma_{j-1}$ is a diagonal matrix with entries $\left( \prod_{j=k+1}^{t} g^{(1)}_{j-1}, \ldots, \prod_{j=k+1}^{t} g^{(N_r)}_{j-1} \right)$.

For the rank corrections, we define
\[
R_{t,m} = \prod_{j=m+1}^{t} \tilde{A}_{j-1}, \qquad
S_{k,m} = \prod_{j=k+1}^{m-1} \tilde{A}_{j-1}.
\]

The expansion to the first order becomes
\begin{equation}
\label{eq:gate-product-multigate}
\begin{aligned}
&\prod_{j=k+1}^{t} M_{j-1} \approx
\Bigg( \prod_{j=k+1}^{t} \Gamma_{j-1} \Bigg) \mathcal{P}_{t,k} \\
&+ \sum_{m=k+1}^{t}
\Bigg( \prod_{j=m+1}^{t} \Gamma_{j-1} \Bigg)
R_{t,m}  G_{m-1} \,
\Bigg( \prod_{j=k+1}^{m-1} \Gamma_{j-1} \Bigg)
S_{k,m}.
\end{aligned}
\end{equation}

Plugging \eqref{eq:gate-product-multigate} into the gradient formula, we obtain
\begin{equation}
\label{eq:gradient-multigate}
\begin{aligned}
\frac{\partial E_t}{\partial \theta_l} &=
\frac{\partial E_t}{\partial x_t}
\sum_{k=1}^{t}
\left( \prod_{j=k+1}^{t} \Gamma_{j-1} \right) \mathcal{P}_{t,k} \\
&+ \sum_{m=k+1}^{t} \left( \prod_{j=m+1}^{t} \Gamma_{j-1} \right) R_{t,m} \\
&G_{m-1} \, \left( \prod_{j=k+1}^{m-1} \Gamma_{j-1} \right) 
S_{k,m} \frac{\partial x_k}{\partial \theta_l}.
\end{aligned}
\end{equation}

Unlike the scalar gate case, here the gate contributions $\prod_{j=k+1}^{t} \Gamma_{j-1}$ form a \emph{diagonal matrix} rather than a scalar. Consequently, the attenuation of the gradient is \emph{neuron-specific} and depends on the entire input-driven gate trajectory:
\begin{equation}
\label{eq:effective_lr_multi_g}
\mu^{*(i)}_{t,k} \;=\; \mu \prod_{j=k+1}^{t} g^{(i)}_{j-1} \;+\; \text{(perturbative correction terms)} ,
\end{equation}
where $\mu^{*(i)}_{t,k}$ is the effective learning rate associated with neuron $i$. 

Here the perturbative correction terms involve interactions between gates and state updates, producing full-rank contributions that cannot be simplified into a scalar or diagonal factor.
In the multi-gate case: (i) each neuron has its own time-varying effective learning rate determined by its specific gate trajectory; (ii) gradient propagation is substantially anisotropic: some directions may be damped strongly, while others may be preserved, depending on the distribution of gate trajectories across neurons; (iii) the correction terms act as neuron-coupling perturbations, making the effective scaling both magnitude- and direction-dependent.

\section{Analogies between gating mechanisms and adaptive gradient descent methods}
\label{sec:gradient_descent_algorithms}

We now turn to well-known variations of the basic gradient descent algorithm~\cite{ruder2016overview}.
Our goal is to draw parallels between these algorithmic modifications and the implicit effects induced by tunable time-scales in RNN state dynamics. A summary is shown in Table~\ref{tab:gates-vs-optimizers}.
In particular, we examine how gates--by shaping the Jacobian structure--alter the effective optimization behaviour of plain gradient descent, in ways reminiscent of learning rate schedules, momentum terms, or adaptive update rules.

The analysis in Sections~\ref{sec:effective_lr_constant_gate}--\ref{sec:effective_lr_multiple_gates} shows that the Jacobian factors in gated RNNs naturally lead to multiplicative modulation of the backpropagated gradient.  
From the first-order expansion, this modulation directly translates into an effective learning rate that depends on the gate configuration. 

\paragraph{Constant gate ($\alpha$)}
In the constant-gate (leaky integrator) case, the gate value $\alpha \in (0,1]$ is fixed throughout training and across all units.  
The dominant multiplicative factor in the backward product over $(t-k)$ steps is $\alpha^{\,t-k}$, independent of the input sequence or hidden state trajectory.  
Unlike the time-varying or multi-gate cases, there is no data dependence: the scaling depends solely on $(t-k)$ and $\alpha$, remaining fixed during training.  
This is analogous to a fixed preconditioning factor in gradient descent, where every parameter update is scaled by the same precomputed value.

\paragraph{Single time-varying gate ($g_{j-1}$)}
With a scalar gate $g_{j-1}$, the gradient is modulated by a global, input-driven attenuation factor that varies over time. This is conceptually similar to a learning rate schedule (e.g., exponential decay), with the key difference that the modulation emerges from the network's own state dynamics rather than from an externally prescribed schedule.

\paragraph{Multiple time-varying gates ($g^{(i)}_{j-1}$)}
In the multi-gate case, the modulation is neuron-specific, so each neuron $i$ has its own effective learning rate, determined by the trajectory of its gate values. This mirrors adaptive optimizers such as Adam or RMSProp, which assign each parameter a distinct, dynamically adjusted step size.

\paragraph{Perturbative corrections and directional modulation}
The perturbative terms arising in the first-order expansion introduce 
directional modifications to the gradient. In the single-gate case, 
the local correction $G_{j-1}$ is rank-1, acting as a low-dimensional 
perturbation akin to momentum. In the multi-gate case, $G_{j-1}$ is 
full-rank, leading to anisotropic scaling reminiscent of the 
preconditioning performed by Adam and other adaptive methods.

These effects are not determined by $G_{j-1}$ alone: the perturbations 
are propagated and mixed across time through products of recurrent 
factors, so that the resulting directional structure reflects the 
combined action of gate-induced corrections and recurrent dynamics.

\begin{table*}[ht!]
\centering
\caption{Conceptual similarities between gating-induced gradient modulation and adaptive gradient descent methods.}
\begin{tabular}{|p{3cm}|p{5cm}|p{5cm}|}
\hline
\textbf{Gating case} & \textbf{Gradient modulation} & \textbf{Analogy with optimizers} \\
\hline
Constant gate ($\alpha$) & Fixed scaling depending on distance $(t-k)$ & SGD with a constant, precomputed scaling factor \\
\hline
Time-varying scalar gate ($g_{j-1}$) & Global, input-driven scaling over time & SGD with a learning rate schedule \\
\hline
Multiple gates ($g^{(i)}_{j-1}$) & Per-neuron (diagonal) scaling & Adam/RMSProp (per-parameter adaptation) \\
\hline
Perturbative corrections & Directional modulation via transported and mixed perturbations (rank-1 or full-rank) & Momentum / adaptive preconditioning \\
\hline
\end{tabular}
\label{tab:gates-vs-optimizers}
\end{table*}

\section{Simulations}
\label{sec:simulations}

\subsection{Effective learning rate induced by gates}
\label{sec:sim1_effective_lr}

We empirically verify that, even when SGD uses a fixed global step size~$\mu$, gating mechanisms induce a lag-dependent effective learning rate that depends on the state-space model.
The central object we use is the lag-conditioned sensitivity, $S_{t,k}\;=\;\left\|\,\mathcal{M}_{t,k}\,\right\|_2, \mathcal{M}_{t,k}=\prod_{j=k+1}^{t} J_j$, whose magnitude reweights how much a gradient signal at time $k$ influences the update due to a loss at time $t$.
At each training checkpoint~$\ell$ we aggregate these quantities to obtain an empirical effective learning rate profile.
For every temporal distance $h = t-k$, we define
\begin{equation}
\tilde{\mu}_{\mathrm{eff}}(h;\ell)\;\propto\;
\mathrm{median}_{h}\,S_{t,k}(\theta_\ell).
\end{equation}
To ensure comparability across models and checkpoints, we report a
normalized version anchored at unit lag:
\begin{equation}
\tilde{\mu}_{\mathrm{eff}}(h;\ell)\;=\;
\frac{\mathrm{median}_{h}\,S_{t,k}(\theta_\ell)}
     {\mathrm{median}_{h=1}\,S_{t,k}(\theta_\ell)}\,.
\end{equation}
This construction emphasizes the relative attenuation with lag:
all curves start from $1$ at $h=1$, so differences directly reflect
how fast or slow the effective learning rate decays with temporal
distance. The normalization removes global scale effects due to parameter
growth and isolates the temporal structure of the effective learning rate.

\paragraph{Predictors and overlays}
To interpret the empirical effective learning rates, we compare them against
a simple \emph{gate-product predictor} derived from the zeroth-order
theory of Section~\ref{sec:time_scale_interaction}. For each model we define
\begin{equation}
P_{t,k}(\theta_\ell)\;=\;
\begin{cases}
\alpha^{\,t-k}, & \text{leaky (constant $\alpha$)},\\[2pt]
\prod_{j=k+1}^{t} g_{j-1}, & \text{scalar gate},\\[6pt]
\frac{1}{N_r}\displaystyle\sum_{i=1}^{N_r}\;\prod_{j=k+1}^{t} g^{(i)}_{j-1}, & \text{multi-gate},
\end{cases}
\end{equation}
which captures the multiplicative attenuation induced purely by the gates.
We then form its lag-binned median
\begin{equation}
\bar P(h;\ell)=\mathrm{median}_{h}\,P_{t,k}(\theta_\ell).
\end{equation}

From this baseline predictor, we construct two reference curves to overlay
on the empirical profiles:
\begin{align}
\tilde{\mu}^{(0)}_{\mathrm{pred}}(h;\ell)
  &= \frac{\bar P(h;\ell)}{\bar P(1;\ell)} ,
  &\text{(zeroth-order theory, no fitting)}\\[6pt]
\tilde{\mu}^{(\mathrm{fit})}_{\mathrm{pred}}(h;\ell)
  &= \Bigg(\frac{\bar P(h;\ell)}{\bar P(1;\ell)}\Bigg)^{s(\ell)} ,
  &\text{(empirical correction)}.
\end{align}

The first curve $\tilde{\mu}^{(0)}_{\mathrm{pred}}$ is the direct theoretical
prediction obtained by normalizing the gate-product decay.
The second curve $\tilde{\mu}^{(\mathrm{fit})}_{\mathrm{pred}}$ introduces
a fitted exponent $s(\ell)$ that corrects for higher-order (perturbative)
effects not captured by the pure gate product. The slope $s(\ell)$ is estimated
at checkpoint $\ell$ via a robust log–log regression
\begin{equation}
\log S_{t,k}\;\approx\;a(\ell)+s(\ell)\,\log P_{t,k},
\end{equation}
restricted to the central predictor range (1\%–99\% quantiles) to avoid extreme-tail artifacts.

\paragraph{Interpretation of the slope}
The slope provides a quantitative diagnostic of how gates renormalize effective learning rates:
\begin{itemize}
  \item $s(\ell)=0$: no lag dependence, i.e.\ the effective learning rate equals the nominal $\mu$ at all lags.
  \item $s(\ell)=1$: exact agreement with the zeroth-order gate-product predictor, i.e.\ $\tilde{\mu}_{\mathrm{eff}}(h)\propto P_{t,k}$.
  \item $s(\ell)>1$: perturbative corrections \emph{amplify} attenuation beyond the predictor, shortening the effective memory horizon.
  \item $0<s(\ell)<1$: perturbative corrections \emph{counteract} attenuation, extending the effective memory horizon.
\end{itemize}

\paragraph{Setup}
We train each RNN model (leaky~\eqref{eq:rnn_dt_leaky}, scalar~\eqref{eq:rnn_dt_singlegate}, multi-gate~\eqref{eq:rnn_dt_multigate}) with plain SGD (no momentum, no adaptivity) on the adding and sinmix tasks. At selected checkpoints, we (i)~run a fixed probe batch, (ii)~build the exact one-step Jacobians $J_j$ from Section~\ref{sec:jacobian_matrices}, (iii)~compute $S_{t,k}$, (iv)~compute $P_{t,k}$, and (v)~produce the normalized lag profiles with theory overlays. We also track $s(\ell)$ across training. All reported quantities are averaged over 20 independent random seeds. Additional details are given in Section~\ref{sec:supp_simulations}.

\paragraph{Results}
Results are shown in Figures~\ref{fig:s1_adding}--\ref{fig:s1_sinmix}. Across both tasks and all models, the zeroth-order overlay $\tilde{\mu}^{(0)}_{\mathrm{pred}}$ captures the correct qualitative lag dependence but not the full magnitude of the attenuation, while the fitted-power overlay $\tilde{\mu}^{(\mathrm{fit})}_{\mathrm{pred}}$ tracks the empirical profile closely.

\emph{Adding task.}
\begin{itemize}
  \item \textbf{Leaky:} $s(\ell)\approx 2.15$, essentially stable throughout training ($R^2\approx 0.998$). The empirical decay is stronger than the $\alpha^h$ baseline, so perturbative corrections increase attenuation.
  \item \textbf{Scalar gate:} $s(\ell)$ decreases from $\approx 0.43$ to $\approx 0.30$ ($R^2\approx 0.994$). Perturbative corrections counteract the gate-product attenuation and extend the effective horizon.
  \item \textbf{Multi-gate:} $s(\ell)$ increases slightly from $\approx 0.55$ to $\approx 0.58$ ($R^2\approx 0.99$). Corrections again counteract attenuation, but less strongly than in the scalar-gated case.
\end{itemize}
The hierarchy $s_{\mathrm{leaky}} \gg s_{\mathrm{multi}} > s_{\mathrm{scalar}}$ remains stable throughout training.

\emph{Sinmix task.}
\begin{itemize}
  \item \textbf{Leaky:} $s(\ell)$ drops from $\approx 2.11$ to $\approx 0.42$ ($R^2$: $1.00 \to 0.72$), crossing from $s>1$ to $s<1$ during training.
  \item \textbf{Scalar gate:} $s(\ell)$ decreases from $\approx 0.36$ to $\approx 0.18$ ($R^2$: $0.97 \to 0.90$).
  \item \textbf{Multi-gate:} $s(\ell)$ decreases from $\approx 0.35$ to $\approx 0.18$ ($R^2$: $1.00 \to 0.92$), becoming nearly indistinguishable from the scalar gate from iteration $400$ onward.
\end{itemize}
Unlike adding, all slopes decrease during training and the inter-model hierarchy compresses, driving all models toward strongly extended memory horizons.

\paragraph{Takeaways}
Across both tasks, $s(\ell)\neq 1$ confirms that perturbative corrections materially reshape the lag profile beyond the zeroth-order predictor. On adding, the sign of the correction is architecture-dependent but stable: it amplifies attenuation in the leaky model and counteracts it in the gated models. On sinmix, training drives all three architectures toward smaller exponents, eventually pushing even the leaky model into the $s<1$ regime.

\subsection{Directional anisotropy: propagation vs.\ updates}
\label{sec:sim2_anisotropy}

Backpropagated signals in recurrent models are transported by
Jacobian products $M_{t,k}$, which determine both how much gradient
survives with lag (magnitude) and how it is distributed across
directions (geometry).
A concentrated singular spectrum of $M_{t,k}$ indicates that
sensitivities align with a low-dimensional subspace in state space.
However, parameter updates follow the realized gradients, which depend
on the data and on how the loss projects onto these directions.
To distinguish these two aspects, we report two complementary
measurements:

\medskip\noindent
\textbf{(A) Propagation anisotropy (from Jacobians).}
For many $(t,k)$ pairs with lag $h=t-k$, we compute the singular values
$\sigma_1\ge\sigma_2\ge\dots$ of $\mathcal{M}_{t,k}$ and summarize them with two
indices.
The \emph{anisotropy index} (AI) measures spectral spread as the ratio
of the largest to the $r$-th singular value, while the
\emph{cumulative energy} (CE) measures the fraction of squared-singular-value
mass that lies in the top-$r$ subspace of the full spectrum:
\begin{align}
\mathrm{AI}_r(\mathcal{M}) &= \operatorname{median}_{h}\!\Big(\tfrac{\sigma_1}{\sigma_r}\Big), \\
\mathrm{CE}_r(\mathcal{M}) &= \operatorname{median}_{h}\!\Big(\tfrac{\sum_{i=1}^{r}\sigma_i^2}{\sum_{i\ge 1}\sigma_i^2}\Big).
\end{align}
Medians are taken across $(t,k)$ pairs with the same lag $h$ to reduce
sensitivity to extreme singular values that arise in long Jacobian
products. Both indices are scale-free, so they can be compared directly across
architectures and lags.
An AI close to~$1$ indicates a nearly isotropic spectrum, while large
values signal that a single direction dominates propagation; CE close
to~$1$ indicates that most of the energy lies in the top-$r$ directions.

\medskip\noindent
\textbf{(B) Parameter update anisotropy (from gradient covariance).}
We collect per-sample gradients into a matrix
$G\in\mathbb{R}^{m\times p}$, where each of the $m$ rows contains the
flattened gradient of all $p$ trainable parameters for one probe
sequence.
To isolate directional geometry, rows are $\ell_2$-normalized,
columns are mean-centered, near-constant columns are removed, and a
small jitter is added for numerical stability.
The singular values of $G$ are proportional to the square roots of the
eigenvalues of the gradient covariance matrix $C=G^\top G$, so spectral
analysis of $G$ reveals the anisotropy of parameter update directions.
If $\sigma_1\ge\sigma_2\ge\dots$ are the singular values of $G$, we
report the same indices:
\begin{align}
\mathrm{AI}_r(G) &= \tfrac{\sigma_1}{\sigma_r}, \qquad
\mathrm{CE}_r(G) &= \tfrac{\sum_{i=1}^{r}\sigma_i^2}{\sum_{i\ge 1}\sigma_i^2}.
\end{align}
These metrics quantify how strongly parameter updates concentrate into
a low-dimensional subspace under the training distribution.

\paragraph{Setup}
We compare three models with the same state size:
(i) plain RNN trained with Adam (no gates), (ii) scalar-gated RNN~\eqref{eq:rnn_dt_singlegate} trained with SGD, and (iii) multi-gated RNN~\eqref{eq:rnn_dt_multigate} trained with SGD. This pairing is by design: plain+Adam isolates optimizer-induced preconditioning, while gated+SGD isolates gating-induced preconditioning, so that the contrast between the two cleanly separates the two sources of update anisotropy. We evaluate on four synthetic sequence-to-scalar tasks at $T=120$ (\emph{adding}, \emph{AR(2)}, \emph{delay-sum}, \emph{moving-average}) and on permuted sequential MNIST (psMNIST, $T=784$) in regression-surrogate form as a canonical long-range real-world benchmark. We stress that these simulations are intended as validation of the theoretical framework, not as a benchmarking study: the quantities reported (lag-resolved AI/CE of Jacobian products and of the gradient covariance, with $r=10$ and $m=256$) characterize recurrent transport and update geometry rather than task-level performance, and the architecture+optimizer pairing is a controlled design rather than a tuning target. All three pairs are trained under an identical protocol: $1200$ iterations, batch size $B=64$, and no early stopping, with no gradient clipping in the primary protocol.
Curves and bars aggregate 20 independent random seeds using medians and interquartile ranges. Repeating the full analysis with a uniform global gradient-norm clip (max norm $1$) produces the same qualitative ordering, confirming that the observed anisotropy is not an artifact of rare unstable gradient steps.
Additional details, including training-loss and gradient-norm diagnostics, are given in the Supplementary Material, Section~\ref{sec:supp_simulations}.

\paragraph{Results}
A consistent picture emerges when comparing propagation anisotropy (from Jacobians) with parameter-update anisotropy (from gradient covariances):
\begin{itemize}
\item \textbf{Jacobians.} AI rises steeply with lag on all tasks, and CE$_{10}$ is generally close to $1$ at long lags. Plain+Adam often exhibits the largest Jacobian AI at long lags (adding, AR(2), delay-sum, moving-average, and psMNIST), largely because the smaller singular values collapse. The main exception to uniformly high CE is scalar gating on adding, where CE$_{10}$ drops sharply at the longest lags. On psMNIST -- whose maximum lag $T=784$ is well beyond the synthetic horizon -- the same trend is observed at lags up to several hundred steps.
\item \textbf{Parameter updates.} Gradient-covariance analysis gives the opposite picture: gated models consistently induce stronger update anisotropy than plain+Adam. At the final checkpoint, CE$_{10}(G)$ is $\approx 0.99$--$1.00$ for gated models versus $\approx 0.92$--$1.00$ for plain+Adam, and AI$_{10}(G)$ is higher for scalar/multi gates than for plain+Adam on every task: adding ($33/30$ vs.\ $24$), AR(2) ($21/21$ vs.\ $6.3$), delay-sum ($421/392$ vs.\ $6.4$), moving-average ($252/299$ vs.\ $21$), and psMNIST ($2280/1880$ vs.\ $22$). The gap is order-of-magnitude on delay-sum, moving-average, and psMNIST; on adding it is narrower but preserved across every seed.
\item \textbf{Scalar vs.\ multi-gate.} Multi-gate usually produces the strongest gradient-covariance anisotropy, with the clearest advantage on moving-average, while scalar gating remains competitive on AR(2) and slightly exceeds multi-gate on delay-sum. This is consistent with the perturbative contributions in Eqs.~\eqref{eq:effective_lr_g} and \eqref{eq:effective_lr_multi_g}: although the leading-order scalar-gate effect is rank-1, the correction terms vary with lag and can still redirect sensitivity across different subspaces.
\end{itemize}

\paragraph{Takeaways}
All models develop low-dimensional propagation at long lags, but Jacobian spectra alone overstate the directional structure of learning. They capture what the dynamics \emph{could} amplify, whereas the gradient covariance captures what optimization \emph{actually} excites. In this sense, Adam can yield highly anisotropic Jacobian transport without comparably concentrated update geometry, while gating reshapes the recurrent dynamics so that realized gradients concentrate into much more coherent low-dimensional parameter subspaces. Across all tasks, gated models induce substantially stronger update anisotropy than plain+Adam. The relative advantage of multi-gate over scalar gating is real but task-dependent: strongest on nonlinear or strongly interacting dynamics, but small or even reversed on specific linear tasks.

\subsection{Broader implications}
\label{sec:sim_broader}
The simulations of Sections~\ref{sec:sim1_effective_lr} and
\ref{sec:sim2_anisotropy} reveal that gating plays two complementary
roles in recurrent learning dynamics: it modulates the magnitude of
effective learning rates across lags
(Section~\ref{sec:sim1_effective_lr}), and it shapes the directional
structure of parameter updates through the geometry of Jacobian products
and gradient covariance (Section~\ref{sec:sim2_anisotropy}).

From this perspective, gating acts as a structural prior on temporal
credit assignment, biasing both the magnitude and geometry of gradient
propagation. The relative advantage of scalar versus multi-gate
architectures is task-dependent rather than following a simple
complexity hierarchy: multi-gate models typically produce stronger
gradient concentration, but scalar gating can match or even exceed this
effect on certain tasks. Taken together, these results suggest that
architectures and optimizers jointly determine not only how fast
parameters move, but also which temporal and directional modes dominate
learning.

\section{Conclusions and future directions}
\label{sec:conclusions}

This work developed a dynamical-systems account of how gating in recurrent neural networks couples state evolution and parameter updates. The central message is that gates do not merely regulate information flow in state space: they reshape the temporal and directional structure of learning itself. In particular, we showed that gating induces lag-dependent effective learning rates and modulates the geometry of gradient transport, thereby coupling state-space time scales with the update dynamics of gradient descent.

Our analysis makes this coupling explicit at two complementary levels. First, exact Jacobian formulas and their first-order expansion reveal how constant, scalar, and multi-dimensional gates control the attenuation of temporal sensitivities, leading to lag-dependent effective learning rates even under plain SGD. Second, the same framework shows that gates also shape the directional structure of learning, determining which subspaces of parameter space are preferentially excited during training.

The simulations support this picture from both sides. Zeroth-order gate products capture the qualitative decay of effective learning rates, while perturbative corrections explain deviations and determine whether the effective horizon is shortened or extended. At the same time, although all models exhibit low-dimensional Jacobian transport at long lags, gated architectures produce significantly more concentrated gradient-covariance structure than plain RNNs trained with Adam. This highlights a key distinction: Jacobian spectra describe what the dynamics could amplify, whereas gradient covariance reveals what optimization actually uses.

Taken together, these results position gating as an architecture-level mechanism for adaptive optimization. Rather than acting as a heuristic analogue of external methods such as learning-rate schedules or Adam, gating embeds temporal geometry directly into the model dynamics. From this viewpoint, architectures and optimizers should be understood as jointly shaping the magnitude and geometry of temporal credit assignment.

This perspective is naturally expressed through the concept of effective learning rates, which emerge here as mesoscopic variables linking state-space and parameter-space dynamics. Training with SGD-like methods can thus be viewed not as a purely parameter-space optimization process, but as a coupled dynamical system in which state evolution and parameter updates jointly determine how information is propagated and learned. This suggests a shift from an optimization-centric view of deep learning toward a dynamical one centered on the interaction between architecture, state, and learning.

Several directions follow from this perspective. Extending the effective-learning-rate framework to richer gated architectures such as LSTMs~\cite{hochreiter1997long} and GRUs~\cite{cho2014learning} is a natural next step. In these models, multiple interacting gates define structured, input-dependent time scales whose combined effect on gradient transport and parameter updates is likely to be substantially richer than in the settings analyzed here.

More broadly, the effective-learning-rate formalism points toward a first-principles theory of learnability. Rather than relying solely on empirical benchmarks, one may aim to predict which temporal dependencies are learnable under a given architecture and training regime, based on the emergent spectrum of effective learning rates governing information propagation.

At the same time, making explicit the role of gating in optimization opens the possibility of designing more principled training algorithms. Instead of treating architecture and optimizer as separate components, one may seek joint designs in which optimization methods explicitly exploit the time-scale structure induced by the network dynamics.

Finally, the underlying mechanism is not specific to recurrent architectures. Because the analysis is based on products of Jacobians and their perturbative structure, it extends naturally to deep feedforward, convolutional, and attention-based models. In feedforward and convolutional networks, temporal depth can be replaced by layer depth, with effective learning rates describing sensitivity propagation across layers. In Transformers~\cite{vaswani2017attention}, attention weights act as input-dependent gating mechanisms, while residual connections and normalization layers contribute additional structured components to the Jacobian. Extending the framework in this direction may provide a unified dynamical perspective on gradient transport and optimization across modern deep learning models.

Overall, these directions suggest that effective learning rates may serve as a unifying set of mesoscopic variables linking architecture, dynamics, and optimization in deep learning.


\bibliographystyle{IEEEtran}

\bibliography{bibliography.bib}

\clearpage
\input{supplementary}

\end{document}

%% file: supplementary.tex
\onecolumn

\section*{Supplementary material}
\setcounter{subsection}{0}

\subsection{Notation summary}
\label{sec:notation}

For reference, Table~\ref{tab:notation} collects the main symbols used throughout the paper and the supplementary material.
Each symbol is introduced at its first appearance in the main text; the table is provided as a quick-access map for readers encountering the framework for the first time.

Throughout the paper we use ``time scale'' as the general term for the characteristic temporal scale over which a quantity of interest evolves: it describes how quickly state, gate, or gradient-propagation effects decay or are integrated in time, and it may itself be either time-varying (as in input- and state-dependent gates) or constant. When the time scale is constant -- for example, the fixed leak rate $\alpha$ in the leaky-integrator RNN -- we also call it a \emph{time constant}. The distinction reflects generality rather than different physical content: every time constant is a time scale, while a time scale need not be a time constant.

In our setting, \emph{rates and time scales are dual variables}: each rate parameter determines an associated temporal scale, and smaller rates correspond to longer time scales. The canonical constant-rate example is the linear decay $\dot x = -\alpha x$, whose solution $x(t) = x(0)e^{-\alpha t}$ identifies the time constant $\tau = 1/\alpha$. The same perspective carries over to discrete time: for $x_{t+1} = \rho x_t$ with $\rho \in (0,1)$, the associated time scale is $\tau = -1/\log \rho$; when $\rho = e^{-\alpha}$, this again gives $\tau = 1/\alpha$. Throughout the paper, we use this duality to move interchangeably between rate parameters and the temporal scales they induce. For a broader treatment of multi-time-scale dynamics, we refer the reader to~\cite{kuehn2015multiple}.

\begin{table}[th!]
\centering
\caption{Summary of the main notation used in the paper.}
\label{tab:notation}
\renewcommand{\arraystretch}{1.2}
\begin{tabular}{@{}l p{0.70\linewidth}@{}}
\hline
Symbol & Meaning \\
\hline
\multicolumn{2}{@{}l}{\emph{Dimensions and indices}} \\
$N_r,\, N_i,\, N_o$ & Recurrent state, input, and output dimensions \\
$T$ & Sequence length \\
$t,\, k,\, j$ & Discrete time indices: $t$ current time, $k$ past time, $j$ generic step \\
$h = t - k$ & Lag between the current step $t$ and a past step $k$ \\
$\ell$ & Training checkpoint (SGD iteration) index \\
\hline
\multicolumn{2}{@{}l}{\emph{States, inputs, losses, parameters}} \\
$x_t \in \mathbb{R}^{N_r}$ & Hidden state at time $t$ \\
$u_t \in \mathbb{R}^{N_i}$ & Input at time $t$ \\
$\mathcal{E}_t,\ \mathcal{E} = \sum_t \mathcal{E}_t$ & Per-step loss and total loss \\
$\theta,\ \mu$ & Trainable parameter vector and nominal (global) learning rate \\
\hline
\multicolumn{2}{@{}l}{\emph{Weights}} \\
$W^r,\, W^i,\, W^o$ & Recurrent, input, and output weight matrices \\
$W^{r,g},\, W^{i,g}$ & Recurrent and input weight matrices of the gate \\
\hline
\multicolumn{2}{@{}l}{\emph{Nonlinearities and derivatives}} \\
$\phi(\cdot),\, \sigma(\cdot)$ & State and gate nonlinearities (generic in the theory; set to $\tanh$ and the logistic sigmoid, respectively, in the experiments) \\
$a_j,\, a^g_j$ & State and gate pre-activations at time $j$ \\
$D_j = \mathrm{diag}(\phi'(a_j))$ & Diagonal of state-nonlinearity derivatives \\
$D^g_j = \mathrm{diag}(\sigma'(a^g_j))$ & Diagonal of gate-nonlinearity derivatives \\
\hline
\multicolumn{2}{@{}l}{\emph{Gates}} \\
$\alpha \in (0, 1]$ & Constant (leaky) gate \\
$g_t$ & Input- and state-dependent gate (scalar or neuron-wise vector) \\
$\Gamma_t = \mathrm{diag}(g_t)$ & Diagonal-matrix form of the multi-gate \\
\hline
\multicolumn{2}{@{}l}{\emph{Per-step Jacobian decomposition (main text)}} \\
$J_j = \partial x_j / \partial x_{j-1}$ & Per-step Jacobian of the recurrent dynamics \\
$A_{j-1}$ & Zeroth-order gated-transport block in $J_j$ \\
$G_{j-1}$ & Gate-induced correction block in $J_j$ \\
\hline
\multicolumn{2}{@{}l}{\emph{Jacobian products, sensitivities, predictor}} \\
$\mathcal{M}_{t,k}=\prod_{j=k+1}^t J_j$ & Iterated Jacobian product from step $k$ to step $t$ \\
$S_{t,k} = \|\mathcal{M}_{t,k}\|_2$ & Operator 2-norm (largest singular value) of $\mathcal{M}_{t,k}$ \\
$\mathcal{P}_{t,k}$ & Normalized matrix product appearing in the gate-product decompositions of the main text \\
$P_{t,k}$ & Zeroth-order scalar gate-product predictor (empirical diagnostic), defined case-wise below: \\
\quad leaky & $P_{t,k} = \alpha^{\,t-k}$ \\
\quad scalar gate & $P_{t,k} = \prod_{j=k+1}^{t} g_{j-1}$ \\
\quad multi-gate & $P_{t,k} = \tfrac{1}{N_r}\sum_{i=1}^{N_r}\prod_{j=k+1}^{t} g^{(i)}_{j-1}$ (per-neuron product averaged across neurons) \\
\hline
\multicolumn{2}{@{}l}{\emph{Effective learning rates}} \\
$\mu^{*}_{t,k}$ & Scalar lag-dependent effective learning rate \\
$\mu^{*(i)}_{t,k}$ & Neuron-specific effective learning rate (multi-gate) \\
\hline
\multicolumn{2}{@{}l}{\emph{Empirical diagnostics}} \\
$s(\ell)$ & Fitted log-log slope of $\log S_{t,k}$ vs.\ $\log P_{t,k}$ at checkpoint $\ell$ \\
$\mathrm{AI}_r,\, \mathrm{CE}_r$ & Anisotropy index and cumulative energy on the top-$r$ singular values (of $M_{t,k}$ or of the gradient covariance) \\
\hline
\multicolumn{2}{@{}l}{\emph{Perturbative expansion (Appendix~\ref{sec:matrix_product_expansion})}} \\
$\varepsilon$ & Perturbation strength \\
$F(\varepsilon)$ & Perturbed matrix product \\
$B_j$ & Generic perturbation matrix used in the abstract expansion $F(\varepsilon) = \prod_j (A_j + \varepsilon B_j)$ \\
$r_j = \|B_j\|_2 / \|A_j\|_2$ & Per-step perturbative ratio (perturbation block relative to the dominant block) \\
$T_1(\varepsilon),\, T_2(\varepsilon)$ & First- and second-order Taylor truncations of $F(\varepsilon)$ \\
$C_2(\varepsilon),\, C_3(\varepsilon)$ & Rescaled remainders $\|F - T_1\|/\varepsilon^2$ and $\|F - T_2\|/\varepsilon^3$ \\
\hline
\end{tabular}
\end{table}

\subsection{Matrix product expansion via Fr\'echet derivatives}
\label{sec:matrix_product_expansion}

We derive here the first-order expansion of a product of matrices with structured perturbations, starting from the \emph{product rule for the Fr\'echet derivative} (Theorem~3.3 in Higham~\cite{higham2008functions}).  
We also show how the same reasoning naturally extends to higher-order terms.

\subsubsection{Fr\'echet differentiability and the first-order expansion}

Let $\mathbb{C}^{n\times n}$ denote the finite-dimensional vector space of complex $n\times n$ matrices equipped with a matrix norm (specifically, the operator 2-norm unless otherwise stated).  
Since the space is finite-dimensional, all norms are equivalent.

\begin{definition}[Fr\'echet differentiability {\cite{krantz2003implicit,higham2008functions}}]
Let $f : \mathbb{C}^{n \times n} \to \mathbb{C}^{n \times n}$.  
We say that $f$ is \emph{Fr\'echet differentiable} at $A \in \mathbb{C}^{n\times n}$ if there exists a bounded linear mapping
\[
L_f(A,\cdot) : \mathbb{C}^{n\times n} \to \mathbb{C}^{n\times n}
\]
such that
\begin{equation}
\lim_{\|E\| \to 0}
\frac{\|f(A+E)-f(A)-L_f(A,E)\|}{\|E\|}=0.
\label{eq:frechet-definition}
\end{equation}
The operator $L_f(A,\cdot)$ is called the \emph{Fr\'echet derivative} of $f$ at $A$, and it is unique if it exists.
\end{definition}

\paragraph{First-order expansion.}

If $f$ is Fr\'echet differentiable at $A$, the first-order Taylor expansion reads
\begin{equation}
f(A+E) = f(A) + L_f(A,E) + o(\|E\|),
\label{eq:first-order-taylor}
\end{equation}
where $E \in \mathbb{C}^{n\times n}$ is the perturbation matrix specifying the direction and structure of the infinitesimal change applied to $A$.  
The notation $o(\|E\|)$ means that $\|o(\|E\|)\|/\|E\| \to 0$ as $\|E\|\to0$.

\paragraph{Product rule.}

If $g$ and $h$ are Fr\'echet differentiable at $A$, their product
\[
f(X)=g(X)h(X)
\]
satisfies the \emph{product rule}
\begin{equation}
L_{gh}(A,E)
=
L_g(A,E)\,h(A)
+
g(A)\,L_h(A,E),
\label{eq:frechet-product-rule}
\end{equation}
where $L_g(A,E)$ denotes the Fr\'echet derivative of $g$ at $A$ in the direction $E$; see~\cite[Sec.~3.2, Thm.~3.3]{higham2008functions}.

\subsubsection{Matrix products with structured perturbations}

Consider a product of $n$ factors, each containing a perturbation proportional to a scalar parameter $\varepsilon$:
\begin{equation}
F(\varepsilon)=\prod_{j=1}^{n}\big(A_j+\varepsilon B_j\big),
\label{eq:product-def}
\end{equation}
where
\begin{itemize}
\item $A_j\in\mathbb{C}^{d\times d}$ is the unperturbed factor at position $j$,
\item $B_j\in\mathbb{C}^{d\times d}$ is the perturbation inserted at position $j$,
\item $\varepsilon\geq0$ controls the magnitude of the perturbations.
\end{itemize}

The perturbation direction corresponds to the tuple
\[
E=(B_1,B_2,\dots,B_n),
\]
meaning that each factor in the product receives its own perturbation.

Since each factor in~\eqref{eq:product-def} is affine in $\varepsilon$, the function $F(\varepsilon)$ is a polynomial in $\varepsilon$ of degree at most $n$.

\subsubsection{Recursive application of the product rule}

For any $k\le n$ define
\[
F_k(\varepsilon)
:=
\prod_{j=1}^{k}(A_j+\varepsilon B_j),
\qquad
F_n(\varepsilon)=F_{n-1}(\varepsilon)(A_n+\varepsilon B_n).
\]

Apply the product rule~\eqref{eq:frechet-product-rule} to $F_n$ with
\[
g(\varepsilon)=F_{n-1}(\varepsilon),
\qquad
h(\varepsilon)=A_n+\varepsilon B_n .
\]

At $\varepsilon=0$ we have
\[
g(0)=F_{n-1}(0)=A_1A_2\cdots A_{n-1},
\qquad
h(0)=A_n,
\qquad
L_h(0,E)=B_n .
\]

The product rule gives
\begin{equation}
L_{F_n}(0,E)
=
L_g(0,E)\,A_n
+
(A_1\cdots A_{n-1})\,B_n .
\label{eq:recursive-step}
\end{equation}

\begin{itemize}
\item The first term propagates the derivative into the earlier factors.
\item The second term inserts the perturbation $B_n$ in the last position.
\end{itemize}

Iterating this recursion yields the general expression
\begin{equation}
\label{eq:first-derivative}
L_{F_n}(0,E)
=
\sum_{i=1}^{n}
\left(\prod_{j=1}^{i-1}A_j\right)
B_i
\left(\prod_{j=i+1}^{n}A_j\right),
\end{equation}
where empty products are interpreted as the identity matrix.

\paragraph{Illustration for $n=3$.}

Write
\[
F_3(\varepsilon)
=(A_1+\varepsilon B_1)
(A_2+\varepsilon B_2)
(A_3+\varepsilon B_3).
\]

Applying the recursive expansion gives
\[
L_{F_3}(0,E)
=
B_1A_2A_3
+
A_1B_2A_3
+
A_1A_2B_3.
\]

Each term contains exactly one perturbation $B_i$ inserted at position $i$.

\subsubsection{First-order expansion of the matrix product}

Applying the Taylor expansion~\eqref{eq:first-order-taylor} to~\eqref{eq:product-def} gives
\begin{equation}
F(\varepsilon)
=
F(0)
+
\varepsilon L_F(0,E)
+
O(\varepsilon^2),
\end{equation}
where $F(0)=\prod_{j=1}^n A_j$ and $L_F(0,E)$ is given by~\eqref{eq:first-derivative}.  
Substituting yields
\begin{equation}
\label{eq:first-order-expansion}
F(\varepsilon)
=
\underbrace{\left(\prod_{j=1}^{n}A_j\right)}_{\text{dominant dynamics}}
+
\underbrace{\varepsilon
\sum_{m=1}^{n}
\left(\prod_{j=1}^{m-1}A_j\right)
B_m
\left(\prod_{j=m+1}^{n}A_j\right)}_{\text{perturbative correction terms}}
+
O(\varepsilon^2).
\end{equation}

\paragraph{Higher-order terms}

The same combinatorial structure extends to higher derivatives.  
Since each factor $(A_j+\varepsilon B_j)$ is affine in $\varepsilon$, the expansion terminates at order $n$.

The $r$-th derivative of $F(\varepsilon)$ at $\varepsilon=0$ is
\begin{equation}
\label{eq:higher-order}
F^{(r)}(0)
=
r!
\!\!\!\sum_{1\le m_1<\cdots<m_r\le n}
\!\!\!
\left(\prod_{j=1}^{m_1-1}A_j\right)
B_{m_1}
\left(\prod_{j=m_1+1}^{m_2-1}A_j\right)
B_{m_2}
\cdots
B_{m_r}
\left(\prod_{j=m_r+1}^{n}A_j\right),
\end{equation}
where empty products denote the identity matrix.

Consequently,
\begin{equation}
\label{eq:taylor_expansion}
F(\varepsilon)
=
\sum_{r=0}^{n}
\frac{\varepsilon^r}{r!}
F^{(r)}(0).
\end{equation}

For $r=1$, Eq.~\eqref{eq:higher-order} reduces to the first derivative
\[
F^{(1)}(0)=L_F(0,E)
\]
appearing in Eq.~\eqref{eq:first-order-expansion}.  
Higher-order terms correspond to simultaneous perturbations inserted at multiple positions in the product.

\subsection{Correspondence between abstract and BPTT indexing}

In the main text, we map the abstract product index $j=1,\dots,n$ in 
Appendix~\ref{sec:matrix_product_expansion} to the BPTT time index 
$j = k+1,\dots,t$, so that the product 
$\prod_{j=1}^{n} A_j$ corresponds to 
$\prod_{j=k+1}^{t} A_{j-1}$, with ordering induced by backward propagation through time. 
Under this identification, the placement of perturbation terms in the product expansion 
matches the general Fr\'echet expansion, with perturbations inserted between factors 
associated with earlier and later time steps.

\subsection{Simulations supporting the validity of the first-order approximation}
\label{sec:simulations_first-order_expansion}

The perturbative analysis relies on the matrix product expansion derived in Appendix~\ref{sec:matrix_product_expansion}.
Specializing that abstract derivation to the gated RNN case, the generic perturbation matrix $B_j$ of Appendix~\ref{sec:matrix_product_expansion} is identified with the gate-induced correction block of the per-step Jacobian, namely $B_j \equiv G_j$ (the factor-index analogue of $G_{j-1}$ in the main text). The Jacobian factors are therefore decomposed as
\[
M_j = A_j + \varepsilon B_j,
\]
where $A_j$ represents the dominant component of the Jacobian and $B_j = G_j$ corresponds to the gate-induced correction.
The product
\[
F(\varepsilon) = \prod_{j=1}^{n}(A_j + \varepsilon B_j)
\]
admits the first-order approximation
\[
T_1(\varepsilon) = F(0) + \varepsilon L_F(0,B),
\]
where $L_F(0,B)$ denotes the Fr\'echet derivative of the product (Appendix~\ref{sec:matrix_product_expansion}).
The approximation is accurate when the perturbative terms $\|B_j\|$ are small relative to $\|A_j\|$, so that the neglected $O(\varepsilon^2)$ contributions remain negligible.

We validate the expansion on a synthetic multi-frequency sinusoidal input signal:
each input channel is a slowly varying sinusoid with a distinct frequency,
plus a small amount of Gaussian noise.
This choice is deliberate: the smooth, bounded nature of the signal keeps
the gate pre-activations in a moderate regime where the sigmoid nonlinearity
is neither saturated nor near-linear, producing gate values that are genuinely
time-varying while remaining well-conditioned.
This ensures that the perturbative matrices $B_j$ are non-trivial---i.e.,
the gates are actively modulating the dynamics---yet remain small relative
to the dominant matrices $A_j$, which is precisely the regime where the
first-order expansion is expected to be accurate.
Using a task-specific training signal (such as the adding problem) would
conflate two effects---the validity of the expansion and the particular
optimization trajectory---whereas the synthetic signal isolates the
geometric question of whether the Jacobian factorization satisfies the
perturbative regime.

To empirically verify that this regime holds for the RNN models considered in the paper, we evaluate two complementary diagnostics.
It is important to note that the first-order approximation is not expected to hold uniformly for arbitrary RNN architectures or tasks, since the relative magnitude of the perturbative terms $B_j$ depends on the operating regime of the network and the statistics of the input sequence. The purpose of the following analysis is therefore not to establish a universal property of gated RNNs, but rather to verify that the perturbative assumptions underlying the theoretical analysis are satisfied in the specific experimental settings studied in this work.

First, for a range of $\varepsilon$ values we measure the truncation error
\[
E_{\mathrm{trunc}}(\varepsilon)
=
\|F(\varepsilon) - T_1(\varepsilon)\|.
\]
If the perturbative expansion is valid, the truncation error should
scale as $O(\varepsilon^2)$.
To visualize this behaviour and to assess whether truncation at first order is numerically well justified, we also report the rescaled remainder
\[
C_2(\varepsilon)
=
\frac{E_{\mathrm{trunc}}(\varepsilon)}{\varepsilon^2},
\]
which should approach a constant as $\varepsilon \to 0$ if the leading neglected term is quadratic.
As a stricter validation, we additionally consider the second-order approximation $T_2(\varepsilon)$ and the corresponding rescaled remainder
\[
C_3(\varepsilon)
=
\frac{\|F(\varepsilon)-T_2(\varepsilon)\|}{\varepsilon^3},
\]
which should likewise approach a constant as $\varepsilon \to 0$ if, after subtraction of the second-order contribution, the residual is dominated by a cubic term.

Second, in the operating regime of the paper ($\varepsilon = 1$),
we evaluate the relative magnitude of the perturbative matrices by
measuring the per-step operator-norm ratio
\[
r_j = \frac{\|B_j\|_2}{\|A_j\|_2}.
\]
Small values of $r_j$ indicate that the perturbative terms remain
substantially smaller than the dominant Jacobian contributions,
ensuring that higher-order corrections remain negligible.

All reported diagnostics are averaged over 20 independent random
seeds in order to mitigate variability due to random initialization
and stochastic training dynamics.

\begin{figure}[t]
\centering
\includegraphics[width=0.45\textwidth]{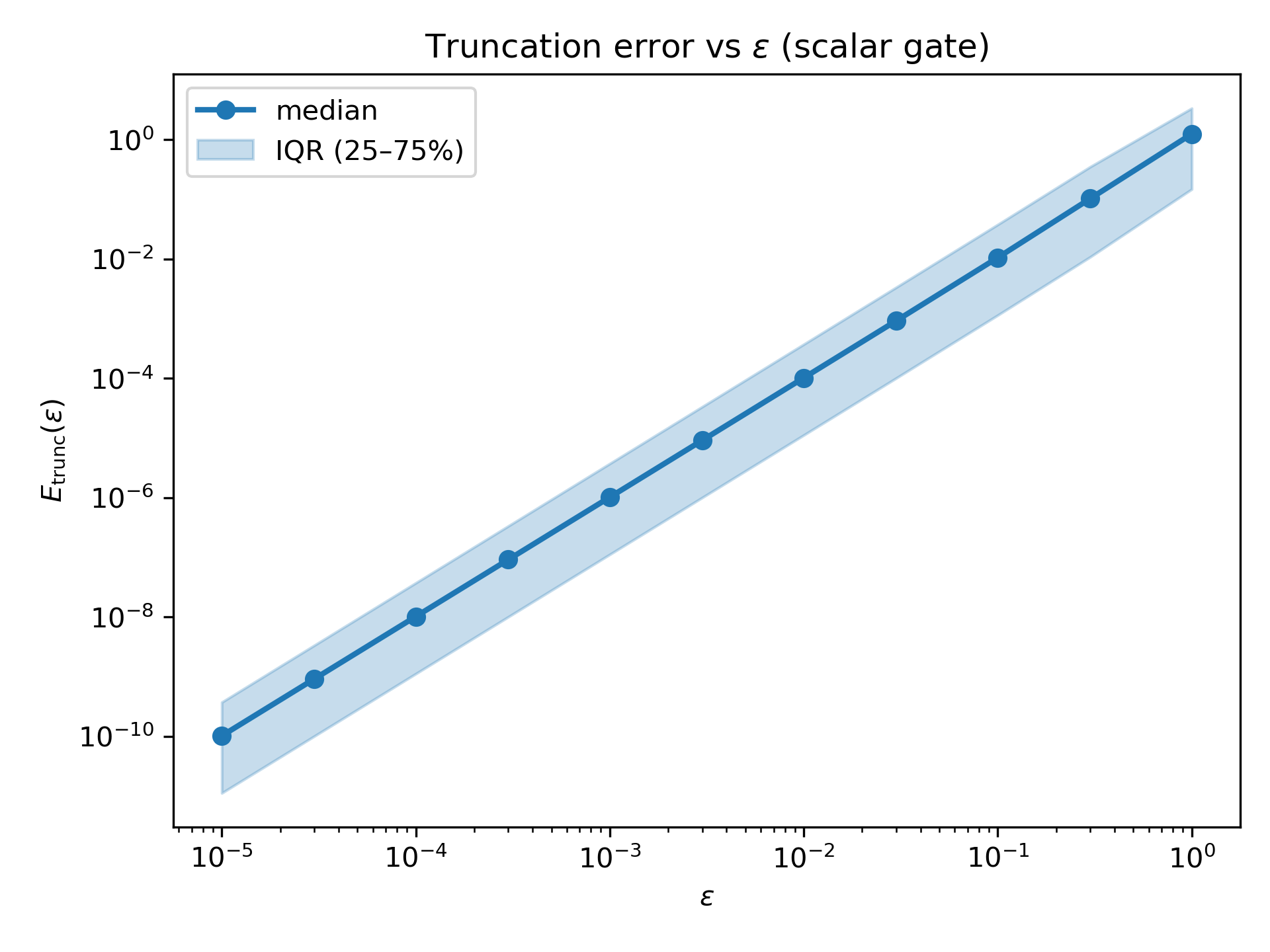}
\includegraphics[width=0.45\textwidth]{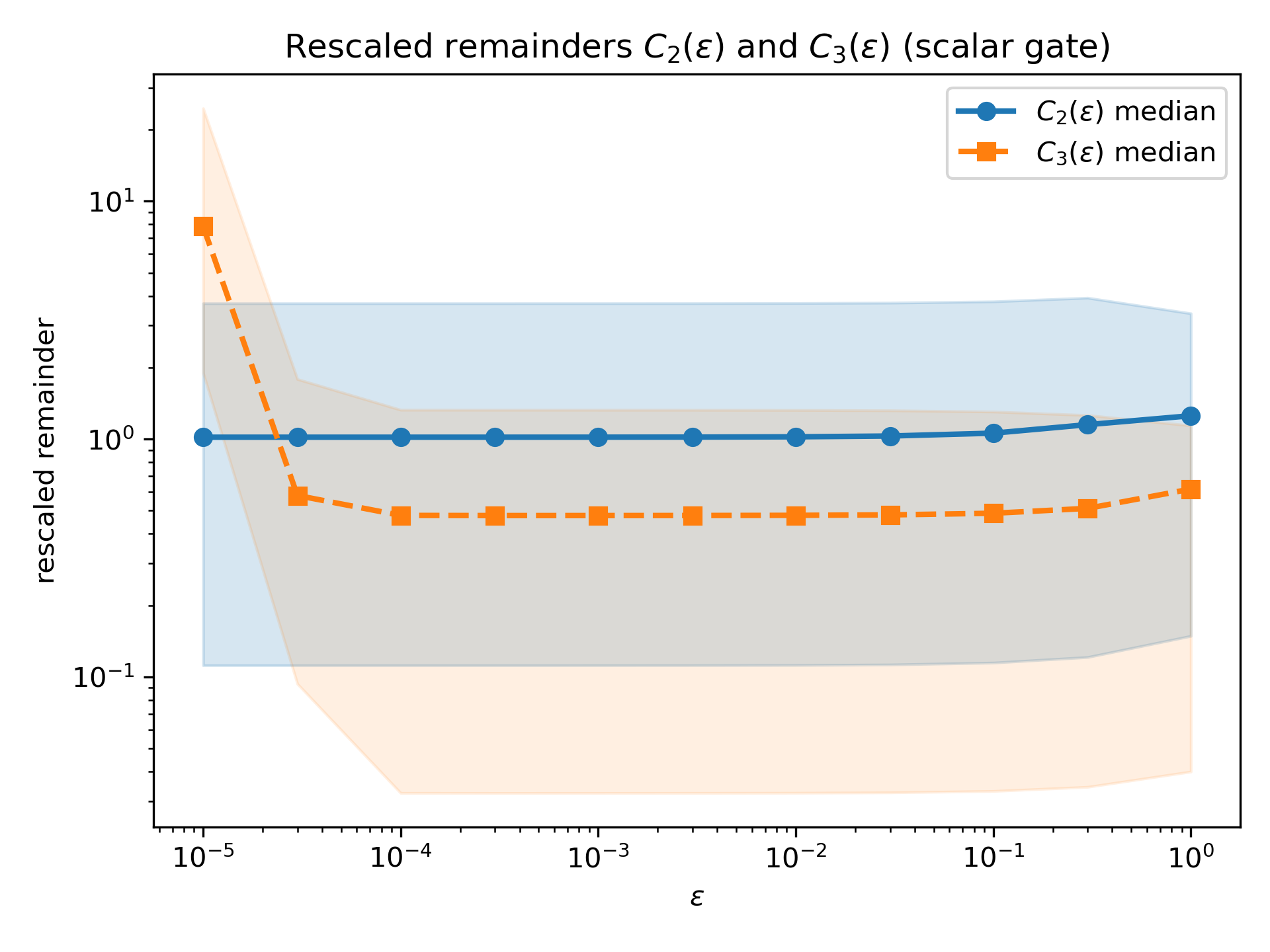}

\medskip

\includegraphics[width=0.45\textwidth]{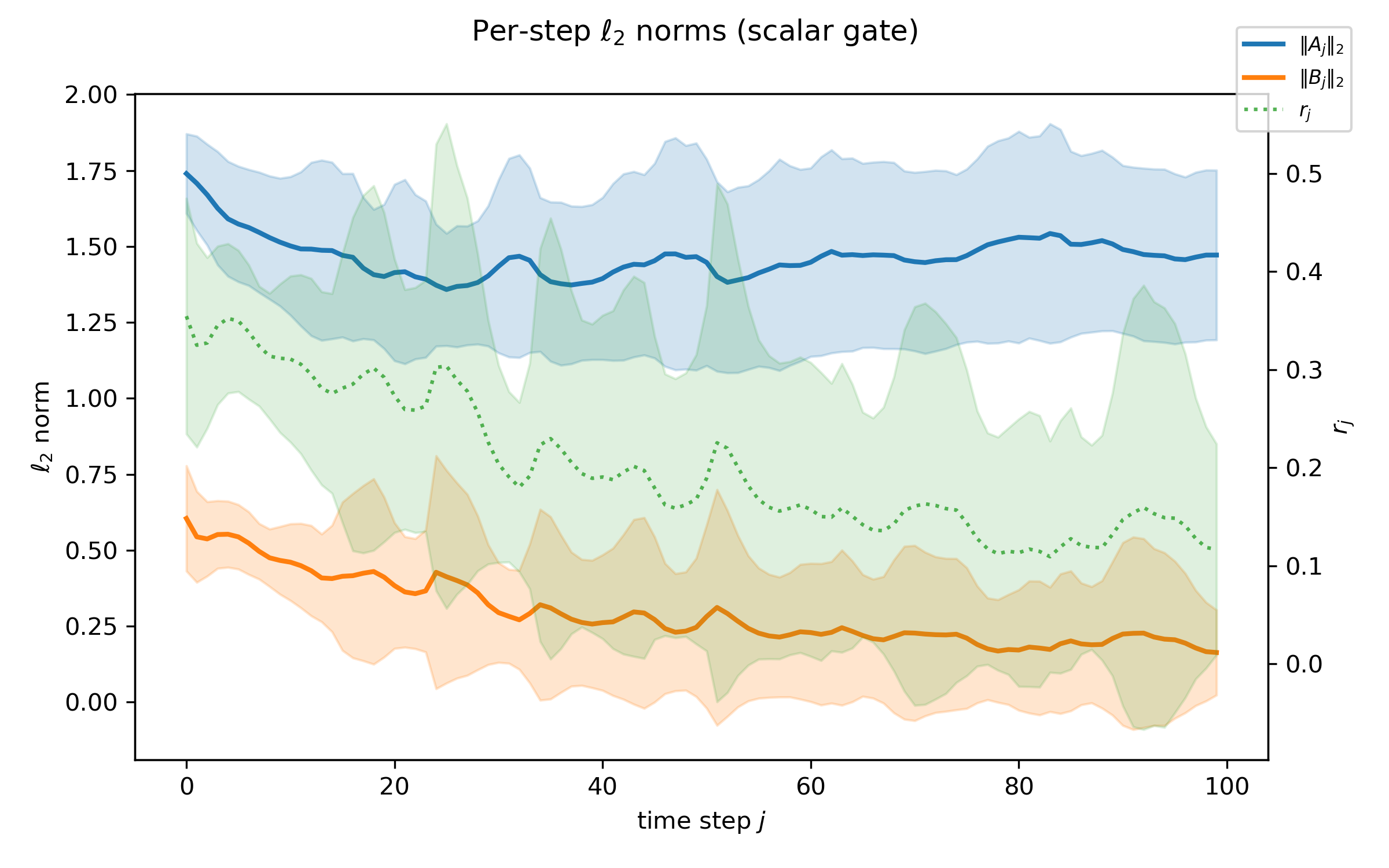}
\includegraphics[width=0.45\textwidth]{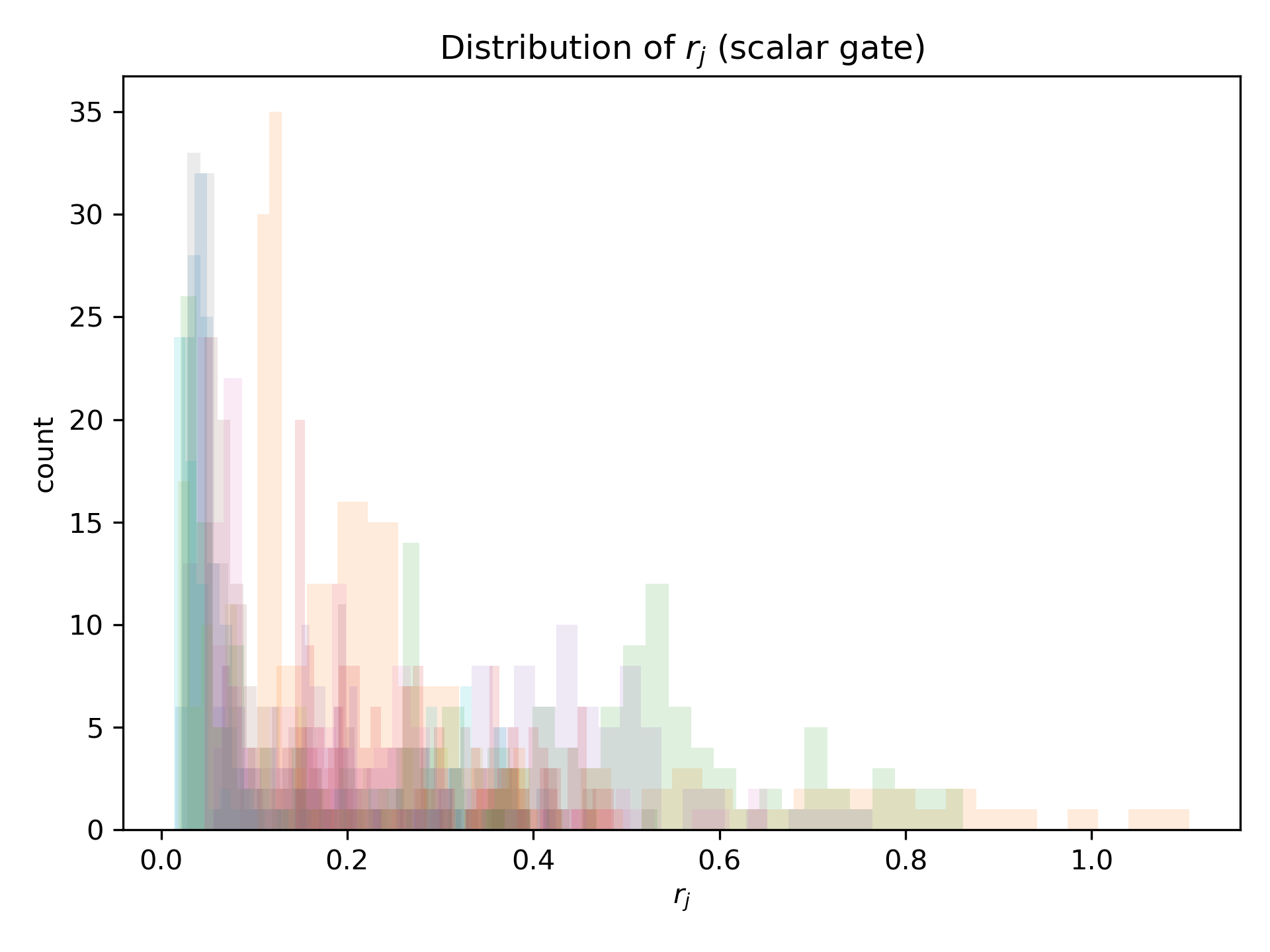}

\caption{Scalar-gated RNN.
Top: truncation error versus $\varepsilon$ (left) and rescaled
remainders $C_2(\varepsilon)$ and $C_3(\varepsilon)$ (right).
Bottom: per-step norms $\|A_j\|_2$ and $\|B_j\|_2$
(left) and distribution of ratios $r_j$ (right).}
\label{fig:first_order_scalar}
\end{figure}

\paragraph{Scalar-gate model}
Figure~\ref{fig:first_order_scalar} reports these diagnostics for the scalar-gated RNN.
The truncation error (top-left panel) follows a clear $\varepsilon^2$ scaling across the full range up to $\varepsilon=1$, confirming the second-order accuracy of the first-order expansion.
The combined remainder panel (top-right) shows that $C_2(\varepsilon)$ remains close to constant across the sweep, while $C_3(\varepsilon)$ is also approximately constant once $\varepsilon$ is large enough to avoid roundoff-dominated behavior at the very smallest tested values.
The key diagnostic is therefore the absence of systematic drift with $\varepsilon$ in the resolved perturbative regime, which confirms that truncation at first order is numerically sound and that the residual after subtracting the second-order term is cubic, as predicted.
The bottom panels examine the perturbative matrices: the per-step norms $\|B_j\|_2$ remain consistently smaller than $\|A_j\|_2$ (bottom-left), and the ratio distribution $r_j$ (bottom-right) is typically small (median $\approx 0.15$, with the bulk of the distribution below $0.5$ for the scalar-gated model), confirming that the gate-induced corrections are a controlled perturbation of the dominant dynamics even where the rank-1 scalar correction concentrates its energy along a single direction.

\begin{figure}[t]
\centering
\includegraphics[width=0.45\textwidth]{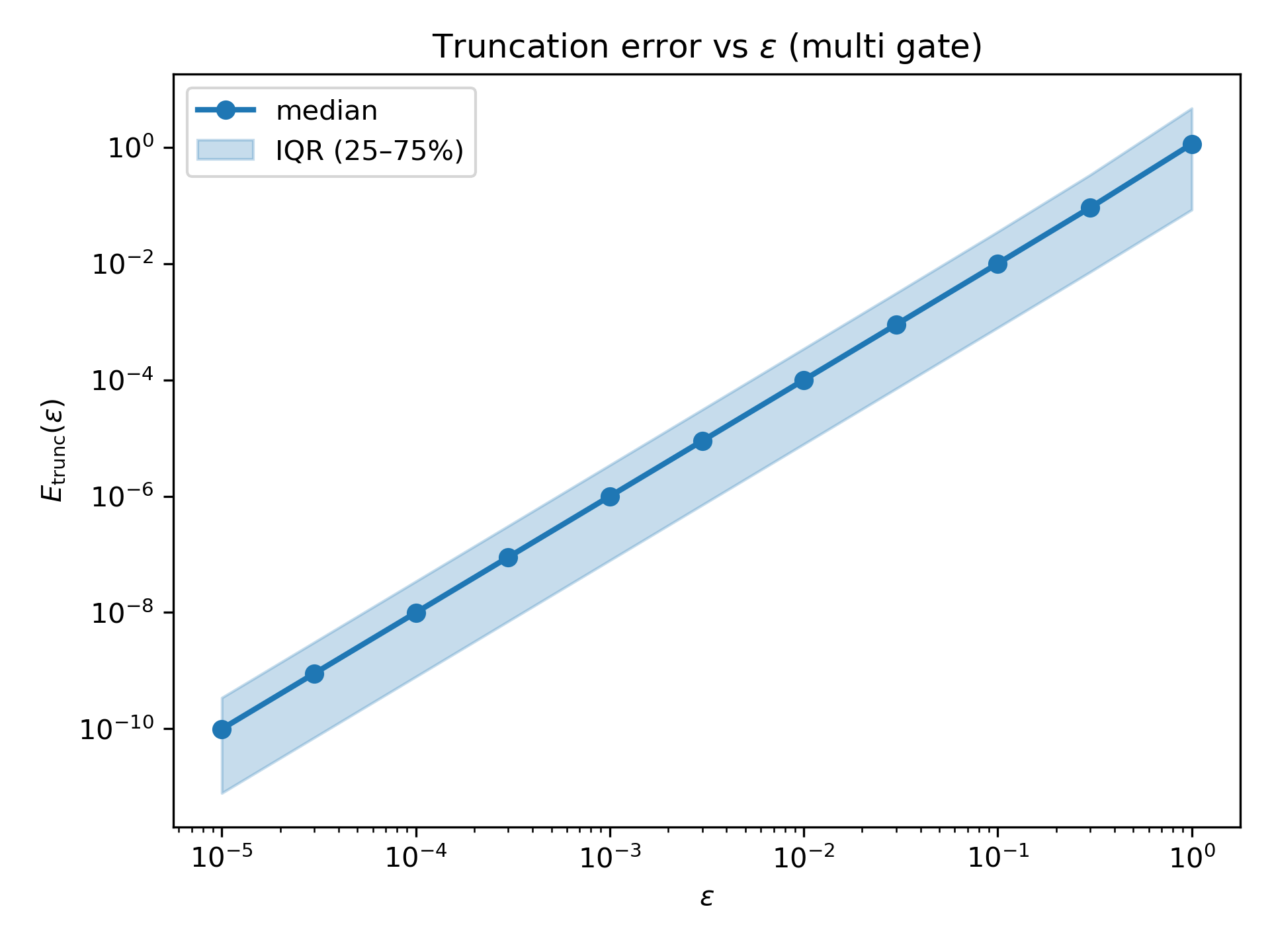}
\includegraphics[width=0.45\textwidth]{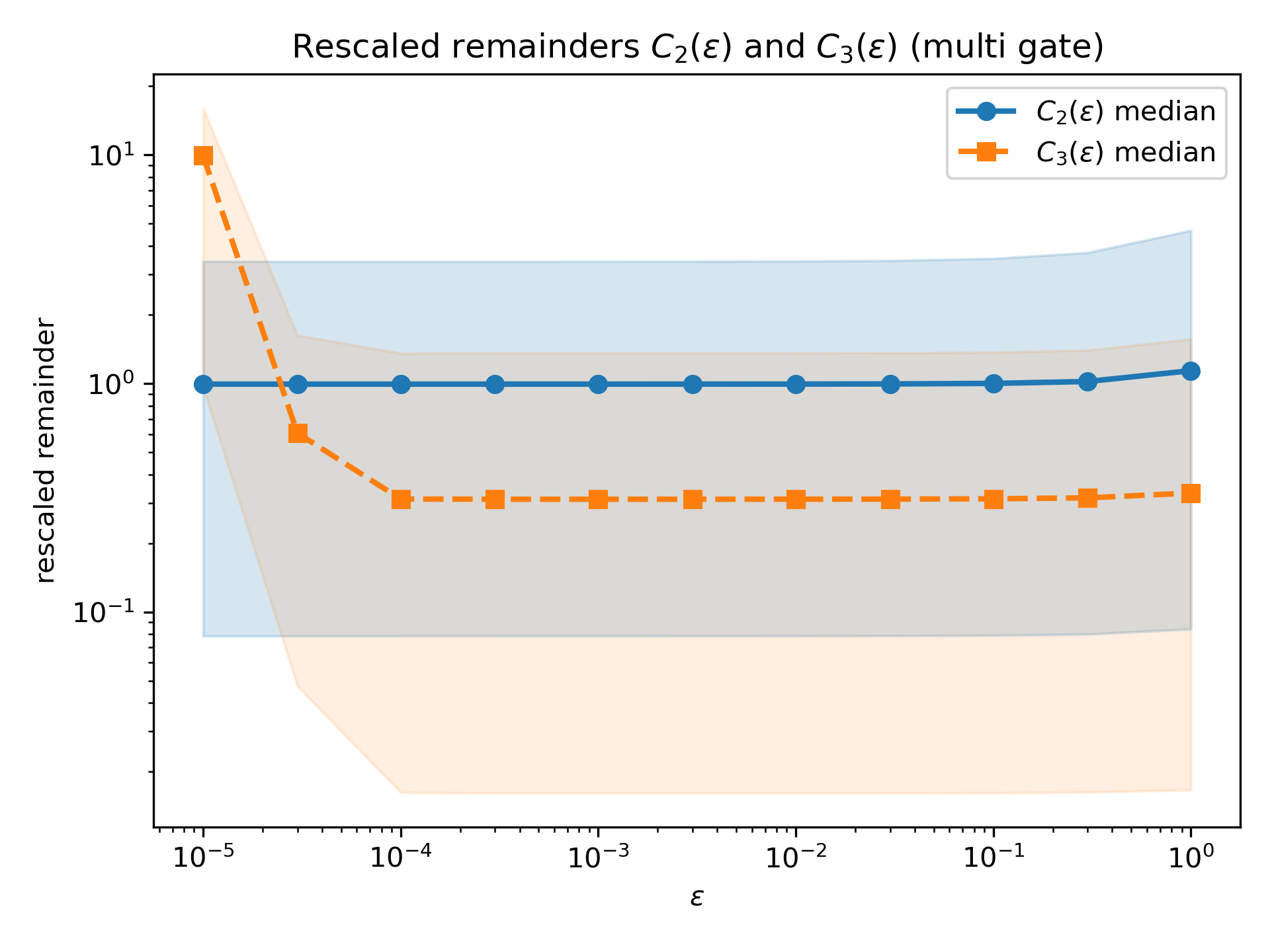}

\medskip

\includegraphics[width=0.45\textwidth]{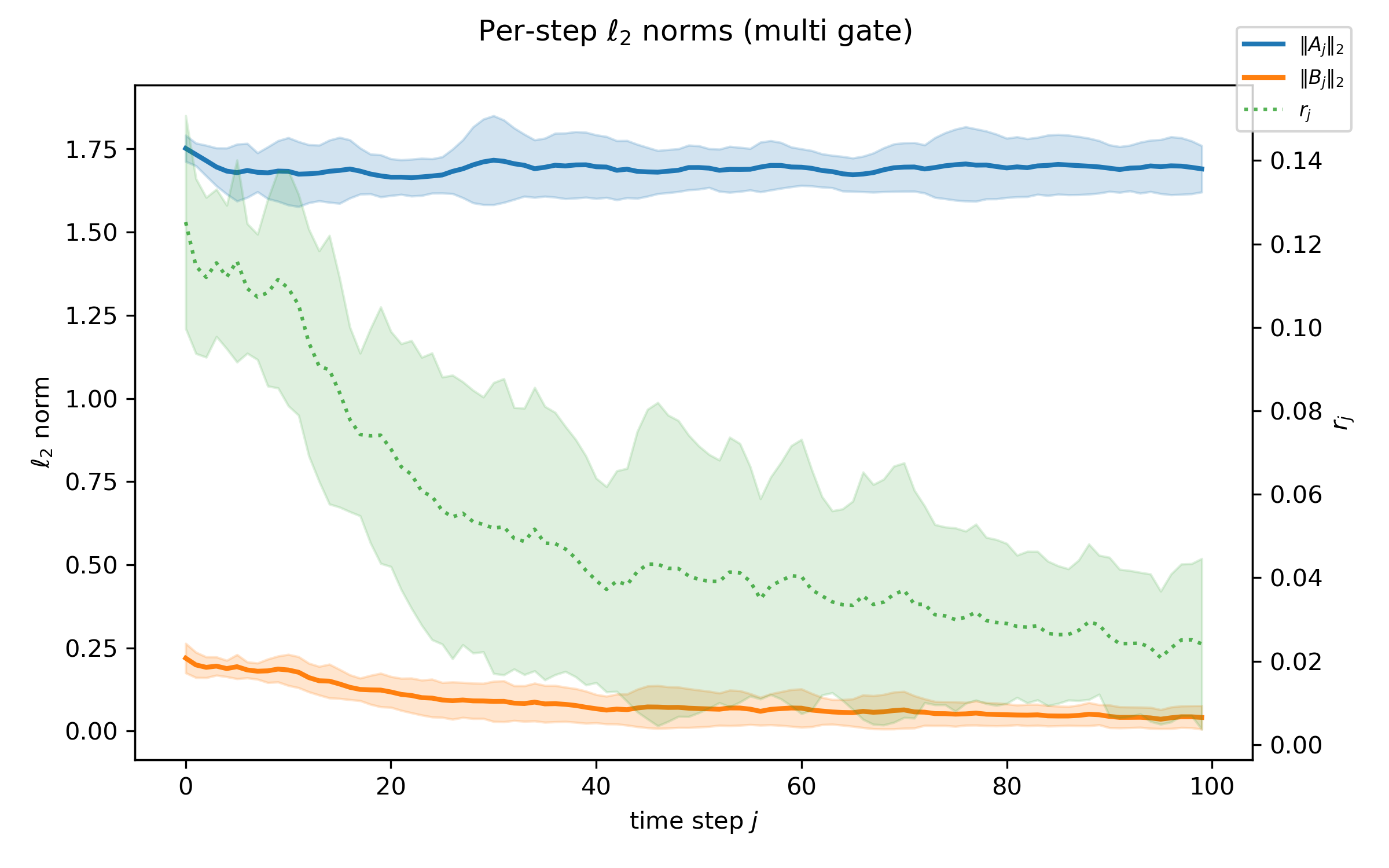}
\includegraphics[width=0.45\textwidth]{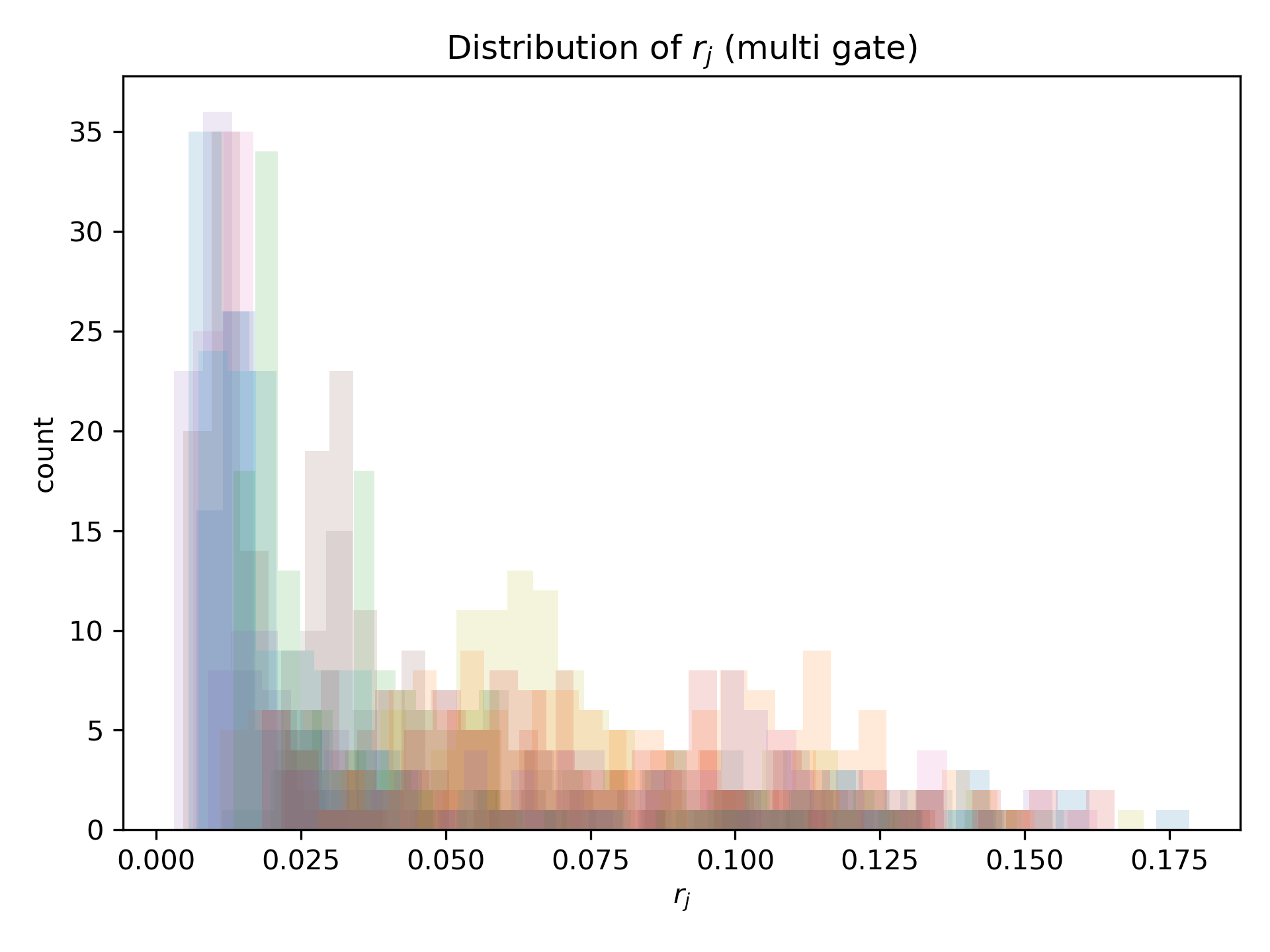}

\caption{Multi-gated RNN.
Top: truncation error versus $\varepsilon$ (left) and rescaled
remainders $C_2(\varepsilon)$ and $C_3(\varepsilon)$ (right).
Bottom: per-step norms $\|A_j\|_2$ and $\|B_j\|_2$
(left) and distribution of ratios $r_j$ (right).}
\label{fig:first_order_multi}
\end{figure}

\paragraph{Multi-gate model}
Figure~\ref{fig:first_order_multi} shows the analogous diagnostics for the multi-gated RNN.
The same $\varepsilon^2$ scaling of the truncation error is observed, and the combined remainder panel again shows a nearly constant $C_2(\varepsilon)$ together with an approximately constant $C_3(\varepsilon)$ away from the roundoff-dominated extreme small-$\varepsilon$ regime.
The ratio $r_j$ is even more tightly concentrated at small values (median $\approx 0.04$, with essentially the entire distribution below $0.2$), reflecting the additional structure of the multi-gate parameterization.
Overall, both architectures comfortably satisfy the perturbative regime assumed by the first-order expansion.

\subsection{Scope and limitations of the first-order approximation}
\label{sec:first_order_scope}

The goal of the first-order expansion in Appendix~\ref{sec:matrix_product_expansion} is
not to provide an exact description of the full Jacobian product
dynamics, but rather to make explicit how gating mechanisms
couple state-space time-scales with parameter updates during
training.
By isolating the leading-order contributions of the Jacobian
factors, the expansion reveals how gate values modulate the
magnitude and direction of gradient propagation.
In particular, the analysis highlights the emergence of neuronwise effective learning rates, which act as mesoscopic variables describing how temporal credit assignment is modulated across recurrent steps.

Focusing on neuronwise contributions is meaningful when the
dominant effect of gating is to modulate the diagonal structure
of the Jacobian factors.
In this regime, the gate products capture the main attenuation of gradients associated with each
neuron.
The simulations reported in
Section~\ref{sec:simulations_first-order_expansion} support
this interpretation: the perturbative matrices $B_j$ remain
consistently small relative to the dominant matrices $A_j$,
the truncation error follows the expected $O(\varepsilon^2)$
scaling, and the rescaled remainders
$C_2(\varepsilon)$ and $C_3(\varepsilon)$ remain approximately
constant over the resolved perturbative range.
Taken together, these diagnostics indicate that, for the tasks and
architectures considered here, the first-order expansion
captures the principal mechanism by which gating shapes
effective learning rates, while the residual corrections remain
controlled.

At the same time, this approximation is not expected to hold
uniformly across all regimes of recurrent learning.
In particular, the perturbative assumption may break down when
off-diagonal interactions dominate the Jacobian structure.
A simple example arises when strong recurrent mixing causes
gradient transport to propagate across many neurons before
returning to the original coordinate.
In such cases, higher-order terms in the matrix product
expansion can accumulate and significantly alter the effective
gradient dynamics.
Similarly, regimes involving highly nonlinear state trajectories
or strongly saturated gates may produce correction terms that
are no longer negligible relative to the dominant dynamics.

These limitations do not invalidate the approach; rather, they
clarify the level of description targeted by the present analysis.
The first-order expansion provides a mathematically tractable
framework for isolating neuronwise time-scales and revealing
their role in the coupling between state dynamics and parameter
updates.
More refined descriptions are possible by extending the
perturbative framework to include higher-order terms or by
explicitly modeling off-diagonal transport effects.
Such extensions would naturally lead to the notion of
\emph{between-neuron time-scales}, in which credit assignment
is described not only by neuronwise retention but also by
transport processes across neurons.

From this perspective, the first-order expansion should be
interpreted as the leading-order component of a broader
hierarchy of models describing temporal credit assignment.
The present work focuses on the neuronwise level of this
hierarchy, while more complex descriptions, including
inter-neuron transport mechanisms, remain an interesting
direction for future investigation.

\clearpage
\subsection{Adam}
\label{sec:adam}

Adam~\cite{kingma2014adam} combines ideas from momentum methods and adaptive learning rate algorithms.  
Like Adadelta and RMSprop, it maintains an exponentially decaying average of past squared gradients $v_{t,l}$, but it also tracks an exponentially decaying average of past (unsquared) gradients $m_{t,l}$, akin to momentum:  
\begin{align}
m_{t,l} &= \beta_1 m_{t,l-1} + (1-\beta_1) \frac{\partial\mathcal{E}_t}{\partial \theta_l}, \\
v_{t,l} &= \beta_2 v_{t,l-1} + (1-\beta_2) \left(\frac{\partial\mathcal{E}_t}{\partial \theta_l}\right)^2,
\end{align}
where $\beta_1, \beta_2 \in (0,1)$ are decay rates, and the squaring in $v_{t,l}$ is applied component-wise.  

To correct the initialization bias introduced by these moving averages, Adam computes bias-corrected estimates:
\begin{align}
\hat{m}_{t,l} &= \frac{m_{t,l}}{1-\beta_1^l}, \\
\hat{v}_{t,l} &= \frac{v_{t,l}}{1-\beta_2^l}.
\end{align}

The parameter update is then:
\begin{equation}
\theta_{l+1} = \theta_l - \frac{\mu}{b} \sum_{t=1}^{b} \frac{\hat{m}_{t,l}}{\sqrt{\hat{v}_{t,l}} + \epsilon}, 
\quad \epsilon > 0,
\end{equation}
where $b$ is the mini-batch size and $\epsilon$ is a small constant to prevent division by zero.

\clearpage
\subsection{Simulations: Experimental details and figures}
\label{sec:supp_simulations}

\subsubsection{Effective learning rate induced by gates}
\label{sec:supp_s1}

\paragraph{Tasks}
We evaluate the effective learning rate profiles on two synthetic sequence-to-scalar regression tasks with contrasting credit-assignment structures.

\emph{Adding task}.
The input sequence has $n_i - 1 = 3$ noise channels drawn i.i.d.\ from $\mathrm{Uniform}(0,1)$ plus one binary mask channel.
At each sample, two time steps are marked---one drawn uniformly from the first half $\{0,\dots,T/2-1\}$ and one from the second half $\{T/2,\dots,T-1\}$, by setting the mask channel to~$1$.
The scalar target is the sum of the first noise channel at the two marked positions:
$y = x_{t_1,\,1} + x_{t_2,\,1}$.
Credit assignment is therefore sparse: only two time steps per sequence carry task-relevant information, and the network must remember them across potentially long lags.

\emph{Sinmix task.}
The input consists of two sinusoidal channels with random frequencies $f_1, f_2 \sim \mathrm{Uniform}(1,7)$ and random phases $\phi_1, \phi_2 \sim \mathrm{Uniform}(0,2\pi)$, sampled over $[0,1]$ at $T$ equally spaced points, plus $n_i - 3$ Gaussian noise channels ($\sigma=0.1$) and one unused mask channel (kept for architectural consistency).
The scalar target contrasts early and late temporal averages of the two sinusoids:
\begin{equation}
  y \;=\; \bigl(\bar{s}_{1}^{\,\mathrm{late}} - \bar{s}_{1}^{\,\mathrm{early}}\bigr)
      \;+\; \bigl(\bar{s}_{2}^{\,\mathrm{early}} - \bar{s}_{2}^{\,\mathrm{late}}\bigr),
\end{equation}
where $\bar{s}_{i}^{\,\mathrm{early}}$ and $\bar{s}_{i}^{\,\mathrm{late}}$ denote the mean of signal~$i$ over the first and last quarter of the sequence, respectively.
Unlike the adding task, credit assignment is distributed: the target depends on extended temporal windows rather than isolated time steps, requiring the network to integrate information over broad intervals.

\paragraph{Experimental setup}
All three RNN architectures---leaky integrator~\eqref{eq:rnn_dt_leaky}, scalar gate~\eqref{eq:rnn_dt_singlegate}, and multi-gate~\eqref{eq:rnn_dt_multigate}---share the same dimensions: input size $n_i = 4$, hidden size $n_h = 64$, output size $n_o = 1$.
The leaky model uses a fixed mixing coefficient $\alpha = 0.8$.
All models are trained with plain SGD (no momentum, no weight decay) at learning rate $\mu = 10^{-2}$ for $800$ gradient steps.
Each step draws a fresh random mini-batch of size $B_{\mathrm{train}} = 64$ from the task generator (online setting; no finite dataset or epochs).
Sequence length is $T = 80$ for both tasks.

Probing is performed at checkpoints $\ell \in \{0, 50, 200, 400, 600, 800\}$.
At each checkpoint, a fixed probe batch of $B_{\mathrm{probe}} = 12$ sequences (drawn once at initialization with a separate random seed) is forwarded through the network.
We build the exact one-step Jacobians $J_j$, compute $\|M_{t,k}\|_2$ from the exact singular values of $M_{t,k}$, and compute the gate-product predictor $P_{t,k}$.
The log--log slope $s(\ell)$ is fitted on the central $98\%$ quantile range of the predictor values (trimming the extreme $1\%$ on each side).
All reported quantities are averaged over $20$ independent random seeds (seeds $1$--$20$).

\clearpage
\paragraph{Results and figures}

\begin{figure}[th!]
  \centering
  \begin{minipage}[t]{0.48\linewidth}
    \centering
    \includegraphics[width=\linewidth]{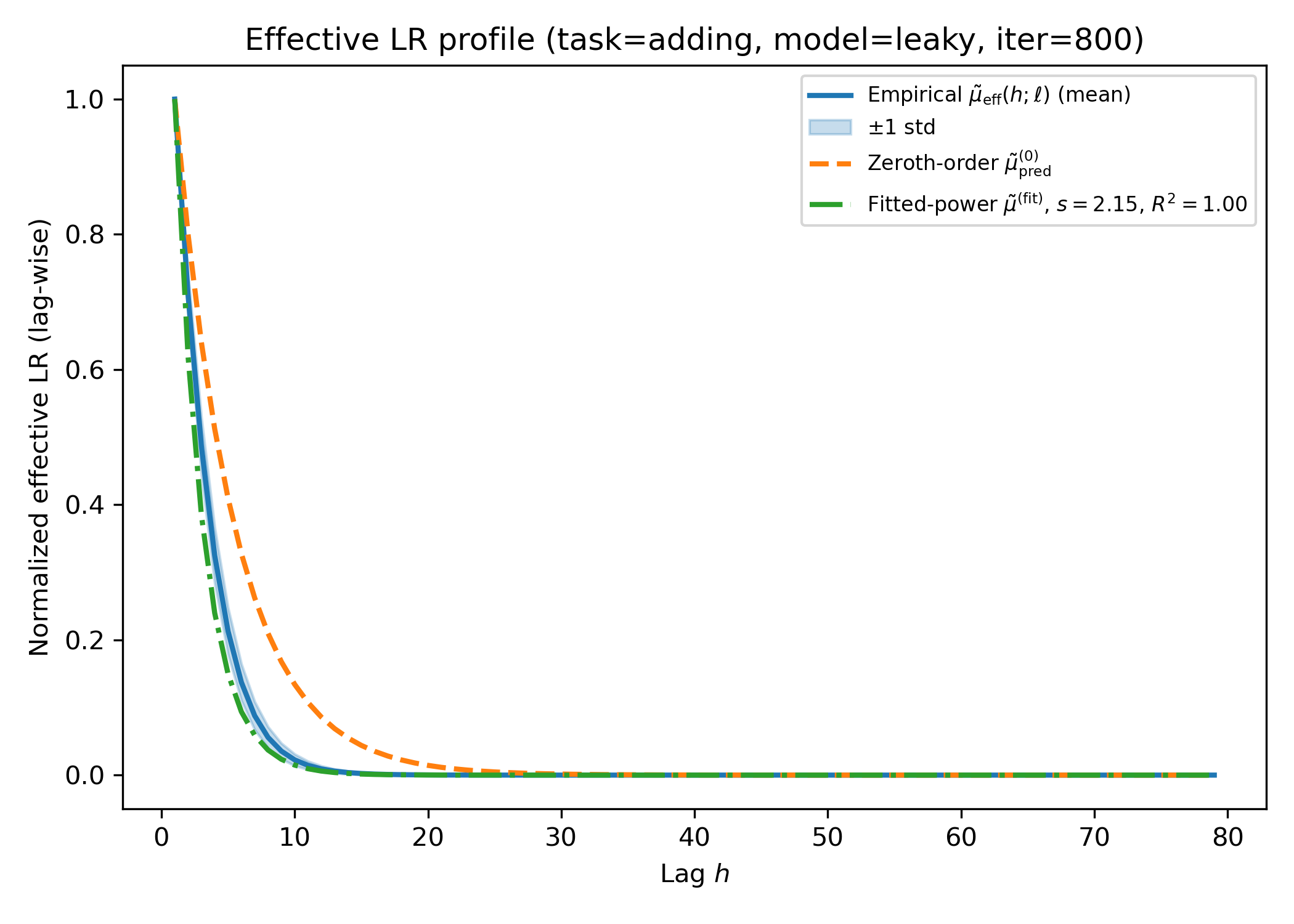}
  \end{minipage}\hfill
  \begin{minipage}[t]{0.48\linewidth}
    \centering
    \includegraphics[width=\linewidth]{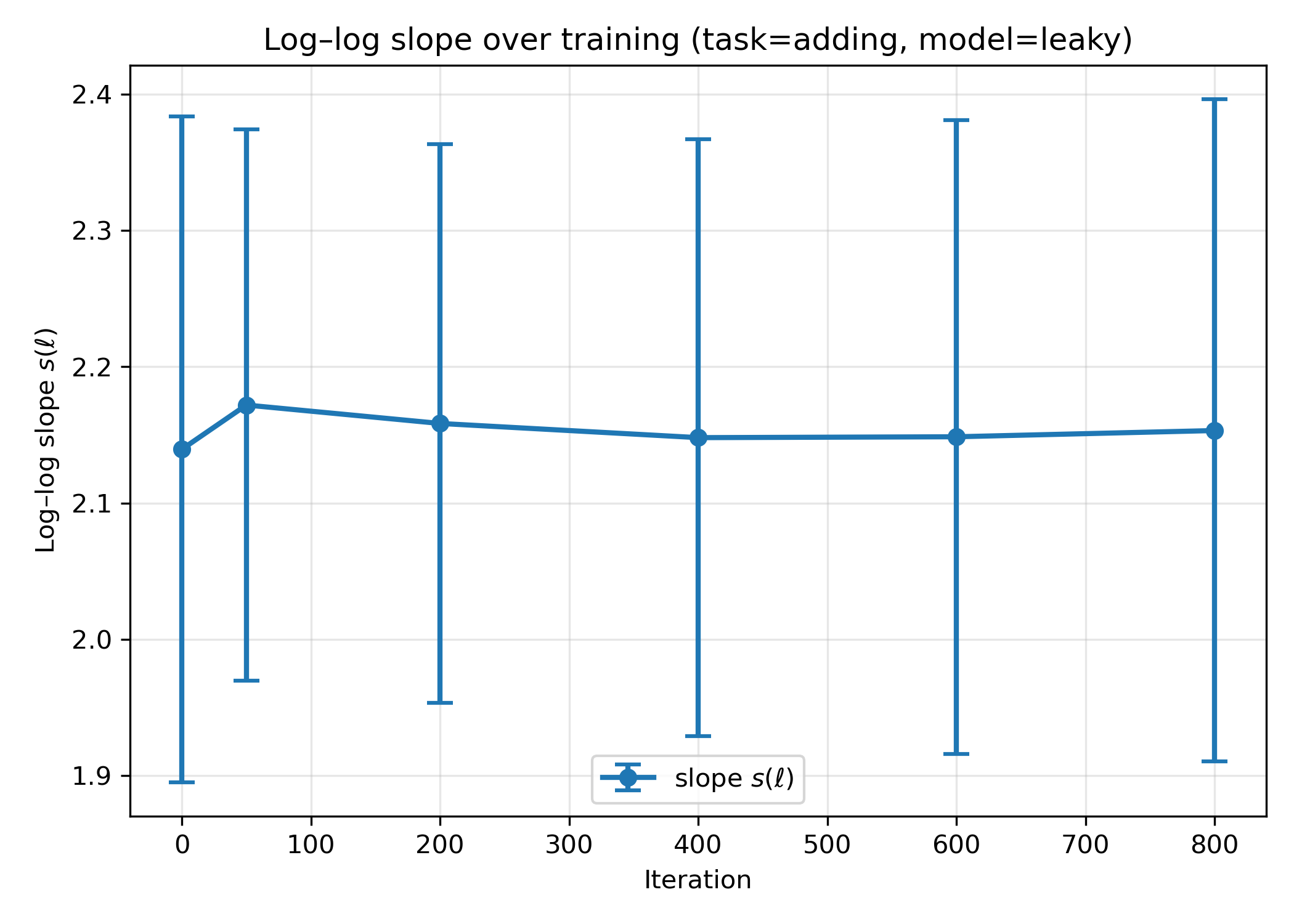}
  \end{minipage}

  \vspace{0.6em}

  \begin{minipage}[t]{0.48\linewidth}
    \centering
    \includegraphics[width=\linewidth]{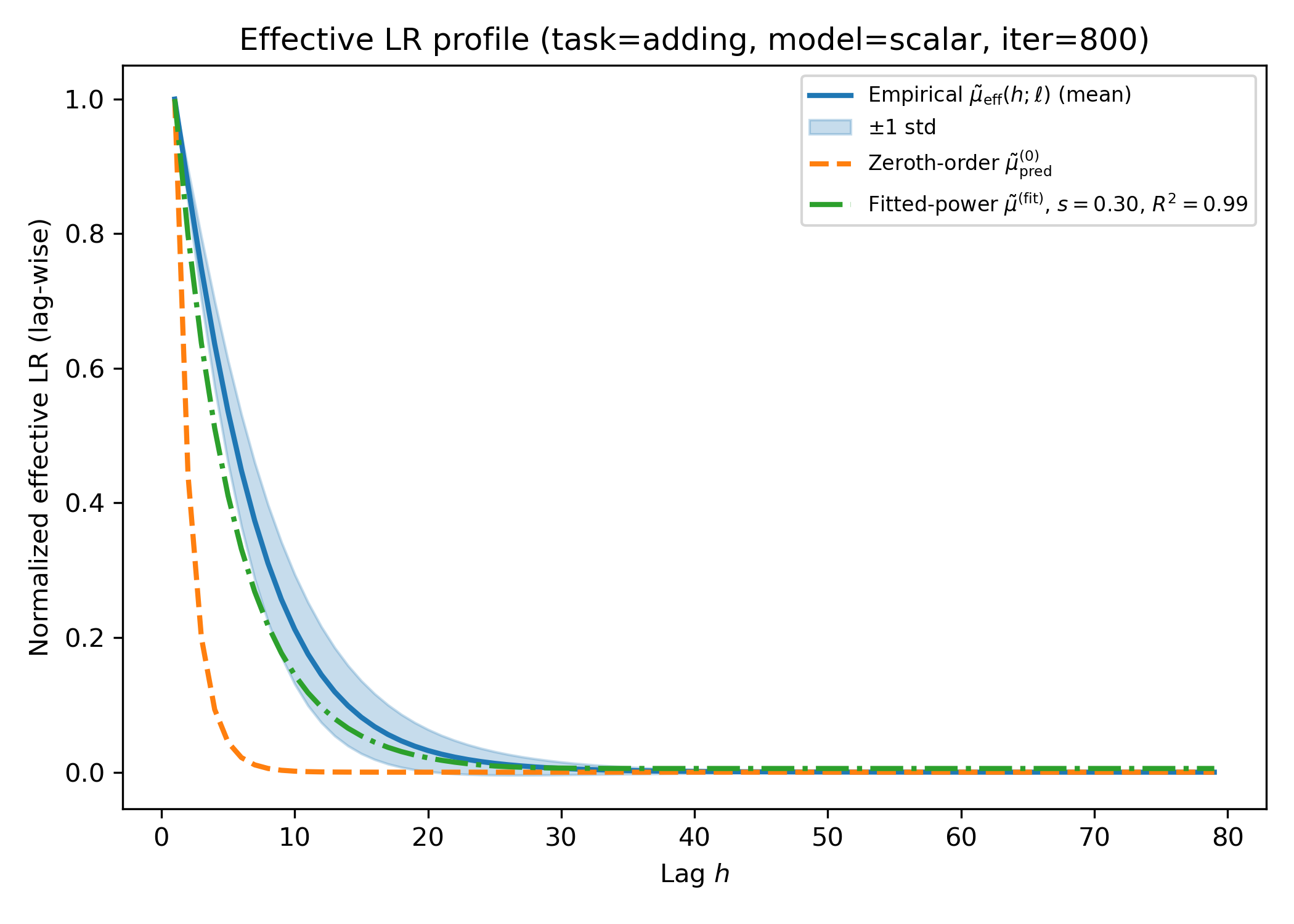}
  \end{minipage}\hfill
  \begin{minipage}[t]{0.48\linewidth}
    \centering
    \includegraphics[width=\linewidth]{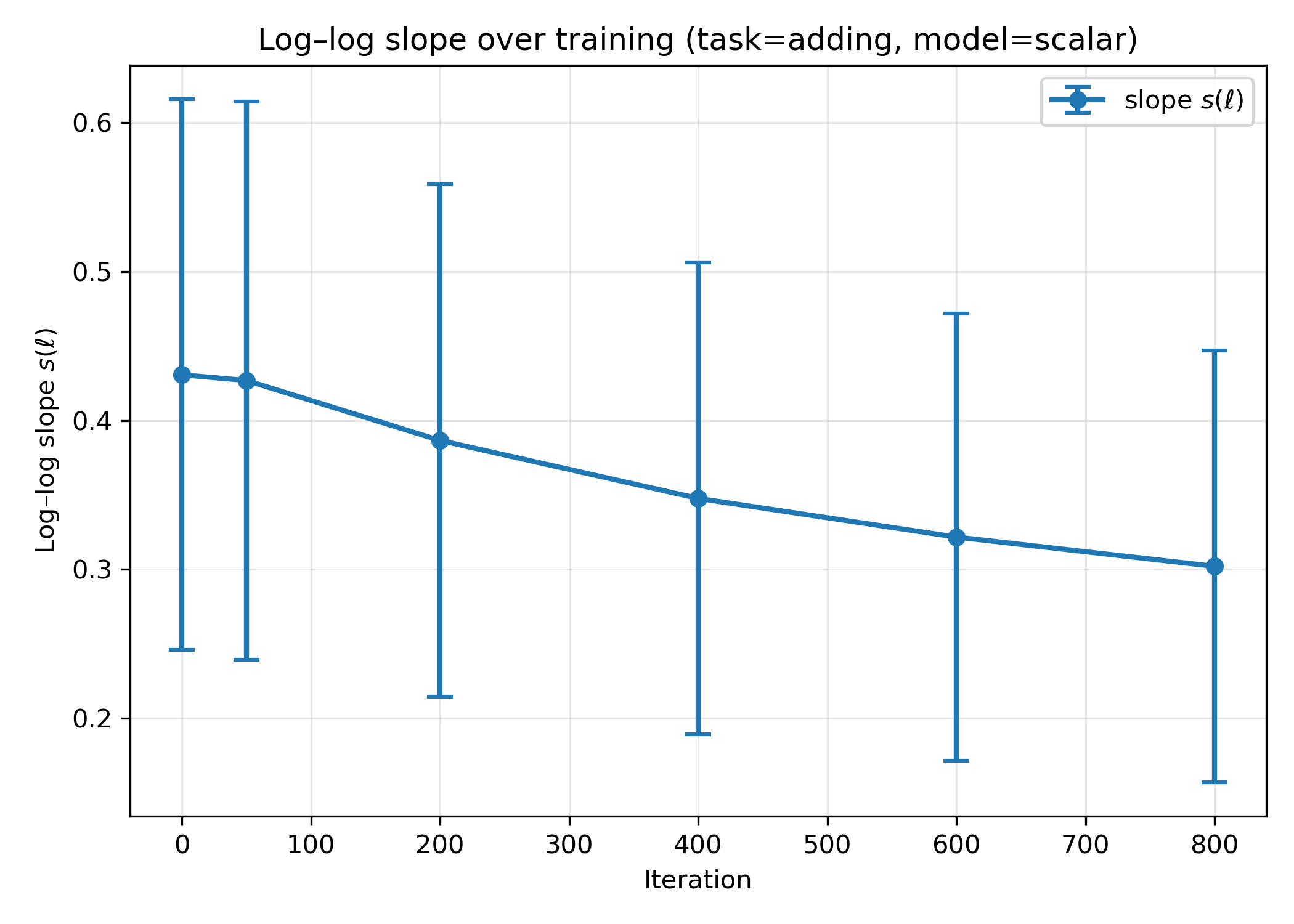}
  \end{minipage}

  \vspace{0.6em}

  \begin{minipage}[t]{0.48\linewidth}
    \centering
    \includegraphics[width=\linewidth]{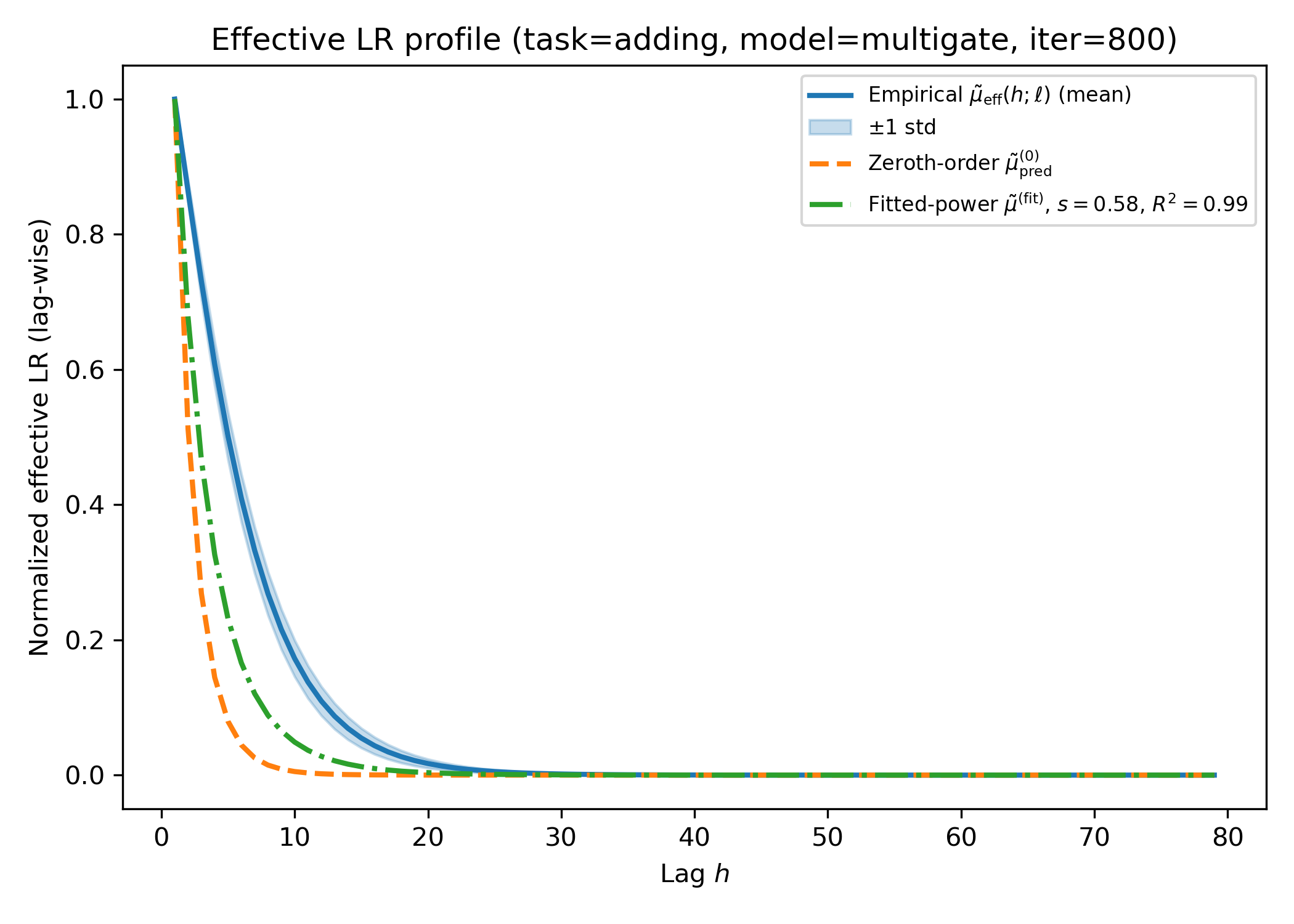}
  \end{minipage}\hfill
  \begin{minipage}[t]{0.48\linewidth}
    \centering
    \includegraphics[width=\linewidth]{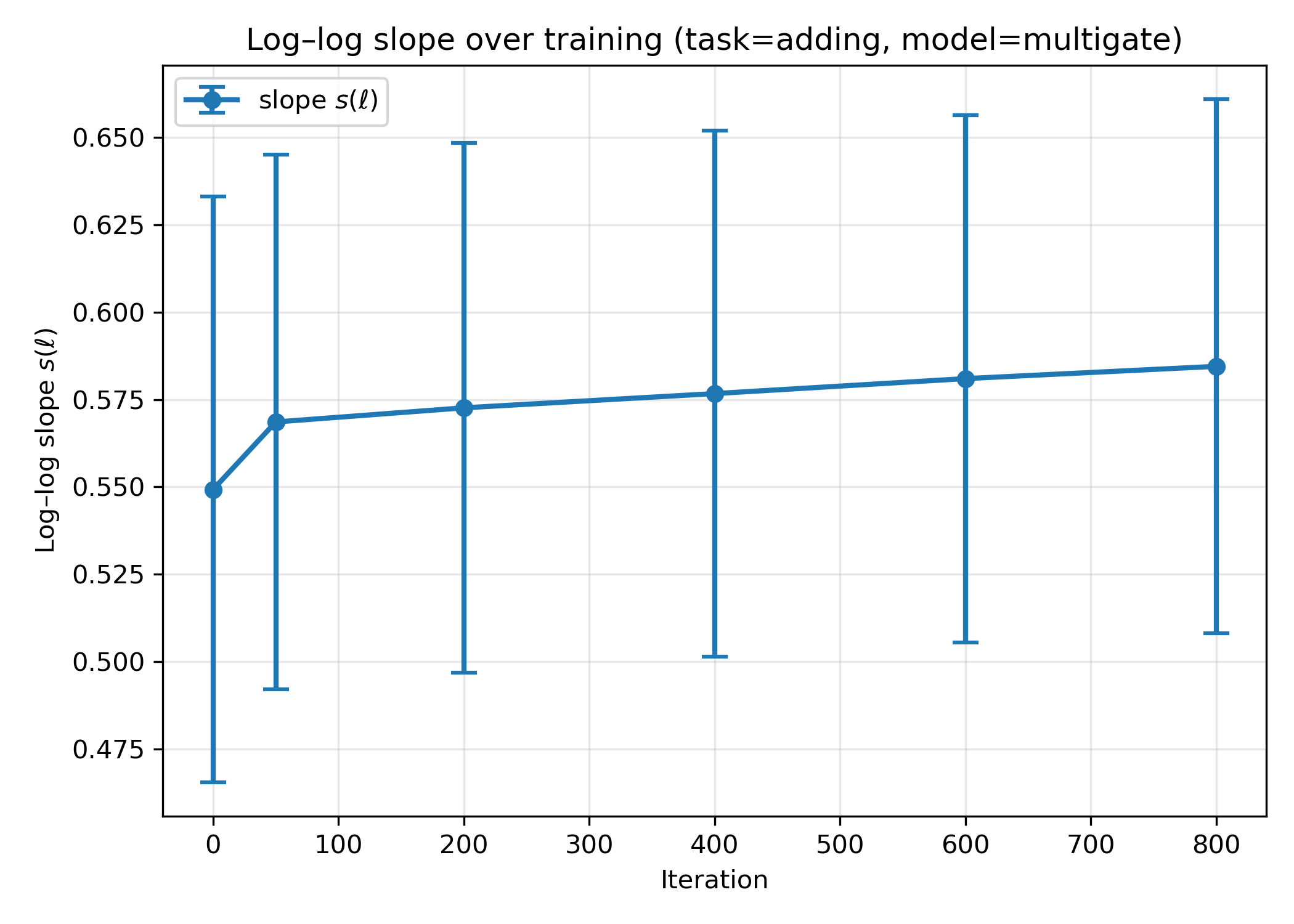}
  \end{minipage}

  \caption{Results on the adding task. Each row corresponds to one model (top: leaky RNN with constant~$\alpha$; middle: scalar-gated RNN; bottom: multi-gated RNN). Left column: normalized effective LR profile $\tilde{\mu}_{\mathrm{eff}}(h;\ell)$ at the final checkpoint with zeroth-order and fitted-power overlays. Right column: fitted log--log slope $s(\ell)$ across training iterations.}
  \label{fig:s1_adding}
\end{figure}

\begin{figure}[th!]
  \centering
  \begin{minipage}[t]{0.48\linewidth}
    \centering
    \includegraphics[width=\linewidth]{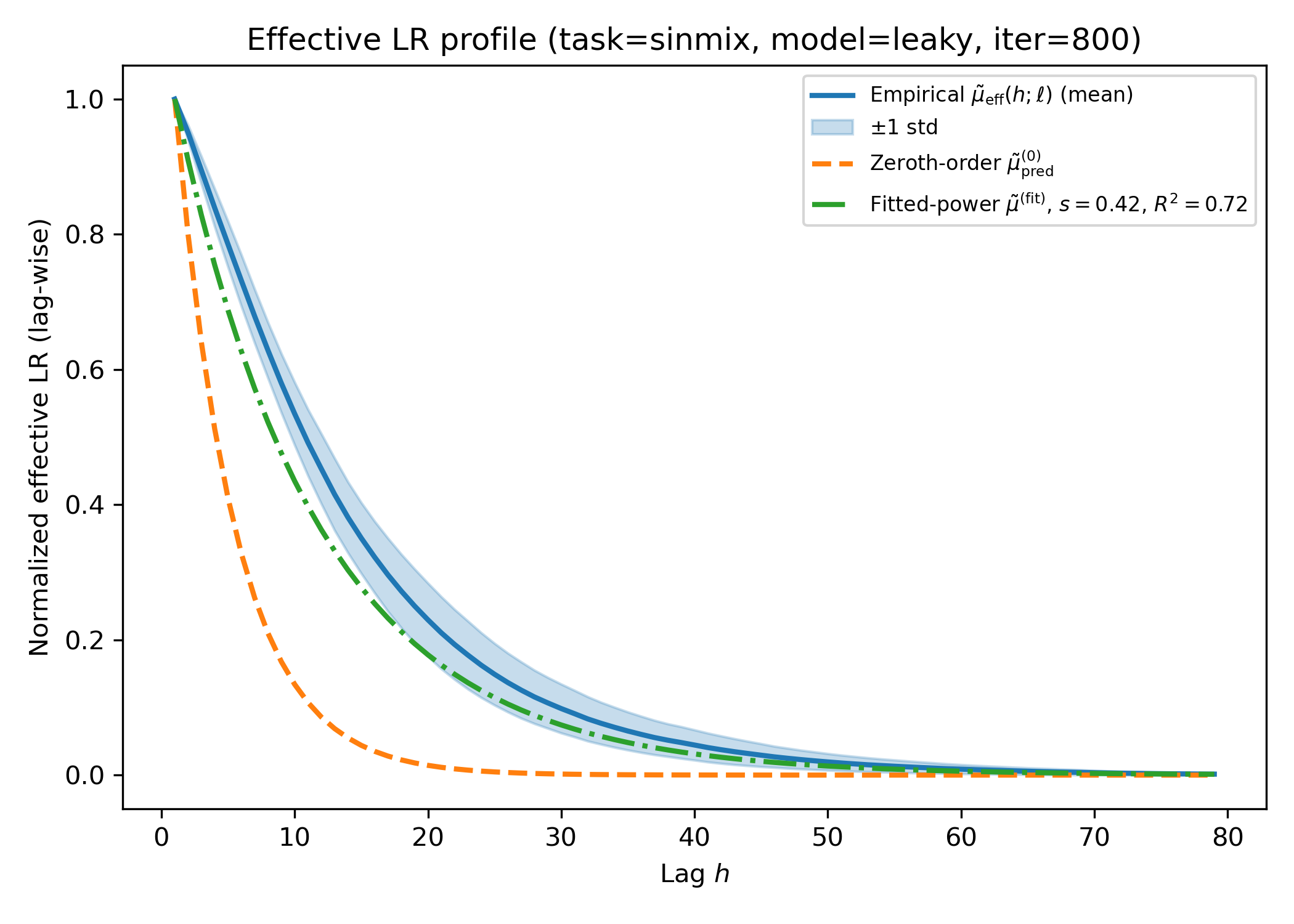}
  \end{minipage}\hfill
  \begin{minipage}[t]{0.48\linewidth}
    \centering
    \includegraphics[width=\linewidth]{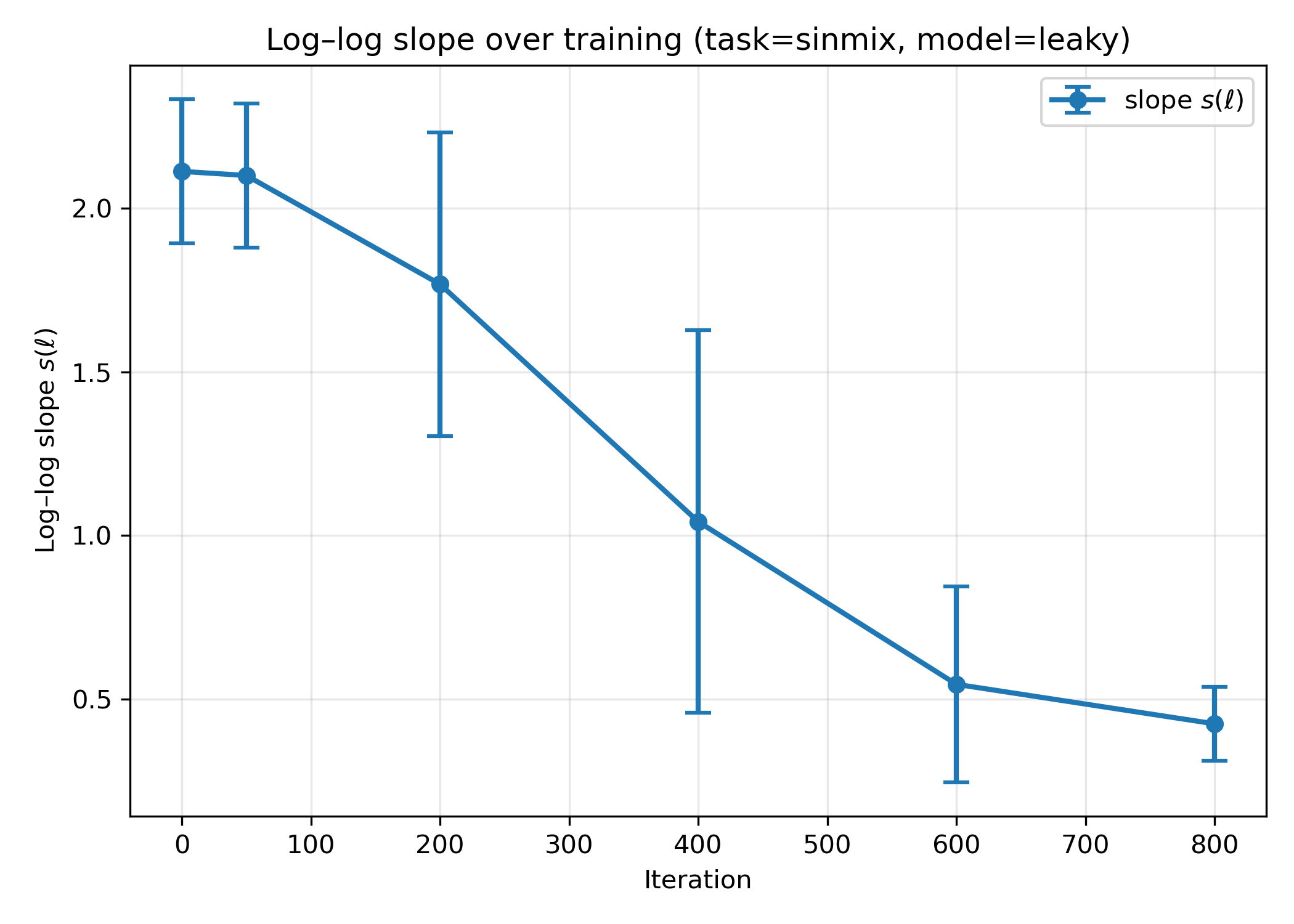}
  \end{minipage}

  \vspace{0.6em}

  \begin{minipage}[t]{0.48\linewidth}
    \centering
    \includegraphics[width=\linewidth]{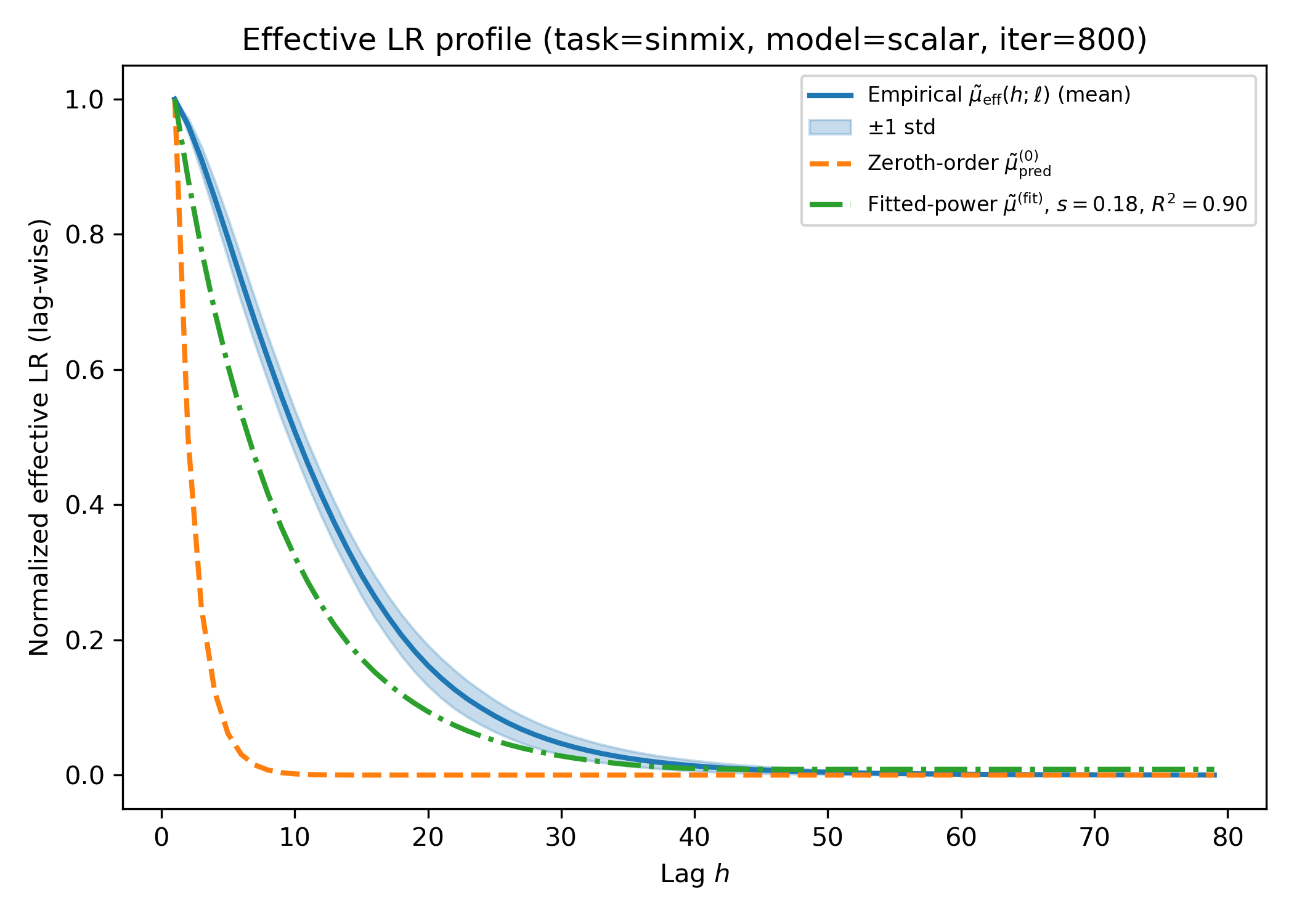}
  \end{minipage}\hfill
  \begin{minipage}[t]{0.48\linewidth}
    \centering
    \includegraphics[width=\linewidth]{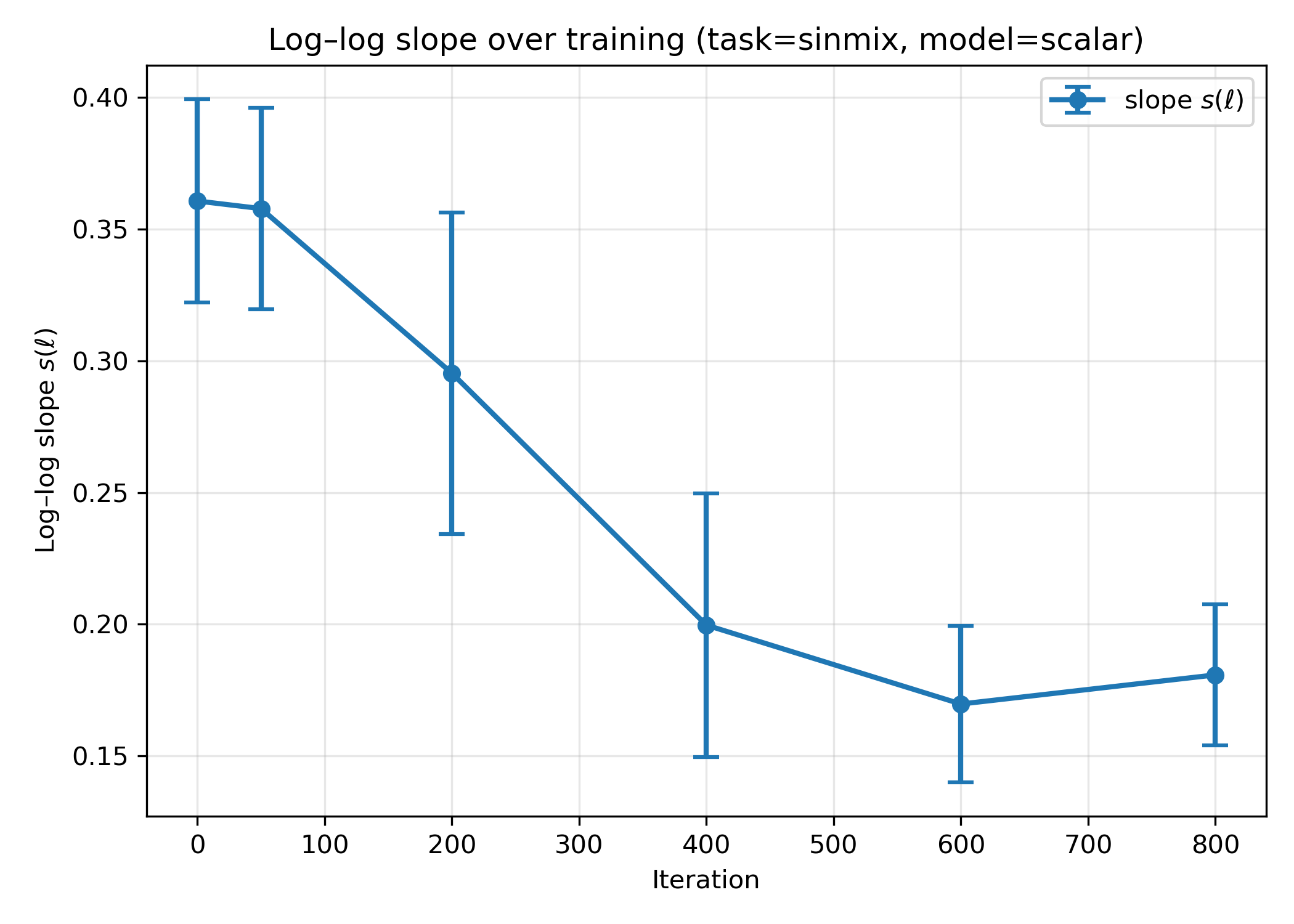}
  \end{minipage}

  \vspace{0.6em}

  \begin{minipage}[t]{0.48\linewidth}
    \centering
    \includegraphics[width=\linewidth]{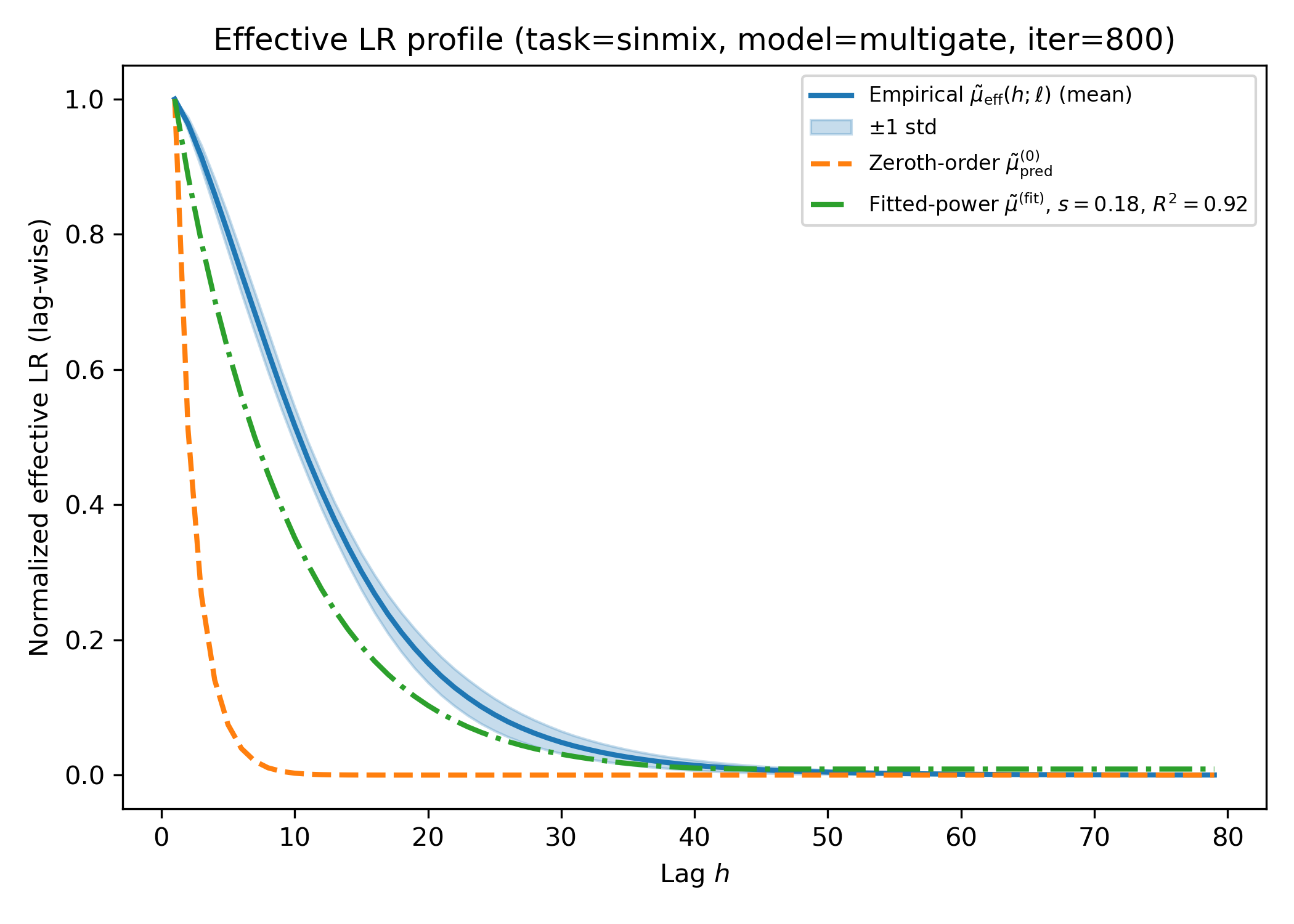}
  \end{minipage}\hfill
  \begin{minipage}[t]{0.48\linewidth}
    \centering
    \includegraphics[width=\linewidth]{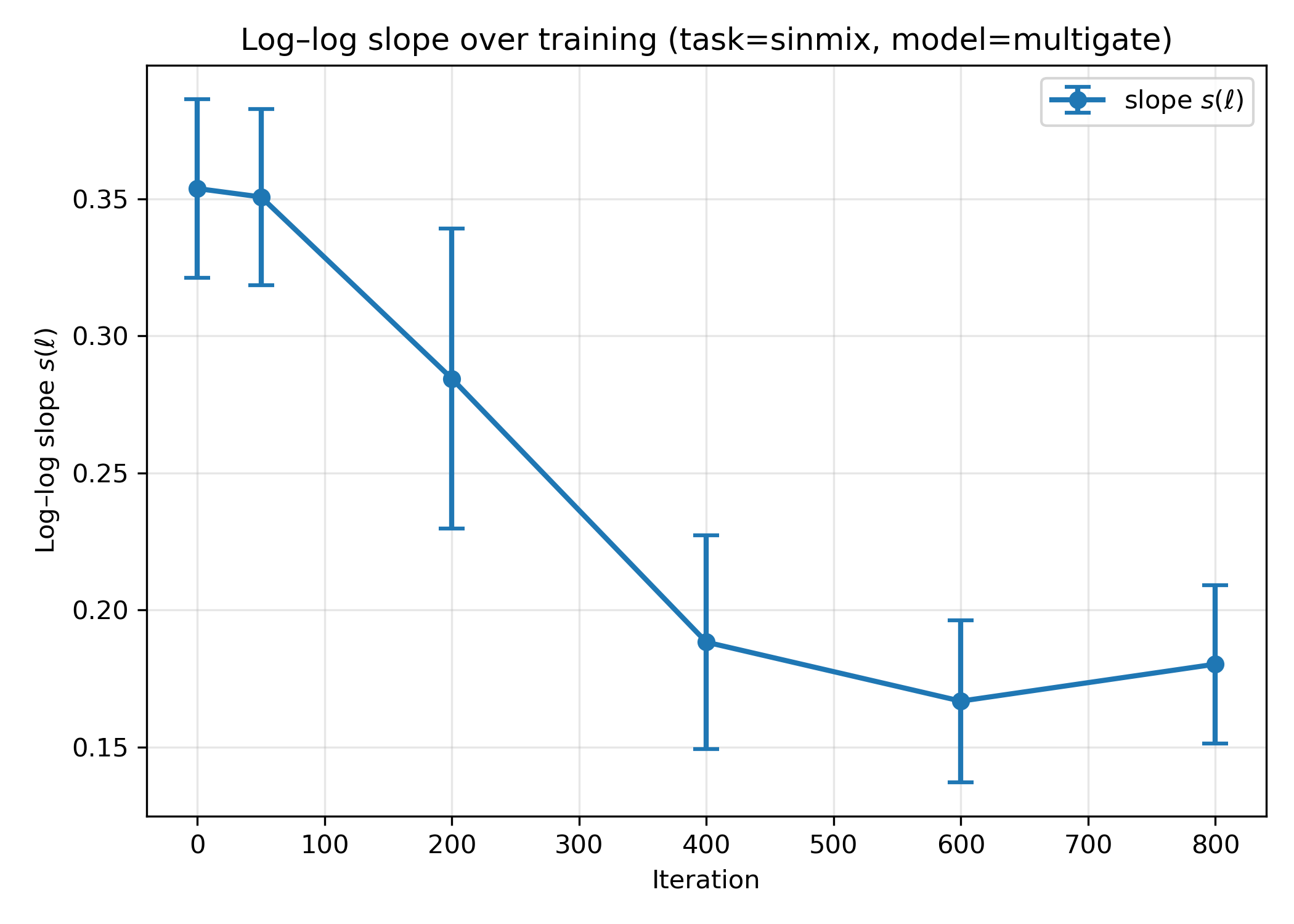}
  \end{minipage}

  \caption{Results on the sinmix task. Each row corresponds to one model (top: leaky RNN with constant~$\alpha$; middle: scalar-gated RNN; bottom: multi-gated RNN). Left column: normalized effective LR profile $\tilde{\mu}_{\mathrm{eff}}(h;\ell)$ at the final checkpoint with zeroth-order and fitted-power overlays. Right column: fitted log--log slope $s(\ell)$ across training iterations.}
  \label{fig:s1_sinmix}
\end{figure}

\clearpage
\subsubsection{Directional anisotropy: propagation vs.\ updates}
\label{sec:supp_s2}

\paragraph{Tasks}
We evaluate propagation and update anisotropy on five sequence-to-scalar regression tasks: four synthetic tasks at sequence length $T=120$ with input dimension $n_i=6$, plus permuted sequential MNIST (psMNIST) at $T=784$ as a canonical long-range real-world benchmark.
Depending on the task, non-target channels of the synthetic inputs are distractor noise, a mask channel, or zeros.

\emph{Adding}.
Same two-marker structure as in Section~\ref{sec:supp_s1}: one marker is drawn uniformly from the first half and one from the second half of the sequence; the target is the sum of the first input channel at the two marked positions.

\emph{AR(2).}
A latent autoregressive process $x_t = a_1 x_{t-1} + a_2 x_{t-2} + \varepsilon_t$, with $a_1=1.6$, $a_2=-0.7$, and $\varepsilon_t \sim \mathcal{N}(0,0.05^2)$.
The scalar target is $x_T$.
In the current implementation, the input channels are independent distractor noise and do not contain the latent process, so this task should be interpreted as a control condition rather than as a genuine AR identification problem.

\emph{Delay-sum.}
The target is a uniformly weighted sum of the first input channel at four fixed delays from the end of the sequence:
$y = \tfrac{1}{4}\sum_{i} u_{T-1-d_i,\,1}$, with delays $d \in \{4, 12, 24, 36\}$.
This task probes sensitivity at specific, predetermined lags.

\emph{Moving-average.}
The target is the mean of the first input channel over the last $W=8$ time steps:
$y = \tfrac{1}{W}\sum_{t=T-W}^{T-1} u_{t,\,1}$.
Credit assignment is concentrated on a short recent window.

\emph{psMNIST (regression surrogate).}
Permuted sequential MNIST, a widely used long-range-dependency benchmark in the recurrent and gating literature.
Each $28\times28$ MNIST image is flattened into a length-$784$ pixel sequence and a single, fixed pixel permutation is applied to every sequence before training.
Pixel intensities are mapped into the first input channel, and the remaining channels are held at zero to keep the input dimension $n_i=6$ consistent with the synthetic tasks.
Because the S2 measurements (AI/CE of Jacobian products, AI/CE of the gradient covariance, per-step perturbative ratio $r_j$) characterize transport and update geometry rather than classification accuracy, we train under a regression surrogate: the digit label is normalized to $[0,1]$ (label$/9$) and the network is trained with the same mean-squared error loss used for the synthetic tasks, so that results are directly comparable across all five tasks.
The scalar target is the normalized label emitted at $t=T$, with $T=784$.
Within our framework, psMNIST serves two complementary roles: (i) as a real-world sequence benchmark complementing the four synthetic tasks, and (ii) as a long-lag stress test of the propagation-anisotropy measurements, since $T=784$ is well beyond the horizon probed on synthetic data.

\paragraph{Experimental setup}
Three RNN architectures are compared with the same hidden size $n_h=64$ and output size $n_o=1$:
(i)~a plain RNN (leaky integrator with $\alpha=1$, i.e., $x_t = \tanh(W_r x_{t-1} + W_i u_t)$) trained with Adam (learning rate $10^{-3}$);
(ii)~the scalar-gated RNN~\eqref{eq:rnn_dt_singlegate} trained with SGD ($\mu=10^{-2}$);
(iii)~the multi-gated RNN~\eqref{eq:rnn_dt_multigate} trained with SGD ($\mu=10^{-2}$).
The pairing of plain+Adam with gated+SGD is an intentional design choice that isolates two distinct sources of update anisotropy. The plain model is paired with Adam so that any anisotropy observed in its gradient covariance can be attributed to optimizer-induced preconditioning alone, through Adam's per-parameter adaptive second-moment rescaling. The gated models are paired with plain SGD so that any anisotropy observed there can be attributed to the recurrent architecture and gates only, with no optimizer-state contribution. Contrasting these two configurations is precisely what is needed to disentangle the optimizer-induced and gating-induced components of the realized update geometry, which is the main goal of this experiment.

\emph{Shared training protocol.}
All three architecture+optimizer pairs are trained under an identical protocol, so that any remaining differences reflect architecture or optimizer rather than the training recipe.
All models are trained for $1200$ gradient steps (online setting, batch size $B=64$) with no early stopping.
The primary protocol does not apply gradient clipping: as shown by the optimization diagnostics reported below, gradient L2 norms remain finite throughout training for every (task, architecture, seed) configuration, and the training loss shows a clear downward trend over training on all task/model pairs, so clipping is not needed as a safeguard.
As a robustness check, the full S2 analysis has been reproduced under an otherwise identical configuration with a uniform global gradient-norm clip (max norm $1$) applied before each update; the resulting gated-configuration AI values are within a fraction of a percent of the unclipped values on every task (largest observed shift: adding scalar-gated, $\sim\!0.23\%$; adding multi-gated, $\sim\!0.11\%$), and CE is essentially unchanged. For plain+Adam the AI shift is usually a few percent, up to about $9\%$ on delay-sum and about $29\%$ on psMNIST, while the gated-vs.-plain+Adam ordering remains unchanged on every task; the gap ranges from narrow on adding ($\sim\!1.3\times$) and a factor of a few on AR(2) ($\sim\!3\times$) to roughly one order of magnitude on moving-average and nearly two orders of magnitude on delay-sum and psMNIST. This confirms that the observed anisotropy structure is architectural rather than protocol-induced.
Probing is performed at checkpoints $\ell \in \{0, 400, 800, 1200\}$.

\emph{Jacobian anisotropy.}
At each checkpoint, a fixed probe batch of $B_{\mathrm{probe}}=64$ sequences is forwarded through the network.
For each lag $h$ -- taken from $\{1,2,4,8,12,16,24,32,40\}$ on the synthetic tasks and extended to long lags on psMNIST -- we sample $64$ $(t,k)$ pairs with $t-k=h$ and compute the Jacobian product $M_{t,k} = \prod_{j=k+1}^{t} J_j$. We then compute the full singular spectrum of $M_{t,k}$ and retain the top $k_{\mathrm{svd}}=16$ singular directions for any downstream diagnostic that uses individual singular vectors; the full spectrum is always used when forming the cumulative-energy denominator, so that CE is never computed against a truncated normalization.
To prevent float32 underflow of $M_{t,k}$ at large lags (particularly on psMNIST where $h$ can reach hundreds of steps), the per-step Jacobians and their running product are accumulated in double precision before the SVD is computed; models and forward passes remain in single precision, so the numerical upgrade is confined to the diagnostic pipeline.
The anisotropy index $\mathrm{AI}_r$ and cumulative energy $\mathrm{CE}_r$ are then computed with $r=10$, using the scale-free definitions $\mathrm{AI}_r = \sigma_1 / \sigma_r$ and $\mathrm{CE}_r = \sum_{i=1}^{r}\sigma_i^2 / \sum_{i\ge 1}\sigma_i^2$; the denominator of $\mathrm{CE}_r$ uses the full singular spectrum of $M_{t,k}$, not a top-$k_{\mathrm{svd}}$ truncation.
When the spectrum becomes numerically degenerate ($\sigma_r$ below machine precision, or total energy below machine precision), the corresponding AI or CE value is recorded as missing rather than saturated at an arbitrarily large number, and the seed aggregation excludes missing entries.

\emph{Gradient covariance anisotropy.}
At each checkpoint, we collect per-sample gradients into a matrix $G \in \mathbb{R}^{m \times p}$ with $m=256$ probe samples and $p$ the total number of trainable parameters.
Rows are $\ell_2$-normalized and columns are mean-centered before computing the SVD of $G$.
$\mathrm{AI}_r(G)$ and $\mathrm{CE}_r(G)$ are reported with $r=10$, using the same scale-free definitions and missing-value convention as above.

All reported quantities are aggregated over $20$ independent random seeds using median and interquartile range (IQR), chosen for robustness to the heavy-tailed distributions that arise in singular-value ratios at long lags.

\paragraph{Optimization diagnostics}

Since the quantities analyzed in Simulation~2 -- lag-resolved Jacobian AI/CE and gradient-covariance AI/CE -- are computed on trained models, it is worth verifying that the underlying optimization runs are themselves well-behaved, so that the reported spectral structure is not an artifact of degenerate training dynamics. To that end, we instrument the training loop with a lightweight diagnostic pass that logs the per-iteration training MSE and the per-iteration gradient L2 norm. Twenty seeds are used at the log density of every $5$ iterations, yielding $241$ samples per iteration trajectory.

\begin{figure}[th!]
  \centering
  \begin{minipage}[t]{0.33\linewidth}
    \centering
    \includegraphics[width=\linewidth]{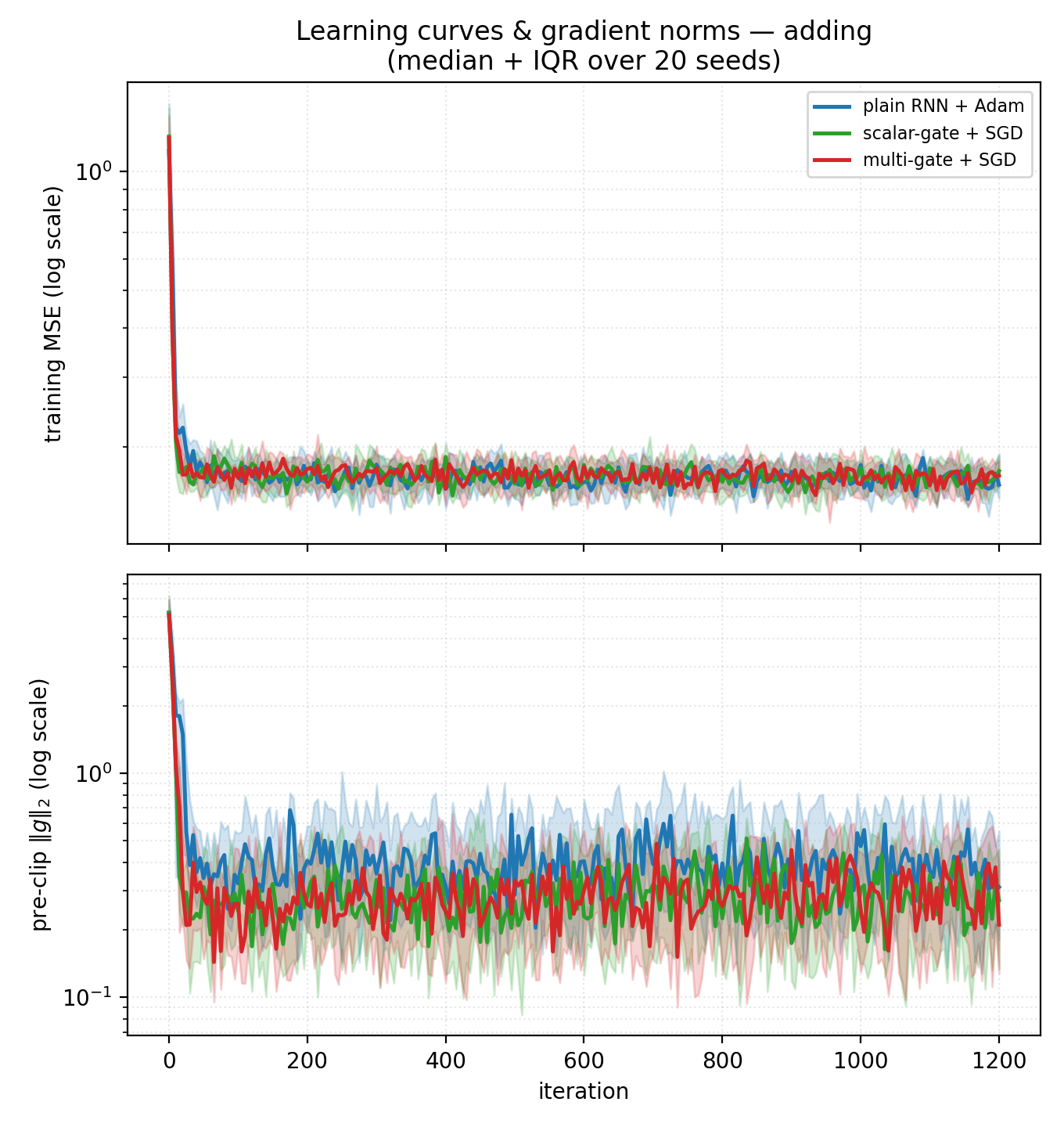}\\(a) adding
  \end{minipage}\hfill
  \begin{minipage}[t]{0.33\linewidth}
    \centering
    \includegraphics[width=\linewidth]{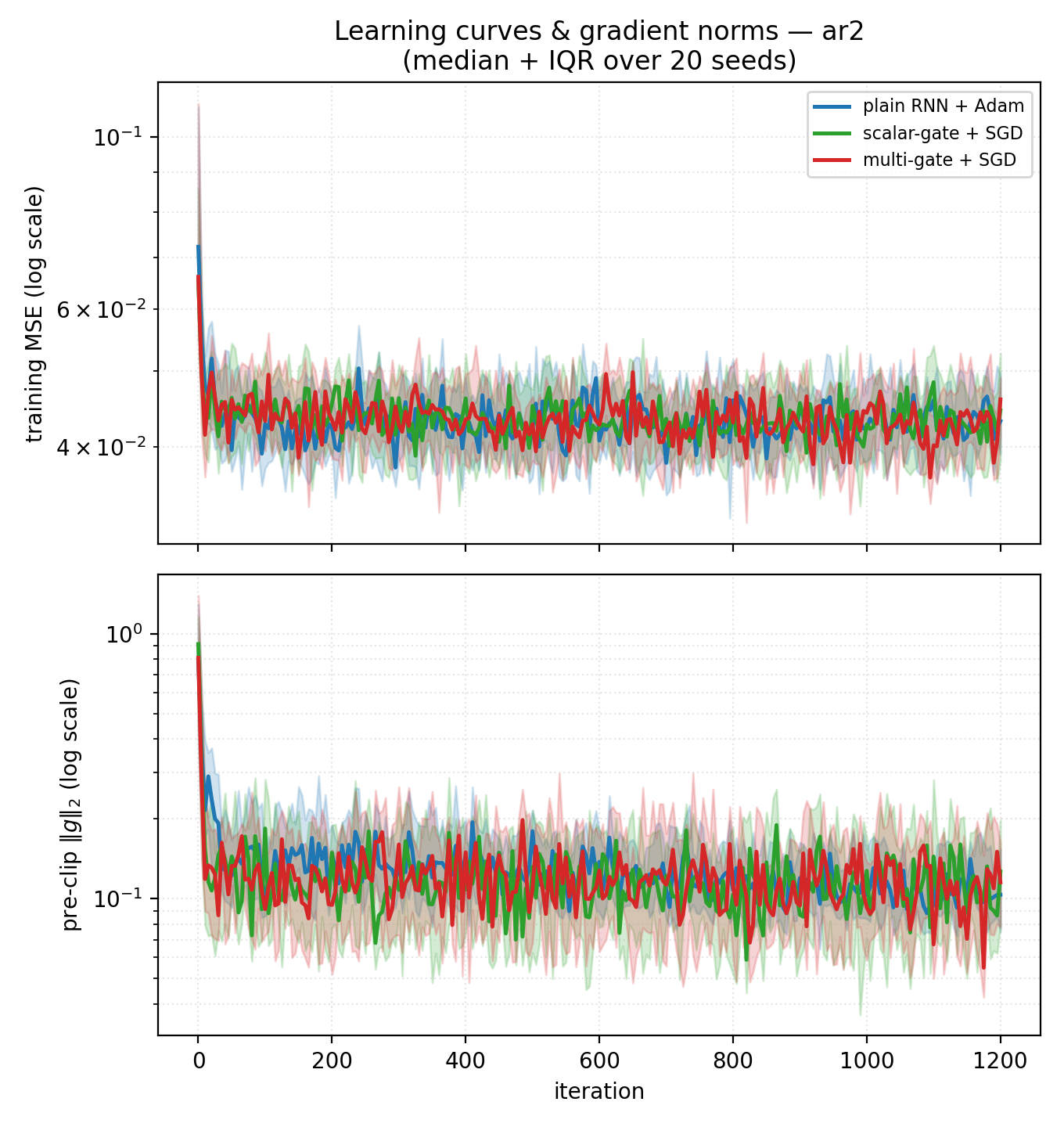}\\(b) AR(2)
  \end{minipage}

  \vspace{0.6em}

  \begin{minipage}[t]{0.33\linewidth}
    \centering
    \includegraphics[width=\linewidth]{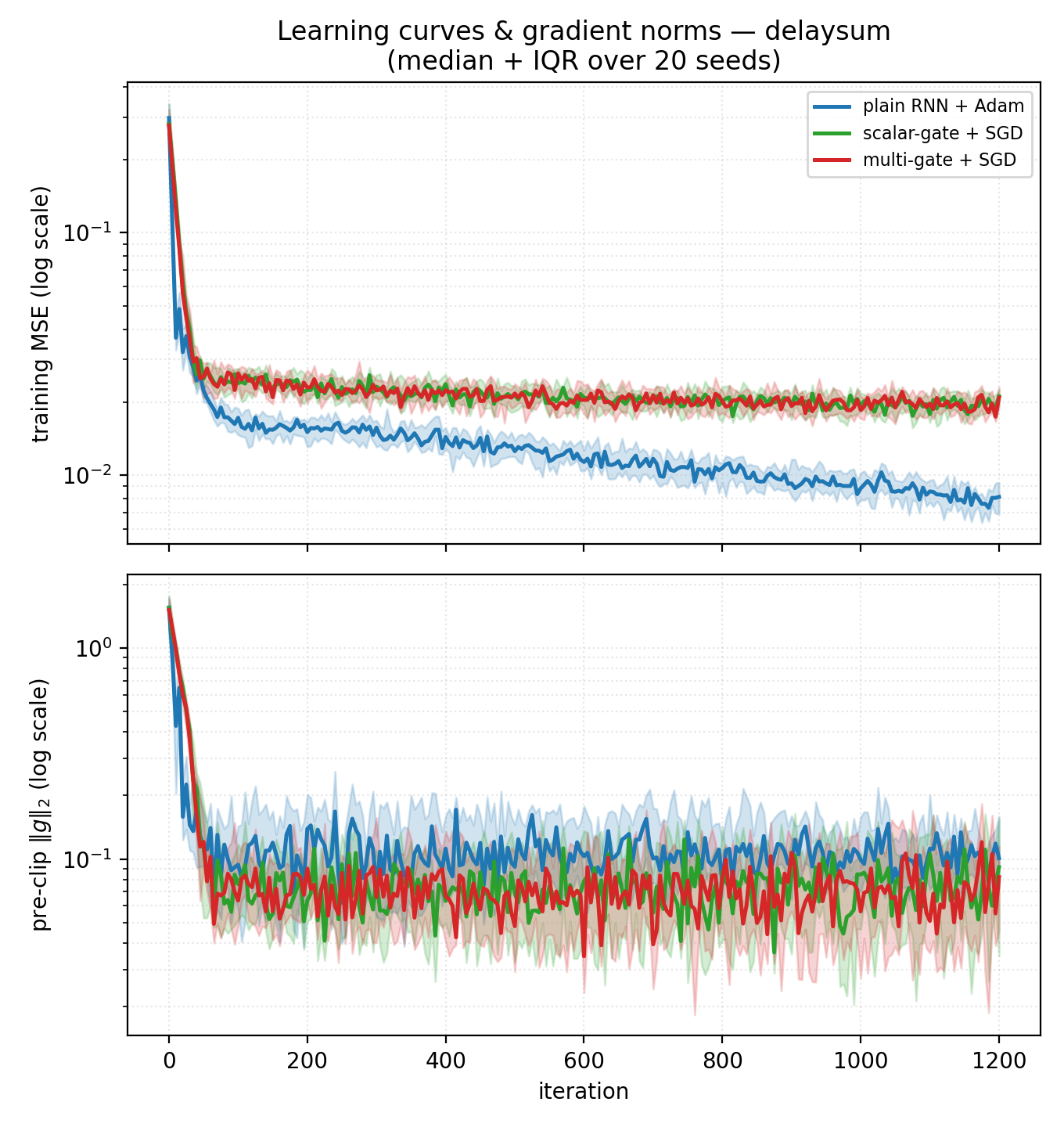}\\(c) delay-sum
  \end{minipage}\hfill
  \begin{minipage}[t]{0.33\linewidth}
    \centering
    \includegraphics[width=\linewidth]{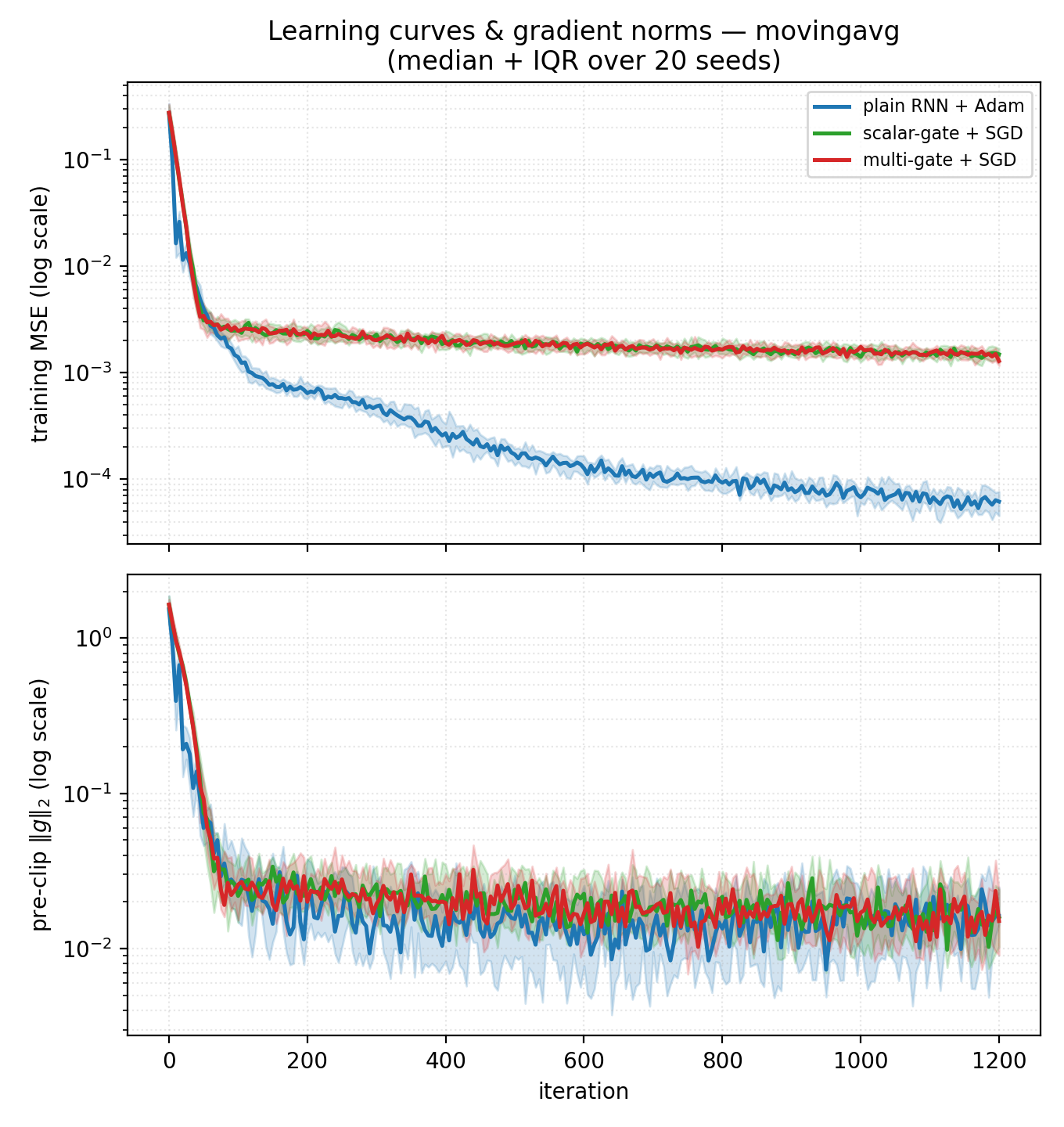}\\(d) moving-average
  \end{minipage}

  \vspace{0.6em}

  \makebox[\linewidth]{%
    \begin{minipage}[t]{0.33\linewidth}
      \centering
      \includegraphics[width=\linewidth]{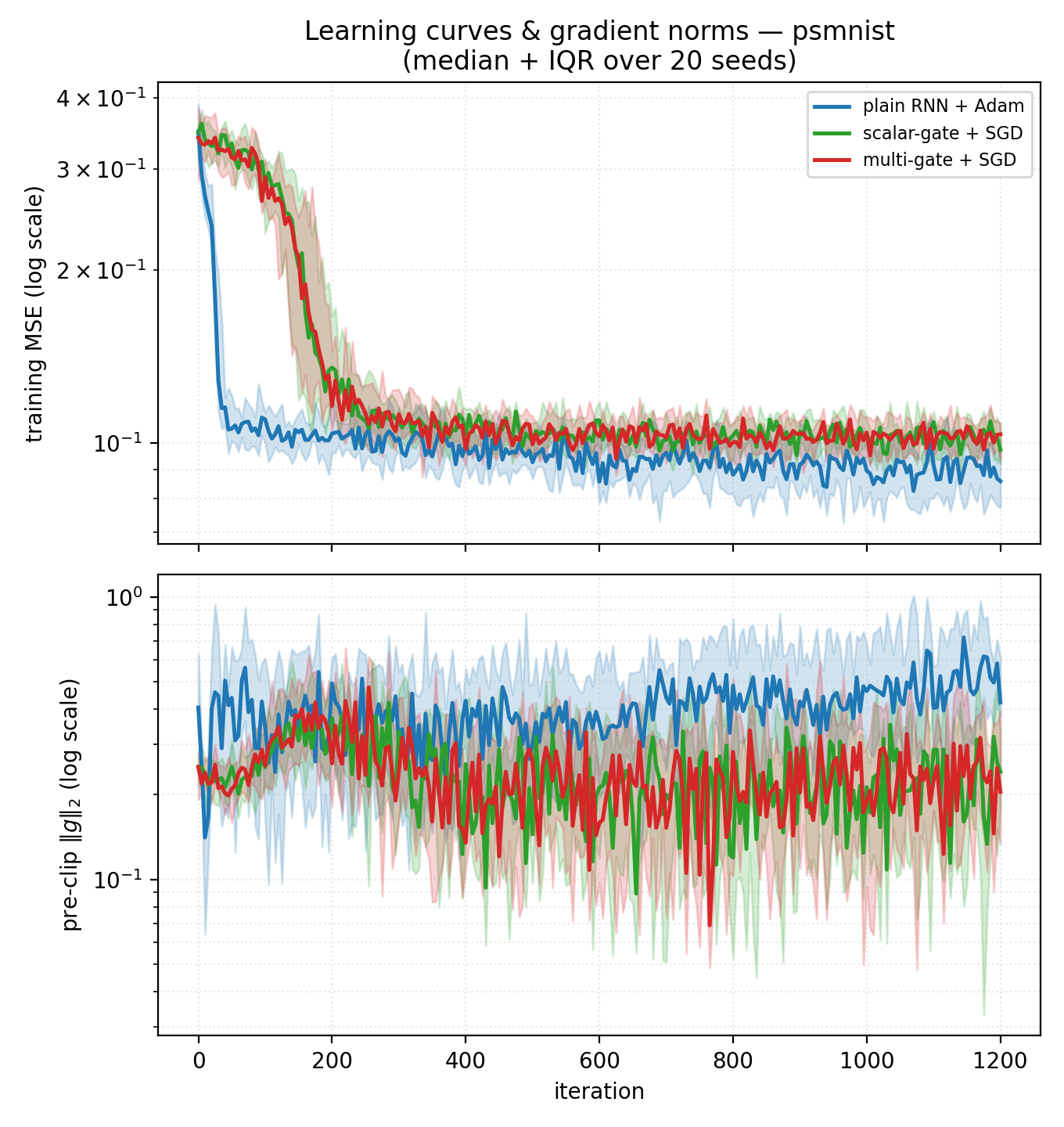}\\(e) psMNIST
    \end{minipage}%
  }
  \caption{Optimization diagnostics for the primary (no-clip) S2 protocol. For each task, the top subpanel shows the per-iteration training MSE and the bottom subpanel shows the per-iteration gradient L2 norm, aggregated as median $\pm$ IQR over $20$ random seeds for each of the three architecture+optimizer pairs (plain+Adam, scalar+SGD, multi+SGD). Losses decrease on every task/architecture pair; gradient norms remain bounded throughout training. psMNIST plain+Adam exhibits the largest transient gradient norms -- a small number of seeds reach max $\|g\|_2 \gtrsim 10$ during early training -- but the loss remains well behaved on those seeds.}
  \label{fig:s2_lossgrad_noclip}
\end{figure}

The resulting trajectories are reported per task in Fig.~\ref{fig:s2_lossgrad_noclip}. Across all five tasks and all three architecture+optimizer pairs, training MSE shows a clear downward trend over training, up to stochastic fluctuations inherent to online SGD: the median relative loss reduction over $1200$ iterations ranges from $\sim\!34\text{--}40\%$ on AR(2) (a task with low target signal-to-noise ratio) to $\sim\!85\%$ on adding, $\sim\!93\text{--}97\%$ on delay-sum, and $\sim\!99.4\text{--}100\%$ on moving-average; gated psMNIST reaches a stable $\sim\!70\%$ reduction across all seeds. Sampled gradient L2 norms remain finite throughout training on every (task, architecture, seed) configuration. On the short-$T$ tasks and on gated psMNIST, per-seed maxima are tightly concentrated: median-over-seeds maxima (Table~\ref{tab:s2_gradnorm}) fall in the range $0.8\text{--}5.2$, with per-seed IQR widths typically below $1$ and global maxima below $8$. The only visibly heavy-tailed condition is psMNIST plain+Adam, where a small number of seeds ($6/20$) exhibit transient gradient spikes with max $\|g\|_2>10$ during early training; nevertheless, the loss remains well behaved on those seeds, and the median relative loss reduction ($\sim\!74\%$) is comparable to that of the gated psMNIST configurations.

Comparing the primary (no-clip) and robustness (clipped) protocols, the optimization diagnostics are essentially unchanged: on the short-$T$ tasks the clipped and no-clip summaries agree to rounding (individual trajectories may diverge slightly because clipping fires on the upper tail, but not enough to shift the loss-reduction medians or the max-norm envelopes), and on psMNIST the gated-SGD runs are identical to rounding while plain+Adam shows a small loss-reduction improvement ($\sim\!74\%\!\to\!76\%$) with a qualitatively similar heavy tail (transient max $\|g\|_2$ up to a few hundred on two seeds). These diagnostics therefore indicate that clipping is not driving the reported S2 geometry, and that the gated models in particular have tightly bounded gradient norms and stable loss reductions across all seeds.

\begin{table}[th!]
  \centering\small
  \caption{Gradient L2 norms under the primary (no-clip) S2 protocol, aggregated over $20$ seeds. Per-seed maximum and mean are computed over $1200$ iterations; entries are medians across seeds with the seed-level interquartile range (IQR) in brackets. Values are finite and modest throughout, consistent with Fig.~\ref{fig:s2_lossgrad_noclip}. A uniform clip at norm $1.0$ (robustness check) fires on $0$--$11\%$ of iterations (highest on adding plain+Adam, essentially zero on AR(2) and gated psMNIST), and the resulting gated-configuration AI values are within a fraction of a percent of these no-clip numbers (CE essentially unchanged). Only psMNIST plain+Adam shows heavy-tailed seed variability; see text.}
  \label{tab:s2_gradnorm}
  \begin{tabular}{llll}
  \toprule
  Task & Arch & $\max \|g\|_2$ med.\,[IQR] & mean $\|g\|_2$ med.\,[IQR] \\
  \midrule
  adding      & plain+Adam  & $5.23\,[4.53, 5.94]$ & $0.513\,[0.493, 0.552]$ \\
  adding      & scalar+SGD  & $5.22\,[4.84, 6.21]$ & $0.368\,[0.359, 0.379]$ \\
  adding      & multi+SGD   & $5.10\,[4.85, 5.98]$ & $0.368\,[0.354, 0.381]$ \\
  AR(2)       & plain+Adam  & $0.80\,[0.56, 1.29]$ & $0.148\,[0.143, 0.156]$ \\
  AR(2)       & scalar+SGD  & $0.91\,[0.56, 1.15]$ & $0.140\,[0.133, 0.149]$ \\
  AR(2)       & multi+SGD   & $0.81\,[0.49, 1.39]$ & $0.144\,[0.135, 0.153]$ \\
  delay-sum   & plain+Adam  & $1.56\,[1.16, 1.73]$ & $0.135\,[0.131, 0.140]$ \\
  delay-sum   & scalar+SGD  & $1.55\,[1.23, 1.77]$ & $0.109\,[0.103, 0.111]$ \\
  delay-sum   & multi+SGD   & $1.51\,[1.22, 1.71]$ & $0.106\,[0.102, 0.111]$ \\
  moving-avg. & plain+Adam  & $1.53\,[1.25, 1.88]$ & $0.038\,[0.036, 0.040]$ \\
  moving-avg. & scalar+SGD  & $1.63\,[1.23, 1.86]$ & $0.050\,[0.047, 0.053]$ \\
  moving-avg. & multi+SGD   & $1.65\,[1.29, 1.77]$ & $0.050\,[0.047, 0.052]$ \\
  psMNIST     & plain+Adam  & $5.04\,[1.96, 12.51]$ & $0.462\,[0.414, 0.611]$ \\
  psMNIST     & scalar+SGD  & $1.05\,[0.95, 1.10]$ & $0.268\,[0.262, 0.278]$ \\
  psMNIST     & multi+SGD   & $1.02\,[0.98, 1.08]$ & $0.277\,[0.267, 0.287]$ \\
  \bottomrule
  \end{tabular}
\end{table}

\clearpage
\paragraph{Results and figure}

\begin{figure}[th!]
  \centering
  \begin{minipage}[t]{0.48\linewidth}
    \centering
    \includegraphics[width=\linewidth]{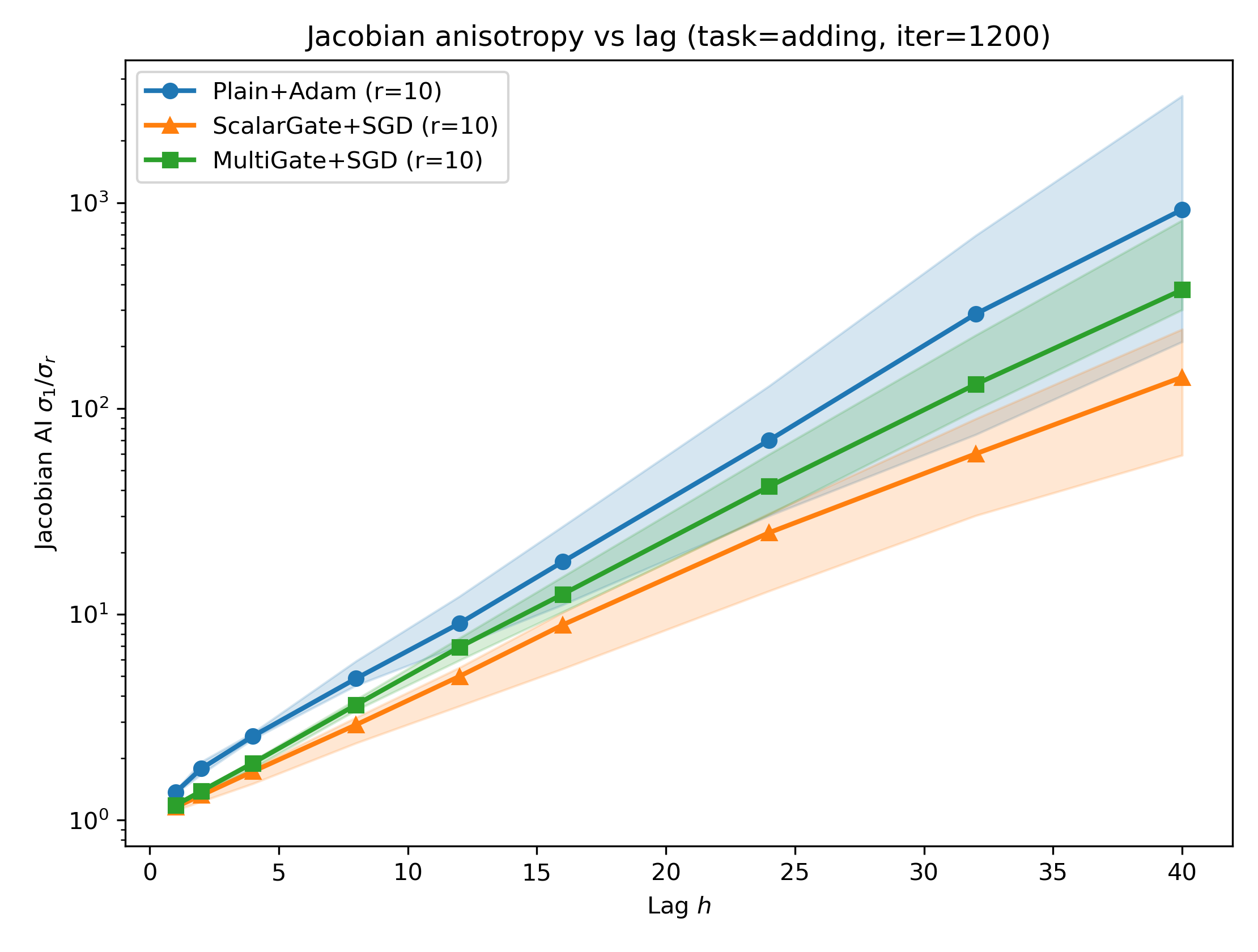}
  \end{minipage}\hfill
  \begin{minipage}[t]{0.48\linewidth}
    \centering
    \includegraphics[width=\linewidth]{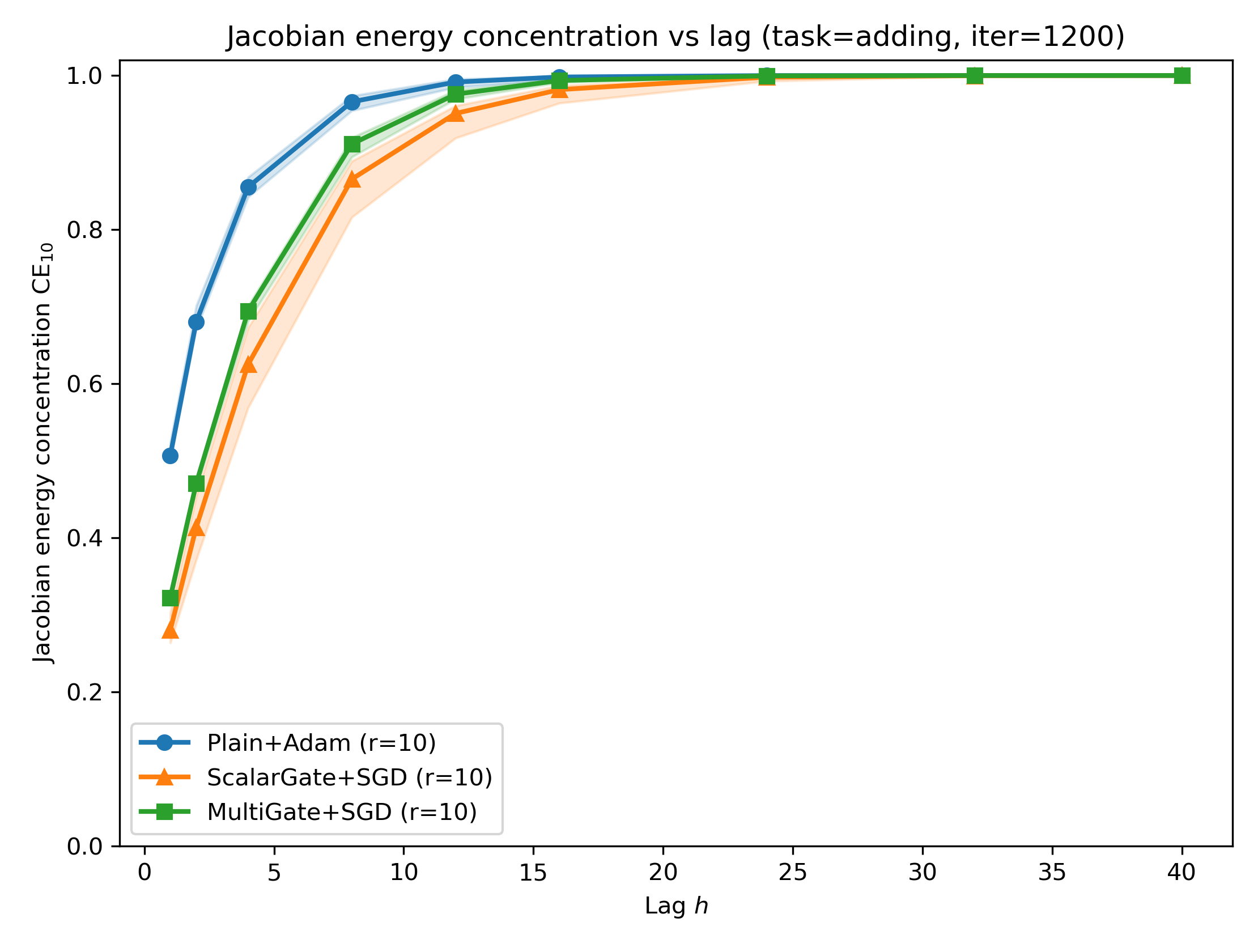}
  \end{minipage}

  \vspace{0.6em}

  \makebox[\linewidth]{%
    \begin{minipage}[t]{0.7\linewidth}
      \centering
      \includegraphics[width=\linewidth]{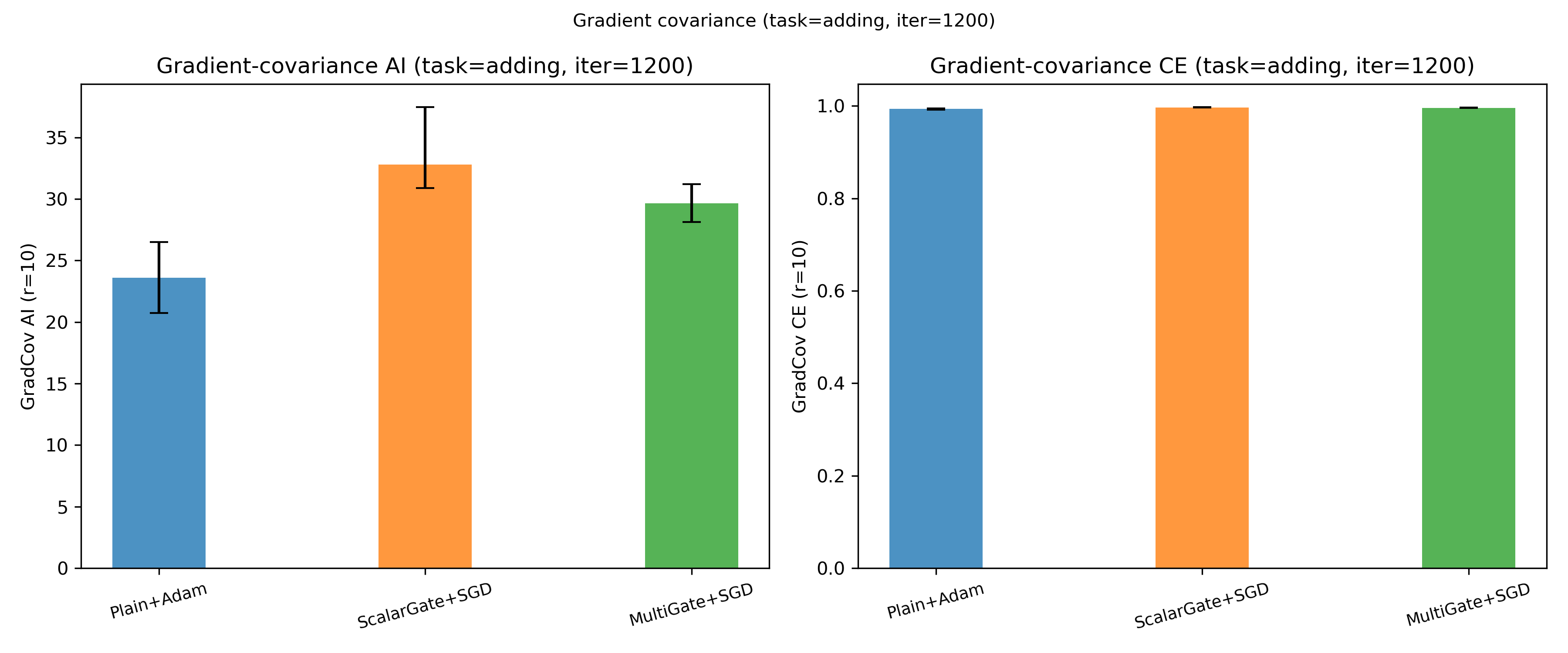}
    \end{minipage}
  }
  \caption{Adding task. Left/middle: Jacobian AI rises with lag; scalar gating shows the one sharp CE$_{10}$ drop at the longest lags. Bottom: gated updates are more concentrated than plain+Adam, with the narrowest gap among the tasks (AI$_{10}(G)\!\approx\!33/30$ scalar/multi vs.\ $24$ plain+Adam) but preserved across every seed.}
  \label{fig:s2_add}
\end{figure}

\begin{figure}[th!]
  \centering
  \begin{minipage}[t]{0.48\linewidth}
    \centering
    \includegraphics[width=\linewidth]{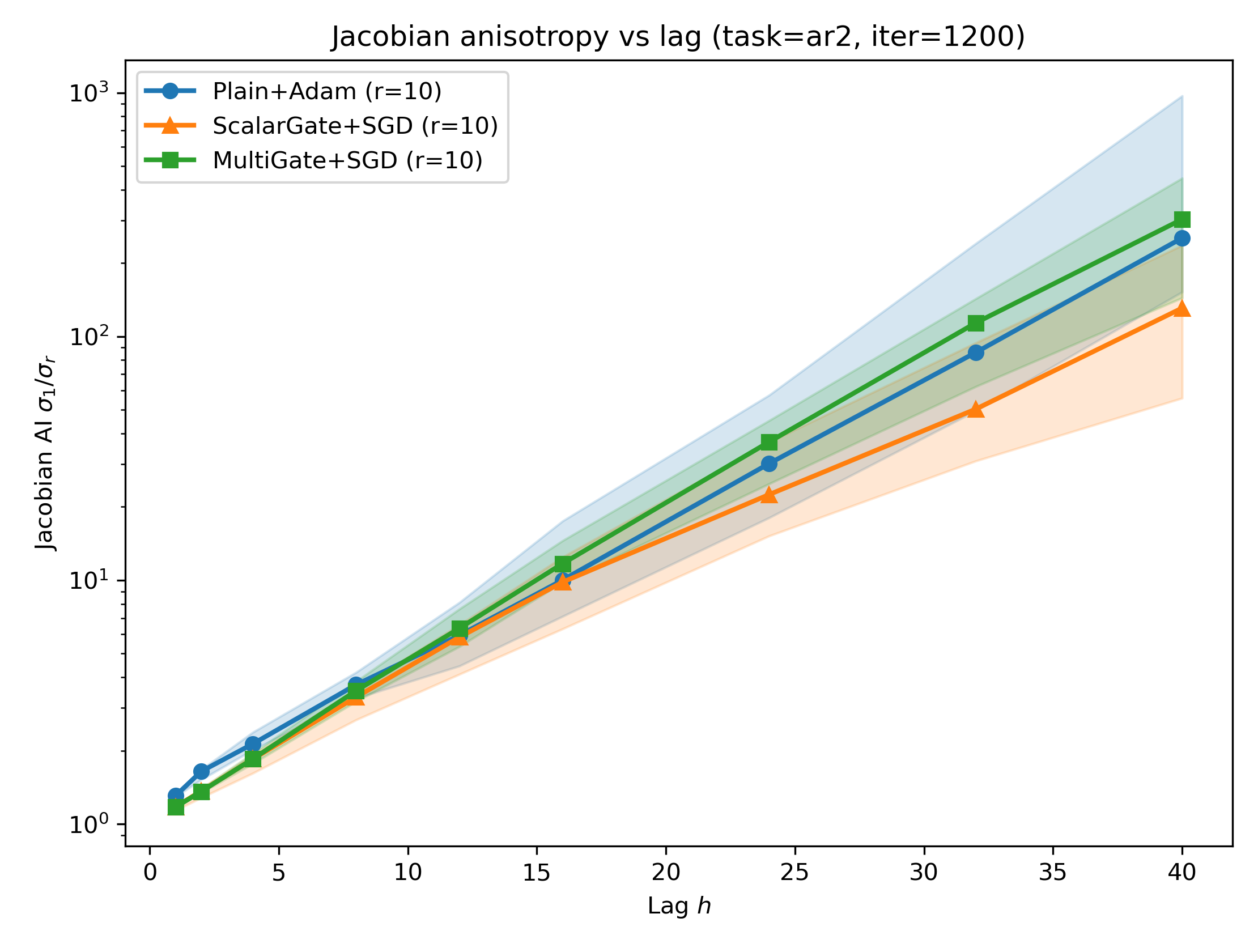}
  \end{minipage}\hfill
  \begin{minipage}[t]{0.48\linewidth}
    \centering
    \includegraphics[width=\linewidth]{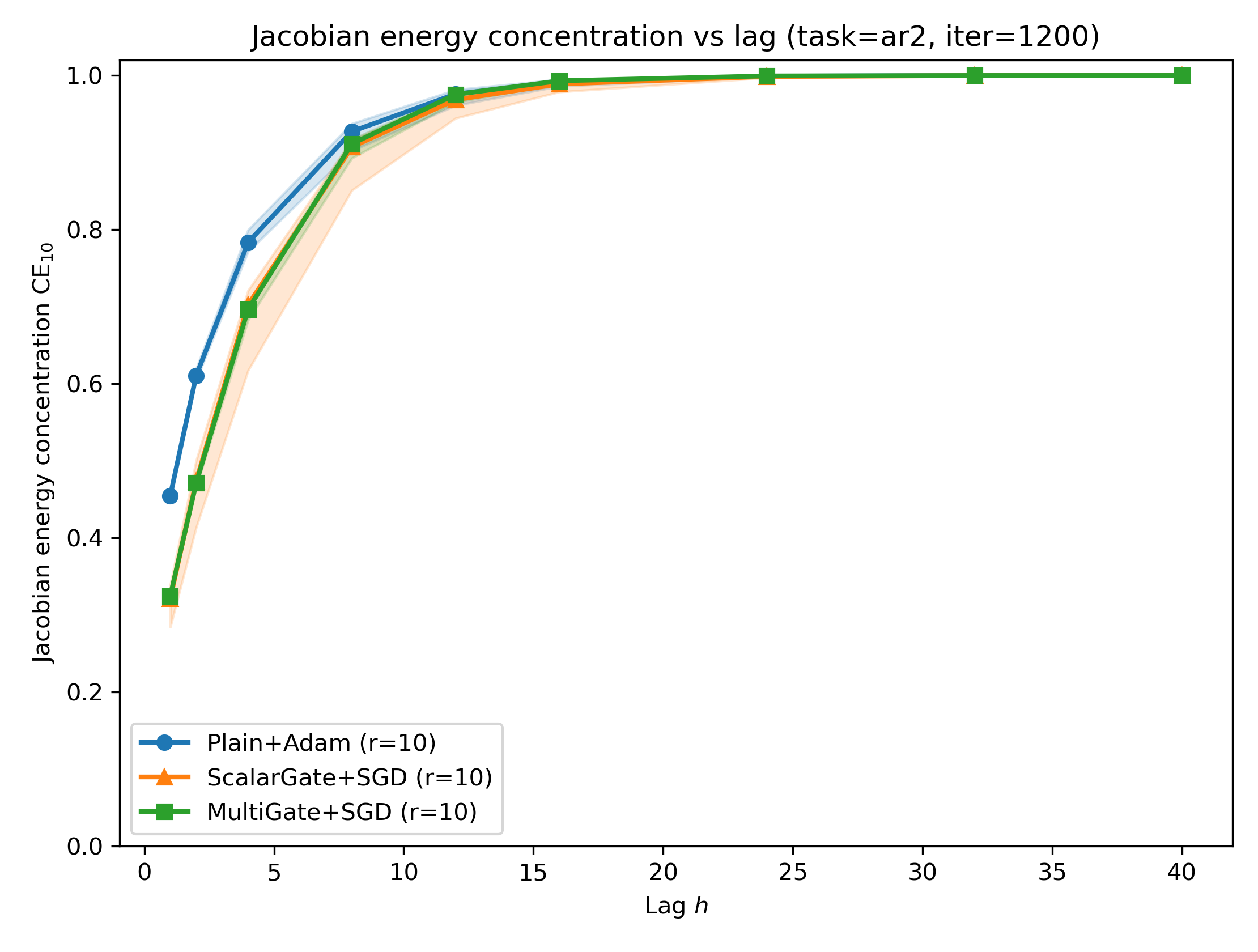}
  \end{minipage}

  \vspace{0.6em}

  \makebox[\linewidth]{%
    \begin{minipage}[t]{0.7\linewidth}
      \centering
      \includegraphics[width=\linewidth]{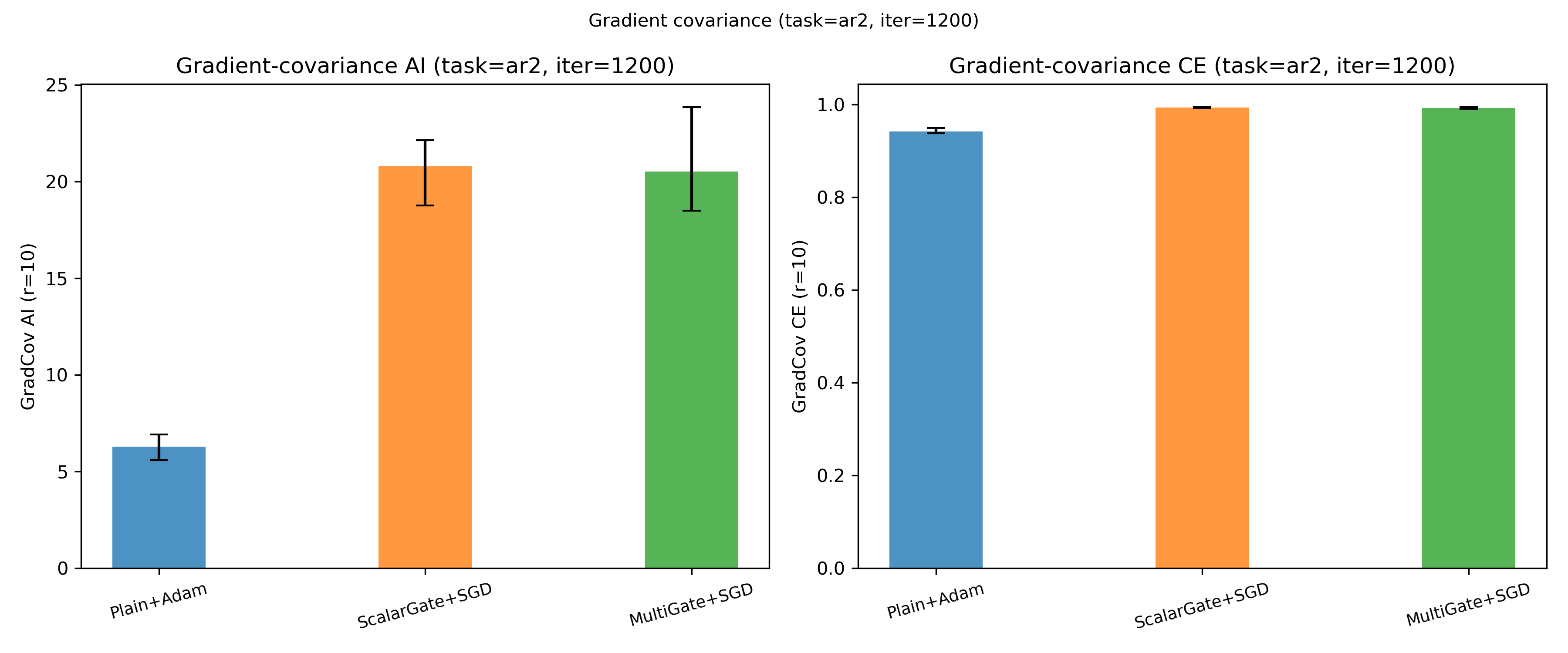}
    \end{minipage}
  }
  \caption{AR(2). Jacobian AI rises steeply with lag and CE$_{10}$ stays near $1$ for all models. Update anisotropy from gradient covariance is markedly stronger under gating (AI$_{10}(G)\!\approx\!21$ scalar, $21$ multi) than under plain+Adam ($\approx\!6.3$).}
  \label{fig:s2_ar2}
\end{figure}

\begin{figure}[th!]
  \centering
  \begin{minipage}[t]{0.48\linewidth}
    \centering
    \includegraphics[width=\linewidth]{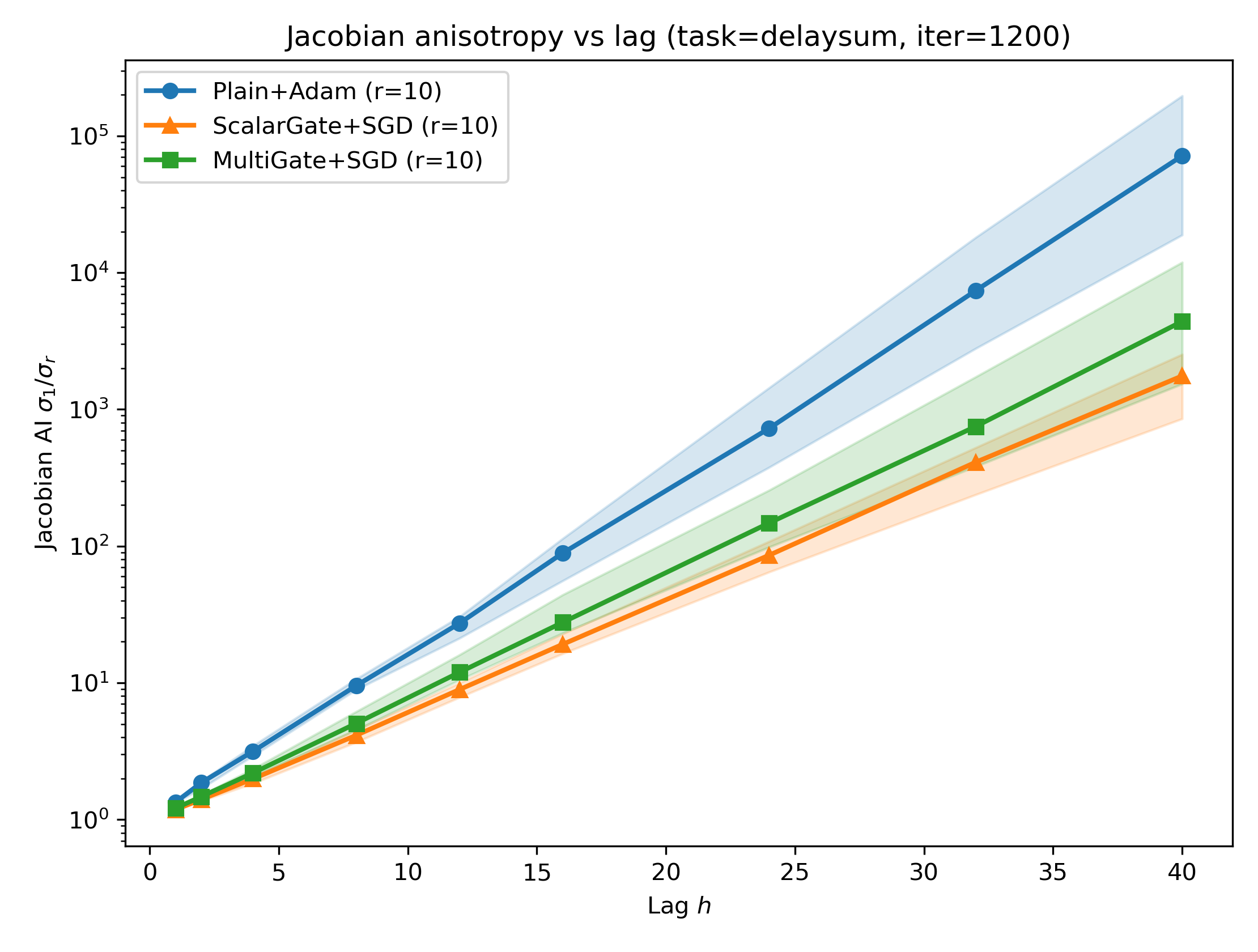}
  \end{minipage}\hfill
  \begin{minipage}[t]{0.48\linewidth}
    \centering
    \includegraphics[width=\linewidth]{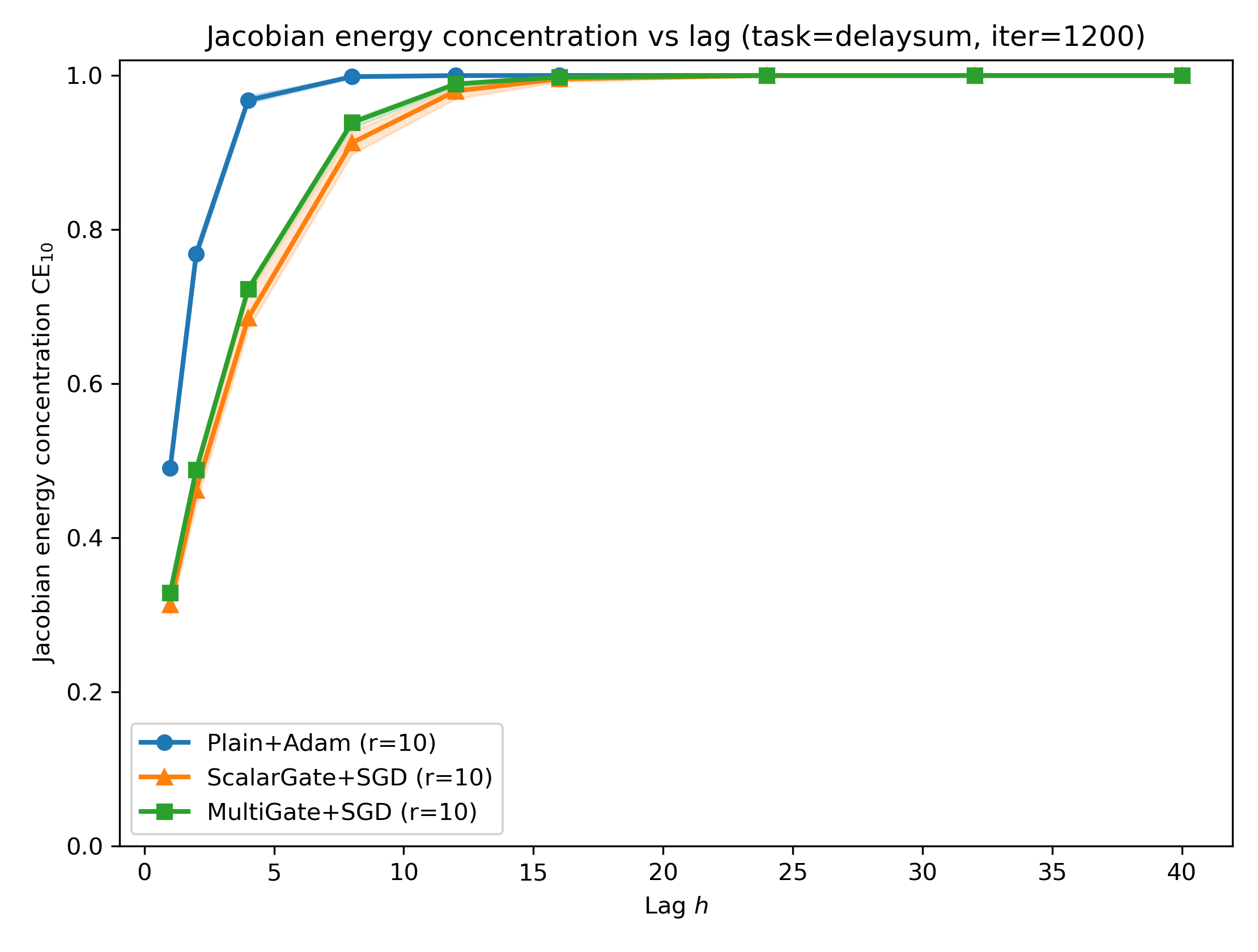}
  \end{minipage}

  \vspace{0.6em}

  \makebox[\linewidth]{%
    \begin{minipage}[t]{0.7\linewidth}
      \centering
      \includegraphics[width=\linewidth]{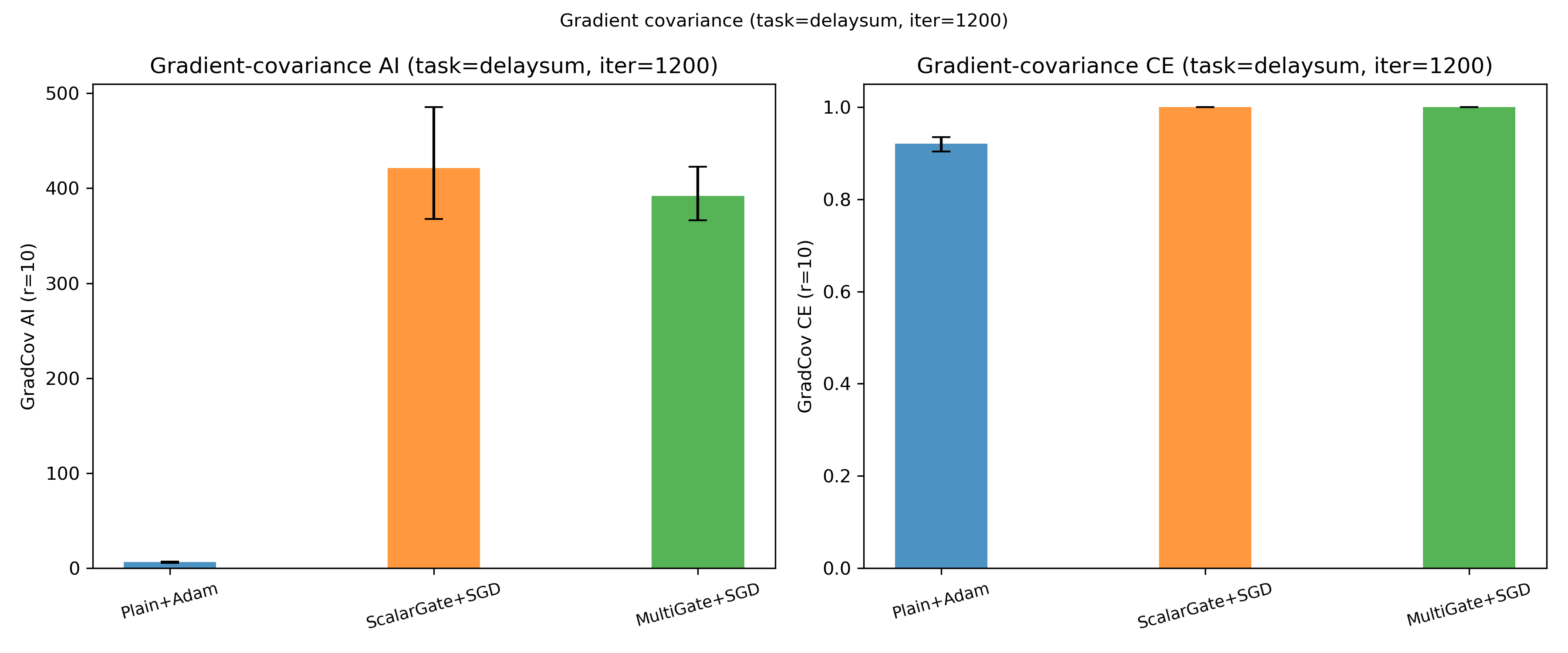}
    \end{minipage}
  }
  \caption{Delay-sum. Jacobian AI grows sharply with lag at near-unit CE$_{10}$. Update anisotropy is extreme for gated models (AI$_{10}(G)\!\approx\!421$ scalar, $392$ multi) versus plain+Adam ($\approx\!6.4$) -- the largest gated-vs.-plain+Adam gap among the synthetic tasks.}
  \label{fig:s2_delay}
\end{figure}

\begin{figure}[th!]
  \centering
  \begin{minipage}[t]{0.48\linewidth}
    \centering
    \includegraphics[width=\linewidth]{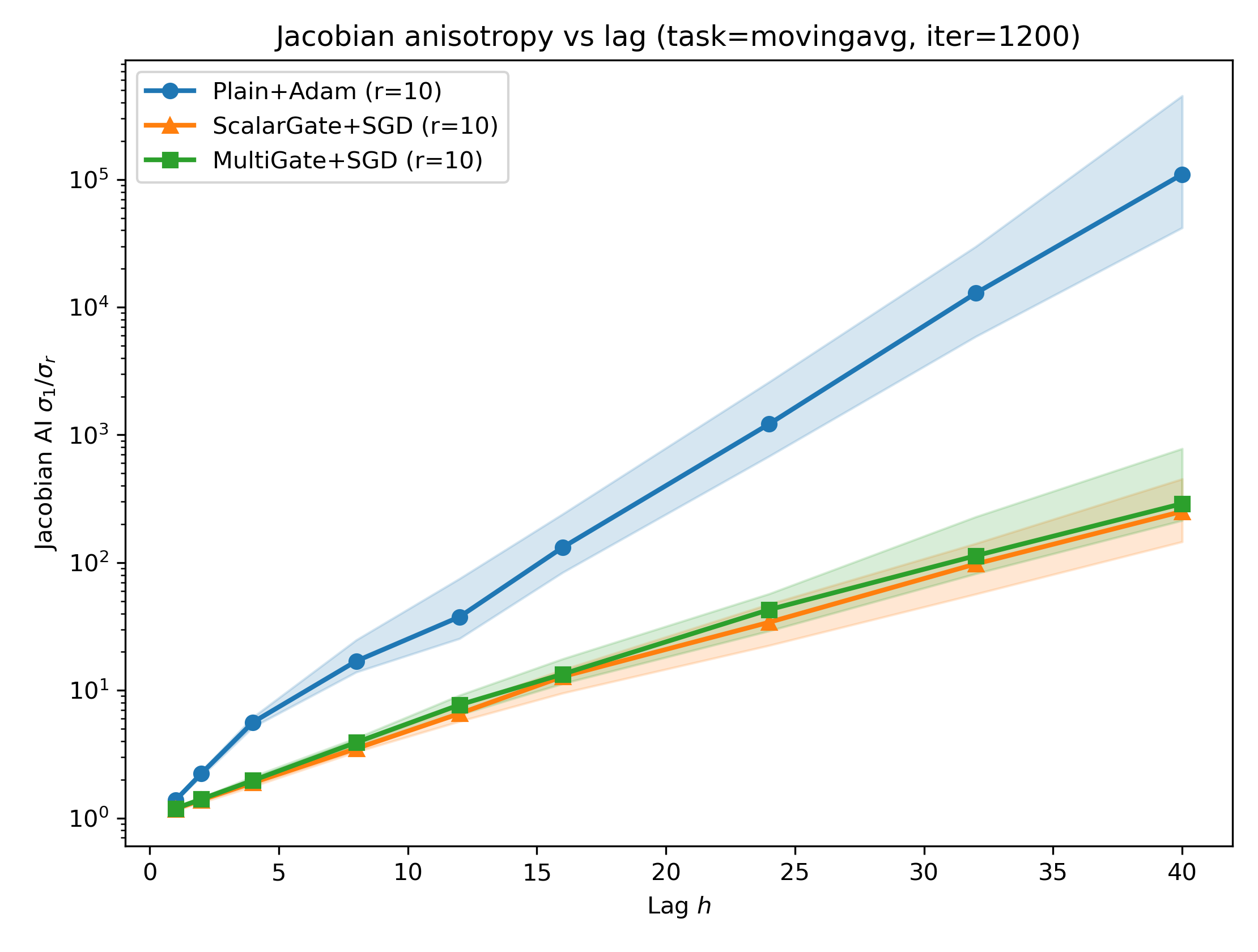}
  \end{minipage}\hfill
  \begin{minipage}[t]{0.48\linewidth}
    \centering
    \includegraphics[width=\linewidth]{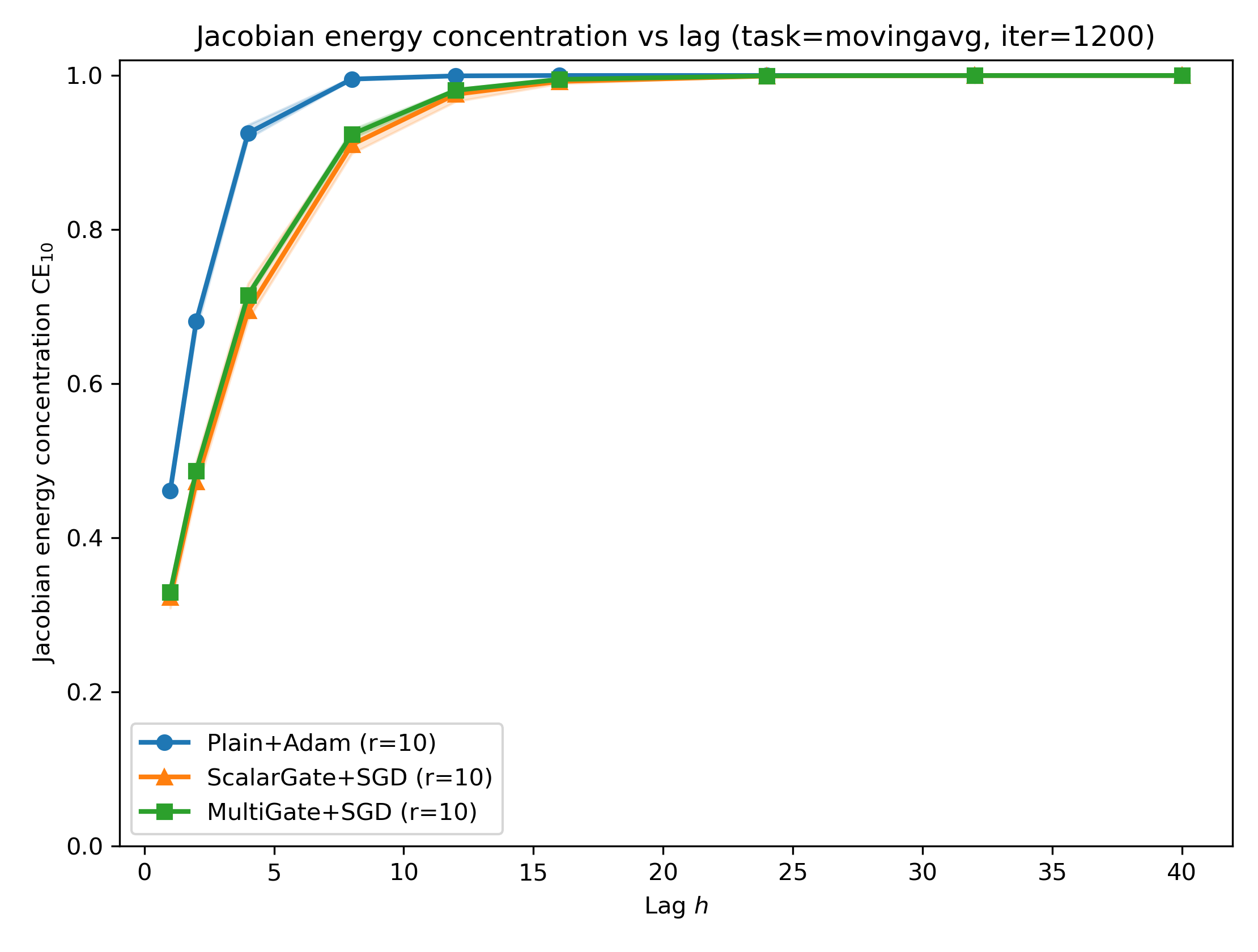}
  \end{minipage}

  \vspace{0.6em}

  \makebox[\linewidth]{%
    \begin{minipage}[t]{0.48\linewidth}
      \centering
      \includegraphics[width=\linewidth]{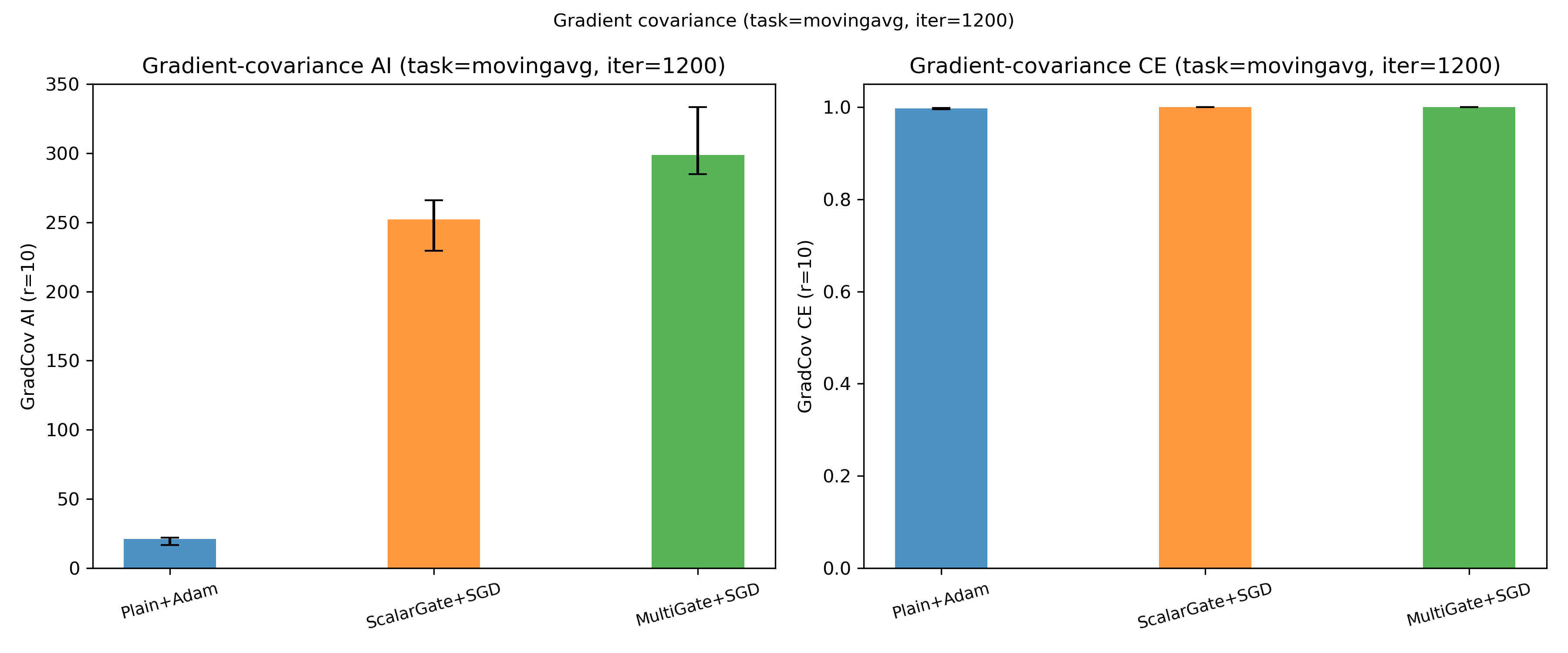}
    \end{minipage}
  }
  \caption{Moving-average. Jacobian AI saturates at long lags with CE$_{10}$ near $1$. Multi-gate shows the strongest update concentration (AI$_{10}(G)\!\approx\!299$), scalar a close second ($\approx\!252$), and plain+Adam an order of magnitude lower ($\approx\!21$).}
  \label{fig:s2_ma}
\end{figure}

\begin{figure}[th!]
  \centering
  \begin{minipage}[t]{0.48\linewidth}
    \centering
    \includegraphics[width=\linewidth]{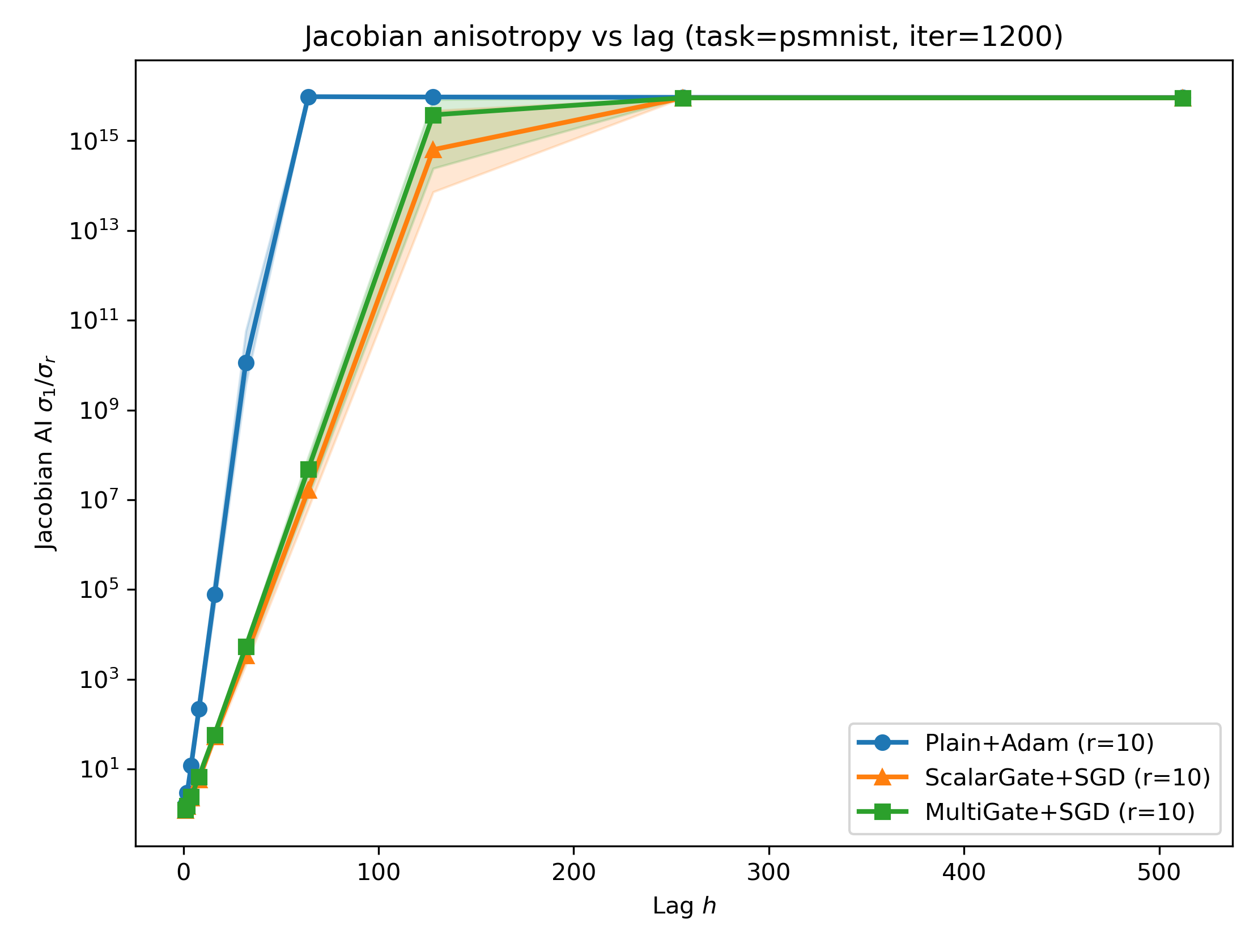}
  \end{minipage}\hfill
  \begin{minipage}[t]{0.48\linewidth}
    \centering
    \includegraphics[width=\linewidth]{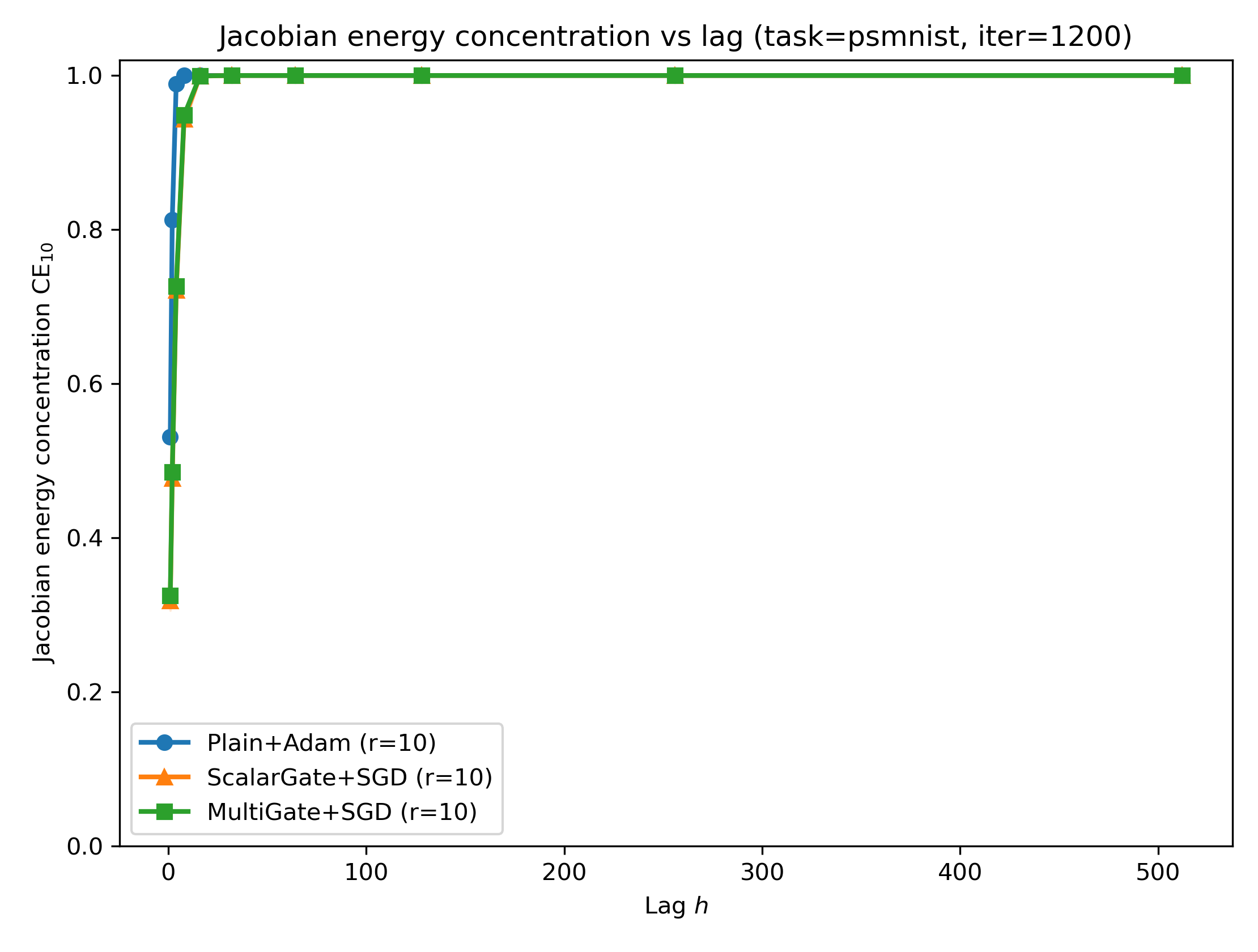}
  \end{minipage}

  \vspace{0.6em}

  \makebox[\linewidth]{%
    \begin{minipage}[t]{0.48\linewidth}
      \centering
      \includegraphics[width=\linewidth]{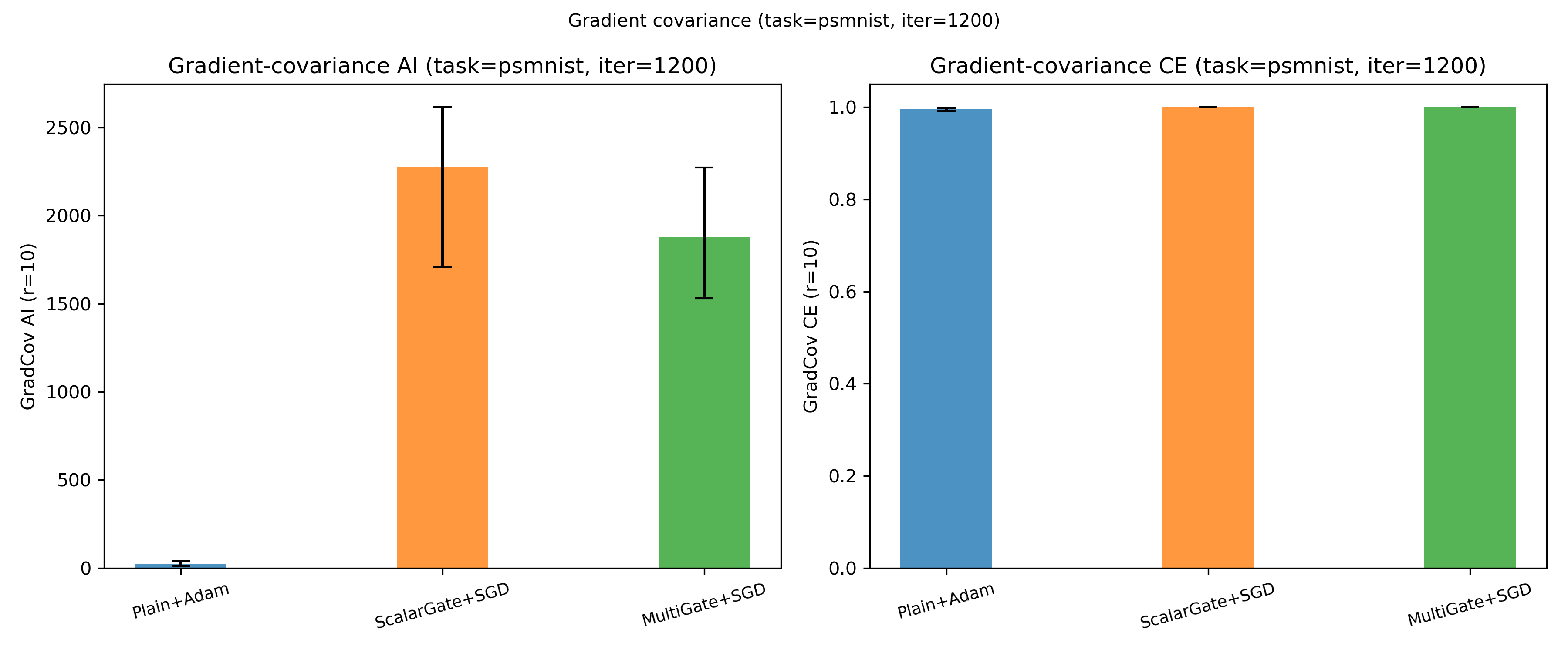}
    \end{minipage}
  }
  \caption{psMNIST (regression surrogate, $T=784$). Long-lag propagation and gradient-covariance anisotropy follow the same qualitative pattern as on the synthetic tasks, with gated models inducing substantially stronger update anisotropy than plain+Adam (AI$_{10}(G)\!\approx\!2280$ scalar, $1880$ multi, vs.\ $\approx\!22$ plain+Adam). Because the Jacobian product is accumulated in float64, its condition number is upper-bounded by $\sim\!1/\varepsilon_{\mathrm{mach}}\approx 10^{16}$; this bound is approached near lag $h\!\approx\!64$ for plain+Adam and $h\!\approx\!128\text{--}256$ for scalar/multi gates, so Jacobian AI beyond those lags should be read as saturated (qualitative) rather than as a quantitative condition-number estimate. Gradient-covariance AI at the final checkpoint, which does not depend on lag, is unaffected by this ceiling.}
  \label{fig:s2_psmnist}
\end{figure}

The optimization diagnostics confirm that the spectral quantities reported in the Figures~\ref{fig:s2_add}--\ref{fig:s2_psmnist} are computed on non-degenerate, numerically well-behaved optimization trajectories.

\clearpage
\subsection{Code availability}
\label{app:code}

The code used to reproduce the experiments reported in this paper is available at

\begin{center}
\url{https://github.com/lorenzolivi/effective-learning-rates}
\end{center}

The repository includes scripts and instructions for reproducing the results.